\documentclass[9pt,lineno]{elife}

\usepackage{lipsum}
\usepackage[version=4]{mhchem}
\usepackage{siunitx}
\usepackage[font=small,labelfont=bf,
   justification=justified,
   format=plain]{caption}
\DeclareSIUnit\Molar{M}

\usepackage{amsmath}
\usepackage{bm}
\DeclareMathOperator*{\argmin}{arg\,min}
\DeclareMathOperator*{\argmax}{arg\,max}
\DeclareMathOperator*{\mean}{mean}
\DeclareMathOperator*{\median}{median}

\def\rot{\rotatebox}
\usepackage{array}
\usepackage{multirow}

\usepackage[linesnumbered]{algorithm2e}

\definecolor{c0}{rgb}{0.12156862745098039, 0.4666666666666667, 0.7058823529411765}
\definecolor{c1}{rgb}{1.0, 0.4980392156862745, 0.054901960784313725}
\usepackage{pgf}
\usepackage{tikz}
\usepackage{varwidth}
\let\pgfimageWithoutPath\pgfimage 
\renewcommand{\pgfimage}[2][]{\pgfimageWithoutPath[#1]{figures/#2}}

\usepackage{siunitx}
\usepackage{nameref}
\usepackage{layouts}
\newcommand{\name}{\textsc{Decompose}}
\newcommand{\git}{\url{https://github.com/bethgelab/decompose}}
\def\eg{e.g.~}
\def\ie{i.e.~}
\newcommand{\topic}[1]{}

\title{Trace your sources in large-scale data: one ring to find them all}

\author[1*]{Alexander B{\"o}ttcher}
\author[1]{Wieland Brendel}
\author[3]{Bernhard Englitz}
\author[1,2*]{Matthias Bethge}
\affil[1]{Werner Reichardt Center for Integrative Neuroscience, Eberhard Karls Universität Tübingen, Bernstein Center for Computational Neuroscience, Institute for Theoretical Physics, Eberhard Karls Universität Tübingen}
\affil[2]{Max Planck Institute for Biological Cybernetics, Tübingen, Germany}
\affil[3]{Department of Neurophysiology, Donders Institute for Brain, Cognition and Behaviour, Radboud University, Nijmegen, The Netherlands}

\corr{alexander.boettcher@bethgelab.org}{AB}
\corr{matthias.bethge@bethgelab.org}{MB}

\begin{document}
\maketitle
\newcommand{\ALG}[1]{\autoref{algorithm:#1}}
\renewcommand{\algorithmautorefname}{Algorithm}
\sisetup{detect-all, detect-display-math}

\begin{abstract}
An important preprocessing step in most data analysis pipelines aims to extract a small set of sources that explain most of the data. Currently used algorithms for blind source separation (BSS), however, often fail to extract the desired sources and need extensive cross-validation. In contrast, their rarely used probabilistic counterparts can get away with little cross-validation and are more accurate and reliable but no simple and scalable implementations are available. Here we present a novel probabilistic BSS framework (\name{}) that can be flexibly adjusted to the data, is extensible and easy to use, adapts to individual sources and handles large-scale data through algorithmic efficiency. \name{} encompasses and generalises many traditional BSS algorithms such as PCA, ICA and NMF and we demonstrate substantial improvements in accuracy and robustness on artificial and real data.
\end{abstract}

\section{Introduction}
\topic{What are BSS algorithms?}
Making sense of experimental data is challenging and requires flexible data analysis pipelines that are able to identify and extract meaningful dimensions in the data. A core component of most pipelines are blind source separation (BSS) algorithms such as principal component analysis (PCA) or independent component analysis (ICA). The goal of these algorithms is to identify a small set of interpretable sources that explain most of the variability in the data.

\topic{Example: desired sources in Neuroscience}
To make this more concrete we will consider applications from Neuroscience throughout this manuscript. A core question in this field is to understand the complex interactions between neurons in the nervous system. Current electrical \citep{jun2017} and optical \citep{ahrens2013} techniques record the intermingled activity of large neural ensembles, with data acquisition rates exceeding 1TB/h. In such data sets BSS aims to identify somas, dendrites, axons, and other neuropil structures by separating their signals from each other and from background noise.

\begin{figure}[t!]
    \centering
    \input{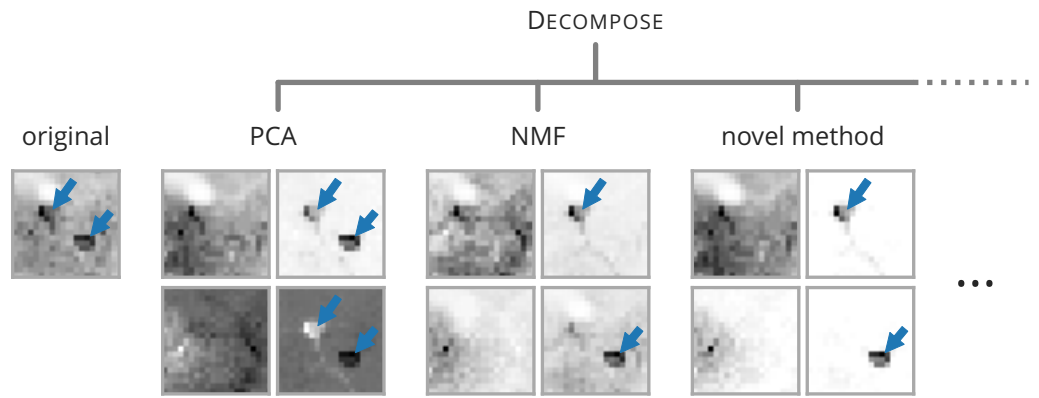}
    \captionof{figure}{
    \name{} consists of a large number of methods including many established ones. Each method enforces certain prior assumptions that have a strong effect on the source separation. (Left) Example data from 2p calcium imaging (average intensity across space). (Center) PCA fails to extract individual cells while NMF fails to separate the a cell from the background. (Right) Novel BSS method implemented in \name{} is able to correctly identify and extract the underlying cells. \name{} encompasses many traditional methods like PCA or NMF as well as many novel ones.}
      \label{fig:pcavsnmf}
\end{figure}

\topic{Desired sources are identified from characteristic properties}
Different BSS algorithms extract different sources from the data  (see \FIG{pcavsnmf}). Which sources are the desired ones is mostly subjective and depends on expert knowledge and the goal of the analysis. This expert knowledge can often be translated into certain statistical properties one expects the desired sources to have. In the case of 2p imaging of neural populations we expect each source to correspond to a neuron, and each neuron is expected to cover only a small part of the image space (sparsity in space), to be seldomly active (sparsity in time) and to generate only non-negative calcium responses. In statistical parlance we would say that the desired sources (the neurons) feature a double-sparse (time and space) non-negative distribution. Every BSS algorithm makes different statistical assumptions about the underlying sources which explains why different algorithms often extract very different sources, \FIG{pcavsnmf}. It is thus crucial to select the right prior assumptions to find the desired sources. The core problem this paper aims to address is that currently available BSS algorithms are often too limited to incorporate these assumptions and require complex hyperparameter tuning to adapt to the data.

\topic{Probabilistic vs Deterministic BSS algorithms}
We differentiate two classes of BSS algorithms: \emph{probabilistic} and \emph{non-probabilistic} methods. A crucial difference between the two is that probabilistic methods track the uncertainty in the estimate of a source. In contrast, non-probabilistic methods track only the most likely state of source and forget about the uncertainty of its estimates. That makes non-probabilistic methods easier to implement and faster to run which is why methods from this class are ubiquitously used in practice. But compared to probabilistic methods that comes with a seldomly discussed drawback: non-probabilistic BSS algorithms require the user to hand-pick all hyperparameters. The hyperparameters are part of the priors and determine their exact shape and thus the exact statistical assumption on the sources. More concretely, a general sparsity assumption about the source can be encoded in the exponential distribution of the prior but the hyperparameters determine the exact level of sparsity we expect. Finding these hyperparameters often requires extensive cross-validation which is convoluted and error-prone \citep{bro2008}.

\topic{Fitting hyperparameters}
Probabilistic methods, in contrast, can tune the hyperparameters of the priors automatically. The reason is that probabilistic methods track the exact likelihood of all extracted sources given the data and the current hyperparameters and can thus tune everything jointly.
In contrast, non-probabilistic methods like iterated conditional modes (ICM) \citep{besag1986} are "too sure" about their estimates. They in turn get the hyperparameters wrong and thereby bias the source estimates which in turn biases the hyper-parameters even more. \cite{szeliski2008} show that ICM is prone to be trapped quickly in poor local minima.

Another key benefit of probabilistic BSS algorithms is that the priors can be adapted to each source individually. For example, this allows these algorithms to differentiate between sources with different levels of sparsity (e.g. neurons and background activity). Non-probabilistic BSS algorithms, on the other hand, typically assume that all sources share the same statistical signature in order to reduce the number of hyperparameters which the user needs to select.

\topic{Contribution of this paper}
Despite the advantages of probabilistic BSS algorithms there exist only a few isolated implementations available to researchers and data analysts \citep{schmidt2009, ahn2015, paisley2014}, none of which are widely used in practice due to their limited scopes and inconsistent interfaces. In this paper we introduce a comprehensive and unifying probabilistic BSS framework named \name{} that 1) can be flexibly adopted to the statistical signatures of the desired sources making it more robust and more accurate than current state-of-the-art BSS algorithms, 2) is easy to extend to novel prior distributions, 3) is easy to use and 4) scales well to large data sets. The framework is \emph{flexible} because it allows varied combinations of different prior assumptions that encompass and generalise a wide range of established methods including probabilistic versions of PCA, sparse PCA, ICA and non-negative matrix factorisation (NMF). The framework is \emph{easy to use} because its interface is close to the popular and widely used sklearn package and because the probabilistic implementation frees the user from parameter guessing or extensive hyperparameter cross-validation. Finally, the framework \emph{scales well} to large data sets because parameters are updated efficiently, memory is saved through an additional dimensionality reduction step and the implementation can utilise GPUs if available. The code of \name{} is available as an open source python package at \git{}.

\section{Results}  \label{sec:results}
\begin{figure}[t]
    \centering
    \input{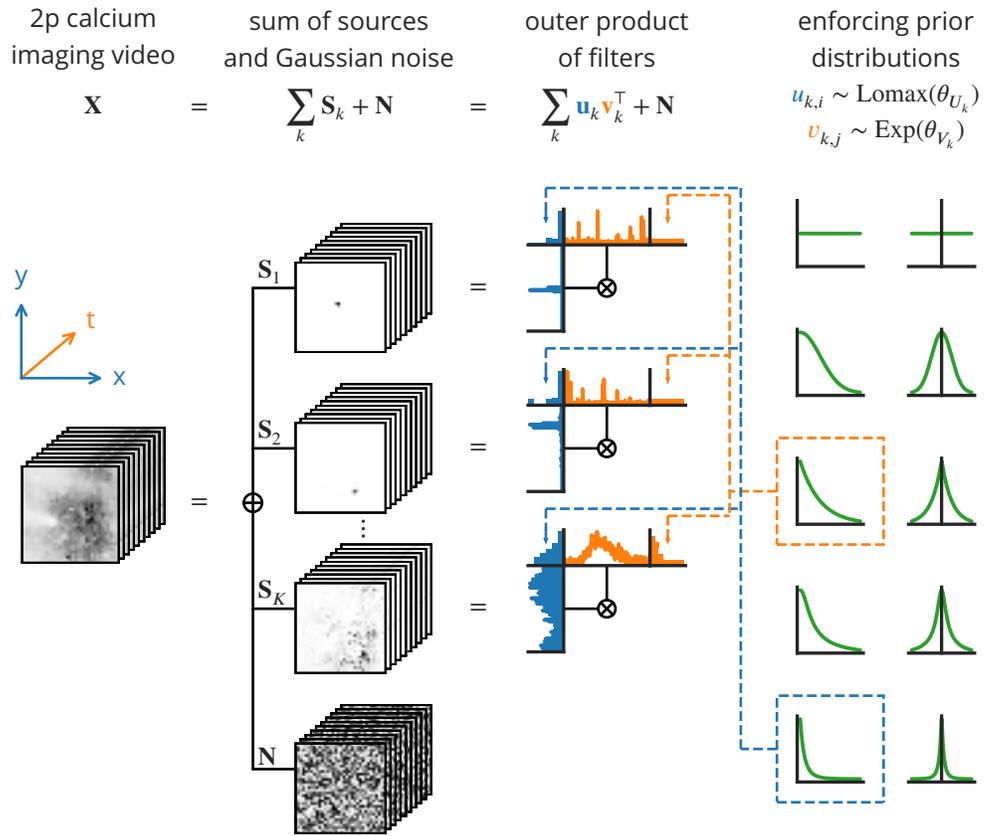}
    \captionof{figure}{
    Overview of the decomposition performed in \name{}. Datasets (left column) can have arbitrary dimension. For illustration, we show spatiotemporal data (Calcium imaging data from the mouse motor cortex, see section \ref{sec:data} for details) with two spatial dimensions $x$ and $y$ (blue) and one temporal dimension $t$ (orange). \name{} assumes an additive decomposition (middle left column) into $K$ sources and Gaussian noise, where each source $S_k$ is given by an outer product (middle right column) of spatial (blue) and temporal filters (orange) respectively. BSS aims to find sources that reconstruct the data but for which the filters (more precisely the histogram over the filter elements) follow certain prior distributions (right column).}
      \label{fig:method}
\end{figure}

\topic{Basic layout of the algorithm and the assumptions}
The probabilistic BSS algorithms implemented in \name{} assume that the true signal is corrupted by additive Gaussian noise during recording (\FIG{method}). The signal itself is assumed to be the sum of the sources. Each dimension of a source (e.g. space and time) is described by a filter or component that models the variability of that source along this dimension (\FIG{method} third column). Finally, the elements of each filter are assumed to be distributed according to a given prior distribution (e.g. Gaussian or Exponential). In other words, the histogram over the values of each filter should roughly fit to the desired prior (\FIG{method} right column). The goal of a BSS algorithm is to find sources that match the chosen prior distributions and reconstruct the data.

\begin{figure*}[t]
    \centering
    \input{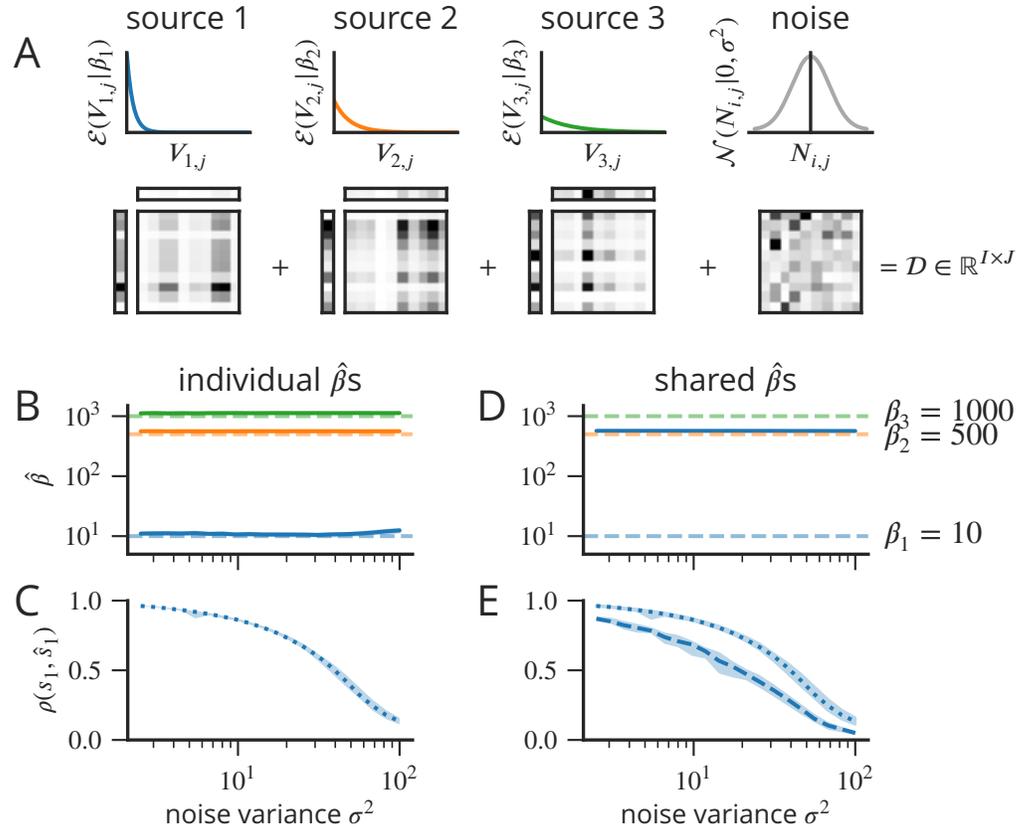}
    \caption{Automatic recovery of multiple sources with different sparsity from synthetic data.
{\bf A}~The data is generated as a sum of three non-negative sources (bottom row) with varying degrees of sparsity $\beta_i$ (distributions in top row) and an additional dense i.i.d. noise term (bottom term, right column) with variance $\sigma$.
{\bf B}~Sparsity of the individual sources as estimated by probabilistic sparse non-negative PCA for different signal-to-noise ratios (solid lines). The estimates are very close to the true sparsity values (dashed lines). The shaded region shows the (very narrow) interquartile range over N=100 samples. 
{\bf C}~Correlation of the weakest extracted source with the weakest ground-truth source. For low signal-to-noise ratios (SNR) the source is perfectly reconstructed by probabilistic sparse PCA. The performance gradually decreases for higher SNR.
{\bf D}~Same as B but we enforce the same sparsity value for all sources. This assumption is used by basically all non-probabilistic BSS algorithms used in practice.
{\bf E}~Same as C. Enforcing the same sparsity value drastically decreases the separation of the weakest source against the background (dashed). }
    \label{fig:hypest}
\end{figure*}

\subsection{Automatic hyperparameter estimation on synthetic data}\label{sec:synthetic}
\topic{Illustration on toy example} Finding the right hyperparameters for a BSS algorithm can be tedious. The hyperparameters are part of the priors and determine the statistical structure of the extracted sources. For example, in sparse PCA one needs to set the 'right' sparsity level in order to balance the sparsity of the sources and the reconstruction of the data set. If the sparsity level is set too high the sources become zero (maximally sparse) and the data is not reconstructed. If the sparsity level is too low then the data is well reconstructed but the sources are dense and do not fit well to the desired statistical signature. We illustrate the advantage of probabilistic BSS algorithms for automatic hyperparameter estimation in a concrete but simple toy example (\FIG{hypest}). Here we generated a synthetic data set that consists of only three sources with different levels of sparsity $\beta_1, \beta_2, \beta_3$ and some background noise with variance $\sigma^2$ (see \nameref{sec:data} for details). Probabilistic sparse PCA as implemented in \name{} (see Methods for details) is able to automatically discover the correct sparsity level $\beta_i$ for each of the sources over a wide range of signal-to-noise ratios (\FIG{hypest}B). That allows it to separate even the weakest source against the background noise (\FIG{hypest}C). In contrast, standard (non-probabilistic) sparse PCA implementations assume that all sources share the same level of sparsity\footnote{It would be practically unfeasible to hand-tune or cross-validate hyperparameters for each source.}. We can restrict \name{} in the same way which still allows it to discover a suitable intermediate sparsity level (\FIG{hypest}D) but makes it much more difficult to separate the weakest source from the background noise (\FIG{hypest}E).

\subsection{Performance on realistic imaging data with ground truth}\label{sec:groundturth}
\noindent The analysis above demonstrates the usefulness of data-driven hyperparameter estimation. Next we evaluate the framework on real data. To this end we choose a two-photon calcium imaging recording in the motor cortex of a behaving mouse (\cite{frady2015}, \cite{peters2014}, for details see the \nameref{sec:data} section). The duration of the recording is about two minutes and we crop an imaging window of the size \SI[product-units = repeat]{118 x 125}{\micro \metre}. The analysis has two goals: first, we want to quantify how much impact the prior has on the extraction of single cells. Second, we want to compare the quality of the reconstruction achieved by \name{} against commonly used non-probabilistic algorithms like ICA or NMF.

\begin{figure}[!ht]
    \centering

    \input{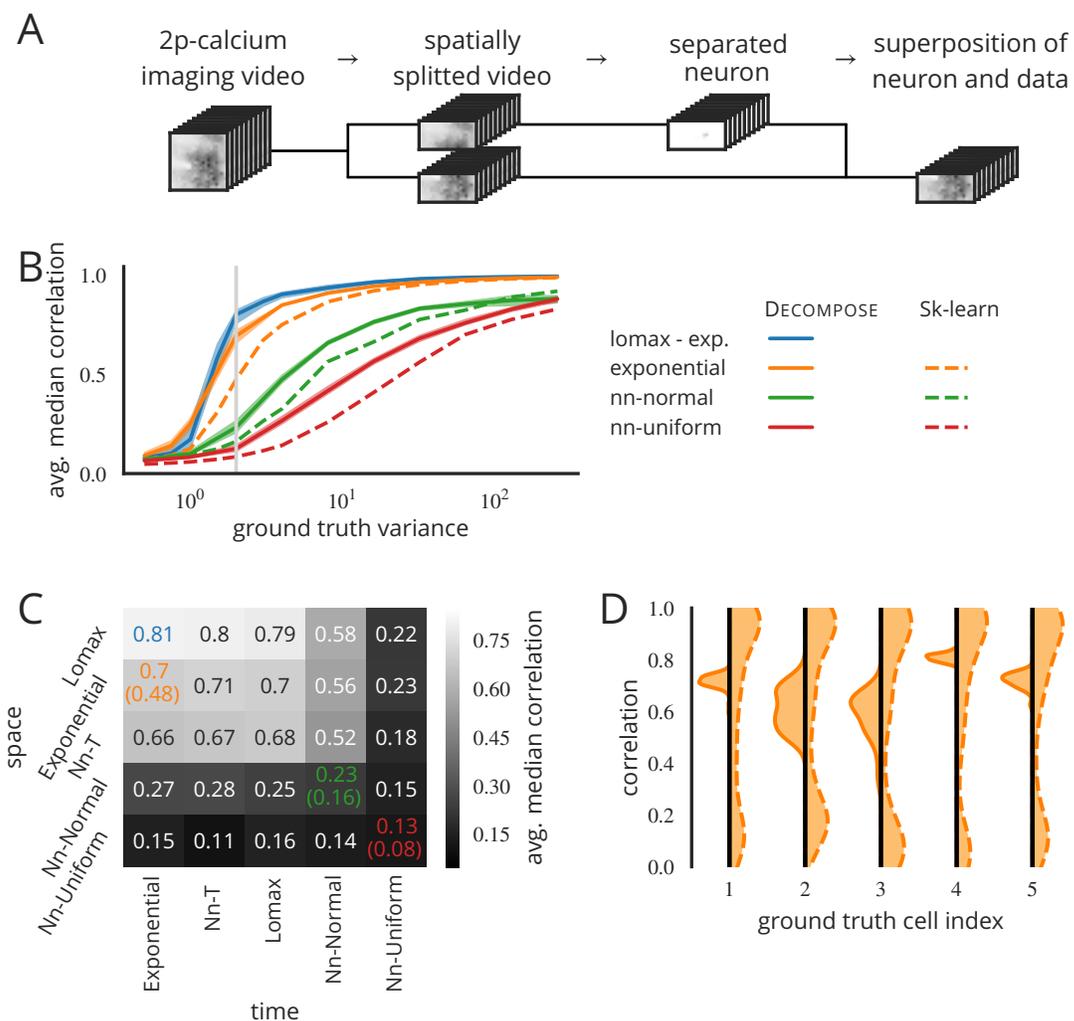}
    \caption{\name{} provides reliable and accurate source recovery for realistic data sets. 
{\bf A}~Procedure to generate a realistic experimental 2p calcium recording with a ground-truth cell. In a first step the field of view is split into two halves. On one half a neuron is isolated and extracted and which is then superimposed on the second half of the video with varying contrast.
{\bf B}~\name{} (solid) outperforms sk-learn (dashed) in recovering the ground-truth source over a wide range of signal-to-noise ratios (x-axis) and for different types of priors (different colours, see legend).
{\bf C}~Source recovery (median correlation with ground-truth source) for different prior combinations. Rows refer to the spatial prior, columns refer to temporal priors.
{\bf D}~Robustness of the source recovery of sparse NMF (sk-learn, dashed) and probabilistic sparse NMF (\name{}, solid). Plotted is the distribution of the source recovery for the five different sources (columns) over 50 different runs.}
    \label{fig:srcRec}
\end{figure}

\topic{Quantification with injected cells}
How can we quantitatively compare different priors and different algorithms on real data for which no simple ground-truth exists? We here solve this problem by injecting single cells as "ground-truth" into the data set as follows (see \FIG{srcRec}A): we spatially split the video into two parts, then we extract one hand-selected cell from one part and place it into the other (see details in the \nameref{sec:data} section). This maintains much of the correlated response statistics between the ground-truth cells and all other cells as well as the background. We control the signal-to-noise ratio, and thus the difficulty of the task, by varying the contrast of the injected cell (see details in the \nameref{sec:data} section). The final quality measure is the maximum correlation that any extracted source has with the injected ground-truth cell.

In a first experiment (\FIG{srcRec}B) we test four different prior combinations. The first three correspond to NMF (red), l2-regularised NMF (green) and double-sparse NMF (orange). All three algorithms are available as probabilistic implementations in \name{} and as non-probabilistic implementations in the sk-learn package. In \name{} the three algorithms are distinguished by their prior assumptions on the spatial and temporal filters. For instance, NMF assumes that the elements of both filters follow a non-negative uniform distribution. In other words, for NMF all values are equally likely as long as the elements are non-negative. Double-sparse NMF, on the other hand, assumes that the elements follow an exponential distribution. That means higher values are assumed to be less likely than smaller values close to zero thus encouraging sparsity. The fourth algorithm tested in \FIG{srcRec}B (blue) does not correspond to any classic BSS algorithm and assumes a Lomax distribution on the spatial filter and a non-negative Student-T distribution on the temporal filter. Both distributions are more heavy-tailed than the standard exponential distribution and thus encourage sparsity even more strongly. The difference between Student-T and Lomax distribution is that the latter more strongly encourages values to be exactly zero.

Every combination of priors is evaluated for different signal-to-noise ratios (\FIG{srcRec}B). For very high signal-to-noise ratios all methods extract one source that basically matches the ground-truth cell (correlation > 0.8). Note that this is only true for the prior combinations evaluated here; other BSS algorithms like ICA or PCA fail to extract the ground-truth cell even for high signal-to-noise ratios (not shown). The difference between the chosen prior combinations is exposed for more challenging signal-to-noise ratios (SNRs). In particular the reconstruction quality of NMF and l2-regularised NMF decays quickly for lower SNR. Within the set of conventional BSS algorithms double sparse NMF performs best at extracting the correct ground-truth cell even for relatively low SNR, but it is still surpassed by the novel combination of Lomax and Student-T priors only available within \name{}.

In a second experiment we compare the probabilistic implementation of NMF, double-sparse NMF and l2-regularised NMF in \name{} against the standard and widely used deterministic implementation in sk-learn (see \FIG{srcRec}B, solid vs dashed). All hyperparameters of sk-learn were intensively cross-validated and the results reported here reflect the maximum performance that can realistically be achieved on a real-world data set. Across the whole range of signal-to-noise ratios and across all three algorithms we consistently find that \name{} outperforms their counterparts in sk-learn by a sizeable margin.

In a third experiment we fix the signal-to-noise ratio (grey vertical bar in \FIG{srcRec}B) and test a range of different prior combinations (\FIG{srcRec}C). Here the rows refer to the prior on the spatial filters of the sources whereas the columns refer to their temporal responses. Overall the results highlight how important the choice of a good prior distribution is. All priors combinations that allow for negative values are basically unable to reconstruct the ground-truth cell. Among non-negative prior combinations the reconstruction performance gradually increases the more sparsity-inducing (i.e. heavy-tailed) the distributions are. That's not surprising given that the desired ground-truth cell is both sparse in space as well as in time.

We finally focus on a single prior combination (both exponential) and test the variability of the estimates. To this end we run sparse PCA from sk-learn and \name{} 50 times and record the correlation with the ground-truth cell for each run. The distribution of correlations is plotted in \FIG{srcRec}D for each of the five ground-truth cells. We observe highly variable results for the non-probabilistic implementation of sparse PCA ranging from nearly optimal to no reconstruction. The results of the probabilistic implementation is much more stable and the correlation with the ground-truth cells is tightly distributed around the mean performance.
 
In summary, the probabilistic algorithm in \name{} performs source separation more accurately and more reliably than existing state-of-the-art implementations and frees the user from extensive cross-validation.

\subsection{Blind source separation with automatic sparsity estimation on real data}\label{sec:realdata}

\begin{figure}
    \centering
    \input{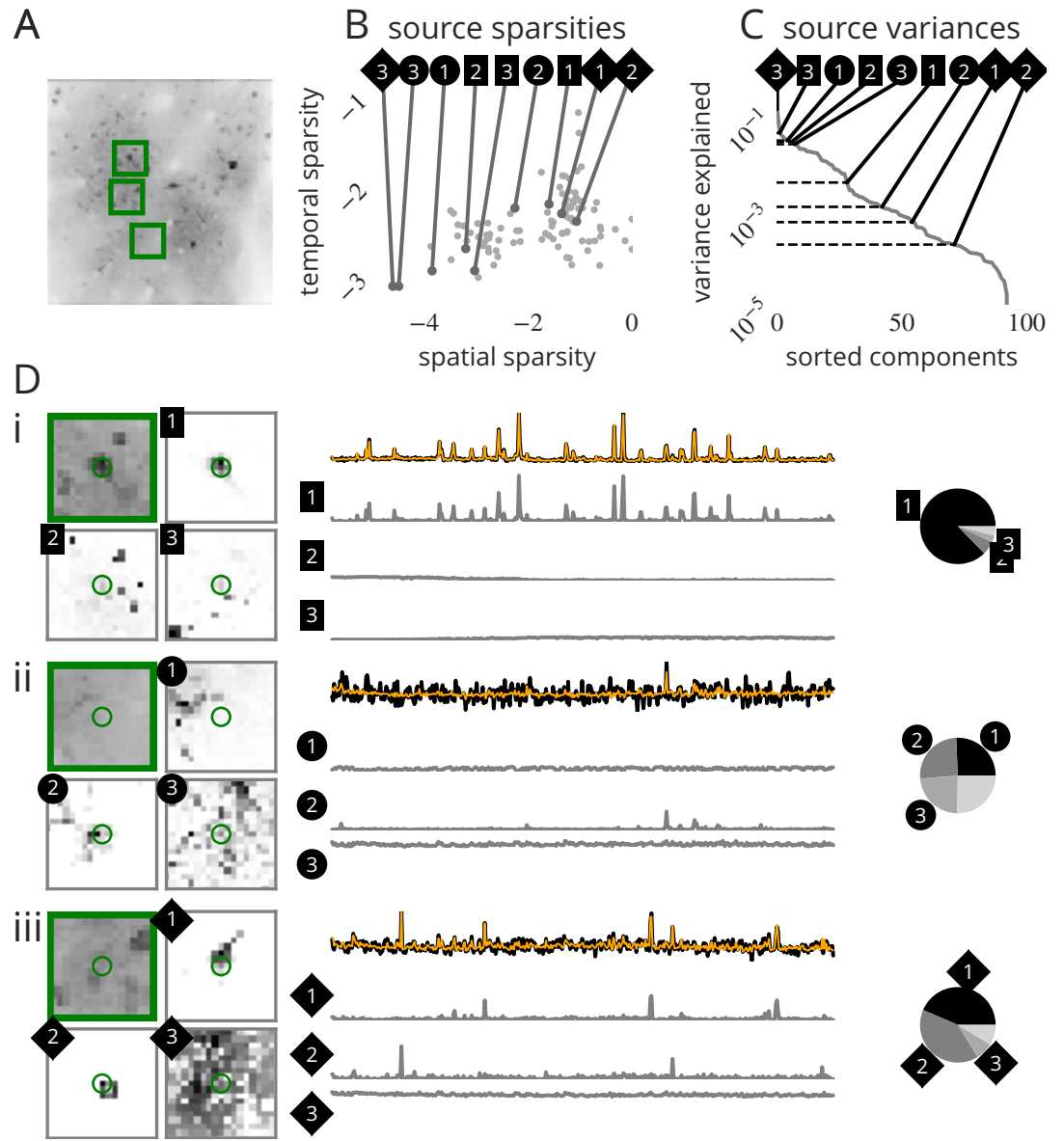}
    \caption{Decomposition of a two-photon calcium imaging recording in the layer 2/3 motor cortex from a behaving mouse with automatic sparsity estimation. 
{\bf A} Temporal average of the calcium activity shown for the complete field of view. Three patches (highlighted in green) are analysed in panel D.
{\bf B} Distribution of temporal and spatial sparsity (measured in terms of the log-ratio between the $L_1$ and the $L_2$ norm) over sources with highlights for the sources shown in panel D. Overall there exists a correlation between temporal and spatial sparsity.
{\bf C} Distribution of the variance of the extracted sources. Values cover around four orders of magnitude. The variances of the extracted sources covered a wide range, almost 4 orders of magnitude. The largest variance is absorbed by background tissue sources (top left), due to their spatial extent. Localised, neural sources contribute much lower variance (2-3 orders of magnitude), but are nonetheless recovered reliably (see panel D).
{\bf D} Decomposition of single pixels in the recording (left column, green circle in top left patch) into contributions from different sources (left column, patches 1-3). The centre columns show the temporal filters for the three most contributing sources with the sum of the individual responses at the top (yellow line). The amount of variance explained by each source is displayed on the right. Three scenarios are shown: (i) a pixel dominated by one source, (ii) a pixel with several weak sources and (iii) a pixel at the boundary of two cells.}
    \label{fig:exdec}
\end{figure}

\topic{Introduce experiment i.e. the model and the data}
Next we perform a complete source separation for a real data set without known ground-truth. The data set is the same as in the previous experiment but now we apply the analysis on the full imaging window spanning \SI[product-units = repeat]{472 x 502}{\micro \metre} (\FIG{exdec}A) and containing approximately 70 neurons in the selected imaging plane (manually counted). Here we perform blind source separation using \name{} with 100 sources and with a Lomax prior for the spatial domain and Exponential for the temporal domain.

\begin{figure}
    \centering
    \input{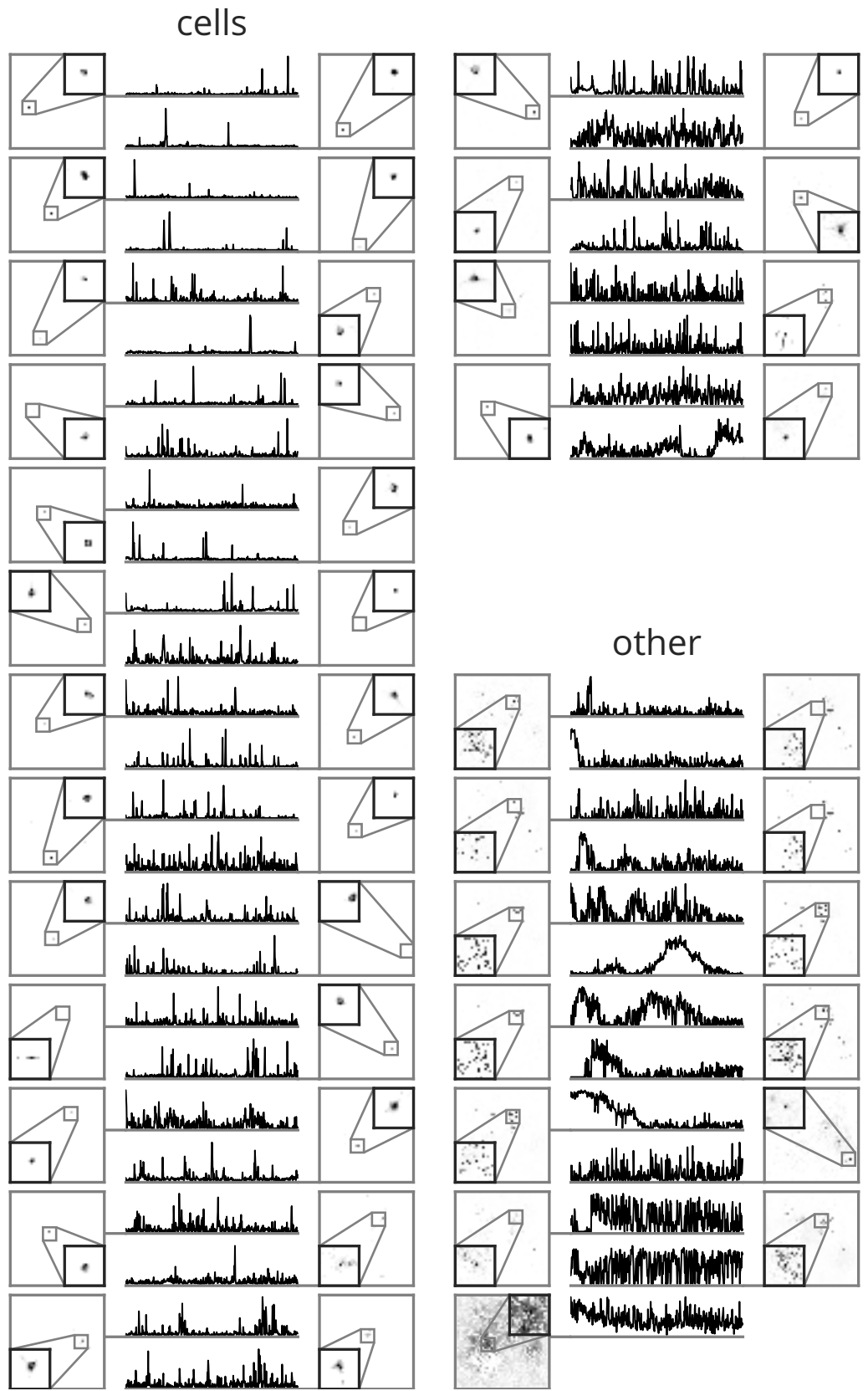}
    \caption{Temporal and spatial filters of an unbiased subset of all estimated sources. We subsampled the sources by ordering them according to their kurtosis and by selecting every second cell.}
    \label{fig:exdecall}
\end{figure}

\topic{We discuss the decomposition at three example locations.}
\FIG{exdecall} shows an unbiased selection of the extracted sources. Sources vary widely in variance (\FIG{exdec}B) and sparsity (\FIG{exdec}C), ranging from dense components that model the background (\FIG{exdecall} bottom right) to many single cells (\FIG{exdecall} left column). Background and single-cells are easily distinguished by their kurtosis. There is a correlation between the spatial and temporal sparsity, i.e. sources that are sparse in space (like cells) tend to be sparse in time (\FIG{exdec}C).

To understand the separation of sources in more detail it is informative to look at how the luminance changes in a single pixel of the video are explained by the individual sources. We look at three cases (\FIG{exdec}A, green squares): (1) a pixel that is dominated by a single cell (\FIG{exdec}D (i)), (2) a pixel that is dominated by the background (\FIG{exdec}D (ii)) and (3) a pixel that sits right at the boundary of two cells (\FIG{exdec}D (iii)).

In the first case a pixel is dominated by a single component (\FIG{exdec}D (i), source 1) that explains more than 70\% of the luminance variance of that pixel (\FIG{exdec}D (i) pie chart). This source clearly exposes a single cell as is visible from its sharply localised spatial distribution as well as its sparse temporal response. Most of the residual variance is explained by non-localised sources that model partially visible cells and potentially dendrites or axions that are vertically aligned to the image plane.

In the second case there is no clear structure that would be identified by the human eye as a meaningful source (\FIG{exdec}D (ii)). Indeed the first and third source (with respect to how much luminance variance they explain) are spatially and temporally dense. Hidden between those two, however, is a source with a localized spatial distribution that resembles the shape of a small soma with its dendritic tree to one side.

Finally, in the last case the pixel sits right between two cells which is again difficult to identify by the human eye (\FIG{exdec}D (ii)). \name{}, however, extracts two cells with a spatial profile that touch at this point but have a very different temporal response profile (\FIG{exdec}D (ii) center column).

\section{Discussion}
\topic{A very short motivation.}
In the context of larger and more complex data sets, blind source separation (BSS) algorithms are essential tools to automate data pre-processing. We here implemented and analysed \name{}, the most comprehensive and scalable open-source framework for probabilistic BSS. Departing from the common non-probabilistic approach, \name{}'s probabilistic formulation gives it five key benefits: (1) ease of use, (2) high flexibility to match the desired statistical properties across many application domains, (3) increased accuracy, (4) increased reliability and, in many cases, (5) reduced overall runtime. Furthermore, we took care to design \name{} both on the algorithmic as well as on the implementation level in a way that makes it (6) scale to massive data sets and (7) easy to extend with new prior distributions. Below we elaborate on these advantages in more detail.

\paragraph{Ease of use}
The probabilistic formulation allows \name{} to estimate optimal parameters of the priors (hyperparameters) alongside the sources in a single run (see \FIG{hypest}). As a result users do not need to manually tune or cross-validate any hyperparameters other than the number of sources.

\paragraph{High flexibility}
\name{} is not restricted to neuroscientific data but is flexible enough to be applied to data sets from many domains. We implemented a wide range of prior distributions that can be flexibly combined to match the statistical assumptions about the data. Most commonly used BSS algorithm like PCA, ICA, NMF, sparse NMF or sparse PCA are part of our framework. The scope of our framework thus encompasses and overreaches all areas in which BSS algorithms have been used in the past. Beyond neuroscience (including fMRI \cite{xie2017} and neurophysiological recordings \cite{lambert2017}) this includes geophysics (\eg extraction of volcanic tremor from seismic data, \cite{Carniel2014}), astronomy (\eg removing the milky way from imaging data recorded with the Green Bank Telescope \cite{wolz2017}), finance (\eg monitoring co-movement of equity prices \cite{raschid2017}) or environmental sciences (\eg source apportionment of pollution in groundwater source area, \cite{guo2017}) to name just a few.

\paragraph{Increased accuracy}
Optimising the hyperparameters directly enables an individualised fit of the priors to each source (\FIG{hypest}). Fitting priors to each source is important because the data typically consists of different sources with varying variance or varying sparsity (\eg dense background versus sparse neuron, see \FIG{exdec}). Non-probabilistic methods typically make the same prior assumption for all sources because cross-validating even a small number of hyperparameters quickly becomes computationally unfeasible.

\paragraph{Increased reliability}
The individualised hyperparameter fit enables probabilistic methods to better separate weak signals from the background (\FIG{hypest}). Overall, \name{} therefore obtains more stable and reliable reconstruction of single cells compared to previous methods (\FIG{srcRec}).

\paragraph{Reduced total runtime}
The total runtime of a BSS estimation is equal to the time required for parameter estimation times the number of crossvalidation runs. Non-probabilistic algorithms typically estimate parameters much faster (\eg ~100-1000$\times$ for the sk-learn algorithms) than their probabilistic counterparts. However, in contrast to probabilistic methods, non-probabilistic algorithms need many cross-validation runs to estimate the hyperparameters. With zero or one hyperparameter the number of runs for cross-validation will be less than 100, thus giving an edge to non-probabilistic implementations. With each additional parameter the number of runs easily grows by a factor of 10 giving an edge to probabilistic algorithms. Hence, in particular for realistically large and heterogeneous data sets the present implementation will provide substantial advantages in run time.

\paragraph{Scaling to massive data}
The present implementation is based on TensorFlow (\cite{abadi2016}) which allows \name{} to automatically take advantage of specialised accelerator hardware like GPUs. \name{} can optionally utilise a low rank approximation of the data to save memory space and computational time (see \nameref{sec:methods} for details). This feature is especially useful for large data sets that contain only a small number of sources.

\paragraph{Extensibility}
New prior distributions can be easily added as long as the maximum likelihood estimates of its hyperparameters can be calculated and the product of the prior density and a normal density is tractable (\ie it is possible to sample and calculate the mode of the conditional posterior distribution efficiently).

\paragraph{Comparison with previous frameworks}
There is no framework that provides a similarly broad spectrum of models as \name{}. The most notable framework used in practice is scikit-learn \citep{pedregosa2011} which provides non-probabilistic implementations for many common models like PCA, SPCA, ICA, NMF, and SNMF. A similar framework with a focus on sparse methods is the Sparse modelling software (SPAMS) (\cite{mairal2011}). Other frameworks focus on a subset of models with certain prior characteristics like non-negativity \citep{li2013,cichocki2006,zhang2017} or exponential distributions (ICA) \citep{rachakonda2008,makeig2000}.

All frameworks mentioned above implement non-probabilistic algorithms and thus do not provide automatic hyperparameter optimisation. They all run exclusively on CPUs and cannot take advantage of specialised accelerator hardware like GPUs. In addition, the number of possible prior combinations available in \name{} goes far beyond any existing framework.

\paragraph{Future work}
\label{sec:future}
An important avenue for future work is to extend the framework with more prior or noise distributions. A very useful extension would be prior distributions that model the smooth changes of a source between two consecutive time steps. In this way, temporal pre-filtering could be directly integrated into the framework, and thus also benefit from automatic hyperparameter estimation.
It would also be useful to generalise our assumptions about the distribution of the residuals. So far we assume that the residuals adhere to a Gaussian distribution, but quite frequently they contain a few entries with very high magnitude. In such cases more heavy tailed distributions like t distributions would provide a better match.

Taken together, in this work we demonstrated the various benefits of probabilistic blind source separation algorithms for automatic source discovery and introduced a comprehensive framework that allows practitioners to use them. 
We hope that the release of \name{} as an open-source python package at \git{} will spur their wide adoption in tomorrow's data analysis pipelines.

\section{Methods and Materials} \label{sec:methods}
\paragraph{BSS as regularised matrix factorisation}
Blind source separation (BSS) can be described as problem of matrix factorisation. Matrix factorisations approximate a data matrix $\mathbf{X} \in \mathbb{R}^{M\times N}$ as a low rank matrix $\mathbf{\hat{X}} \in \mathbb{R}^{M\times N}$ which factorises into $\mathbf{U}\mathbf{V}^\top$ with $\mathbf{U} \in \mathbb{R}^{M\times K}, \mathbf{V} \in \mathbb{R}^{N\times K}$ and $K < N,M$. $\mathbf{\hat{X}}$ is often referred to as the signal, while  $\mathbf{U}$ and  $\mathbf{V}$ are called factors or filter banks. The factors $\mathbf{U}$ and $\mathbf{V}$ may be forced to satisfy certain constraints (e.g. sparsity). Most algorithms used in practice define the solution to the BSS problem as the minimum of the following cost function:
\begin{equation}
  \label{eq:matrixfactorization}
  \mathbf{U}^*, \mathbf{V}^* = \argmin_{\mathbf{U}, \mathbf{V}} ||\mathbf{X} -\mathbf{U}\mathbf{V}^\top||_2^2 + r_\theta(\mathbf{U}, \mathbf{V})\text{.} 
\end{equation}
The cost function consists of two parts: a squared $L2$ loss that enforces that the approximation is close to the data and an optional regularisation penalty $r_\theta(\mathbf{U}, \mathbf{V})$ that enforces the constraints on the factors.

\topic{Hyperparameters cannot be included in the optimisation using a non probabilistic approach}
The regularisation function $r_\theta$ depends on hyperparameters $\theta$. Consider a sparsity-inducing $l1$ penalty as a concrete example: in this case the term $r_\theta$ has the form $\theta |\mathbf{U}|_1$. Optimising \EQ{matrixfactorization} for $\theta$ would in this case yield the trivial solution $\theta = 0$. As a result those hyperparameters cannot be optimised simultaneously with the parameters $\mathbf{U}$, $\mathbf{V}$ but have to be handpicked or cross-validated. Neither of the two options remains practical as the size of the data increases and the number of regularisation parameters becomes larger. To circumvent this problem we now formulate \EQ{matrixfactorization} from a probabilistic perspective.

\paragraph{Probabilistic matrix factorisation}
\topic{Definition of the probabilistic matrix factorisation model}
In the probabilistic matrix factorisation model \citep{mnih2008} we assume that the data $\mathbf{X}$ is a superposition of the signal $\mathbf{\hat{X}}$ and independent Gaussian noise (with precision $\alpha$) and that the signal $\mathbf{\hat{X}}$ factorises into $\mathbf{U}\mathbf{V}^\top$. These assumptions allow us to compute the likelihood of the data given the factors and the precision as
\begin{equation}
\begin{split}
\label{eq:noisedist}
p(\mathbf{X}|\mathbf{U},\mathbf{V}, \alpha) &= \prod_{m,n}\mathcal{N}(X_{m,n}|(\mathbf{\hat{X}})_{m,n}, \alpha^{-1})\\
&=\prod_{m,n}\mathcal{N}(X_{m,n}|(\mathbf{U}\mathbf{V}^\top)_{m,n}, \alpha^{-1})\text{.}
\end{split}
\end{equation}
To model the constraints on the factors we assume that the elements of $\mathbf{U}, \mathbf{V}$ are drawn from some prior distributions,
\begin{equation}
\label{eq:priors}
\begin{split}
p(\mathbf{U}|\theta_U) &= \prod_{m,k} f(u_m^k | \theta_{U_k})\text{,}\\
p(\mathbf{V}|\theta_V) &= \prod_{n,k} g(v_n^k | \theta_{V_k})\text{.}
\end{split}
\end{equation}

\paragraph{Parameter inference}
\topic{Motivation for the probabilistic formulation.}
The goal of probabilistic matrix factorisation is to determine a distribution of factors $\mathbf{U}, \mathbf{V}$ and hyperparameters $\Theta$ (\ie the parameters of the prior distributions $\theta_{U_k}$ and $\theta_{U_k}$ and the noise precision $\alpha$) that maximise the likelihood of the data. More precisely, we are going to minimise the negative log-probability of the data,
\begin{equation}
\label{eq:neglogprob}
\begin{aligned}
\mathbf{\Theta}^*, \mathbf{U}^*, \mathbf{V}^* &= \argmin_{\mathbf{\Theta}, \mathbf{U}, \mathbf{
V}}- \log p(\mathbf{X}, \mathbf{U}, \mathbf{V}, \mathbf{\Theta}) \\
\text{with }\mathbf{\Theta} &= \alpha, \theta_{U}, \theta_{V}\text{.} \\
\end{aligned}
\end{equation}

\topic{The probabilistic and deterministic formulations are similar.}
Note that the cost function derived from the probabilistic formulation given in \EQ{neglogprob} is very similar to the cost function of the non-probabilistic formulation used in \EQ{matrixfactorization}. The links become explicit when we expand the negative log-probability:
\begin{equation}
\label{eq:neglogprobexpanded}
\begin{aligned}
-\log p(\mathbf{X}, \mathbf{U}, \mathbf{V},\mathbf{\Theta}) =& -\log p(\mathbf{X}| \mathbf{U}, \mathbf{V}, \mathbf{\Theta}) - \log p(\mathbf{U}, \mathbf{V}|\mathbf{\Theta}) - \log p(\mathbf{\Theta})\\
=&\frac{\alpha}{2}||\mathbf{X} -\mathbf{U}\mathbf{V}^\top||_2^2+ \tilde{r}_\theta(\mathbf{U}, \mathbf{V}) + c\\
\text{with } \tilde{r}_\theta =& -\log(p(\mathbf{U}|\theta_U)) -\log(p(\mathbf{V}|\theta_V))\text{,}\\
c =& \log\left(\sqrt{2\pi\alpha^{-1}}\right) + \log(p(\mathbf{\Theta}))\text{.}
\end{aligned}
\end{equation}

\topic{Deterministic optimisation of the likelihood is still performing poorly.}
Comparing \EQ{matrixfactorization} with \EQ{neglogprobexpanded} we note several additional terms which come from the normalisation factors of the prior distributions. In principle we could directly optimise \EQ{neglogprobexpanded} in a non-probabilistic manner by finding the factors and hyperparameters with the maximum likelihood. However, empirical evidence suggests that such an optimisation gets trapped very quickly in poor local minima \citep{szeliski2008}.

\topic{Here we apply a combination of EM and BCD to avoid the problems above.}
We can circumvent this problem by optimising \EQ{neglogprobexpanded} in a fully probabilistic way. The crucial difference is that instead of estimating only single points (the maximum likelihood solutions, or MAP estimates) we now infer probability distributions over $\mathbf{U}, \mathbf{V}, \mathbf{\Theta}$ which encodes the uncertainty or confidence in our estimates. To this end we split the optimisation of \EQ{neglogprob} into two subproblems. In the first subproblem we utilise an expectation maximisation (EM) algorithm \citep{dempster1977} to optimise the log probability with respect to the hyperparameters $\Theta$ and an estimate of the distribution over the parameters $p(\mathbf{U}, \mathbf{V}|\mathbf{\Theta})$. In the second subproblem we utilise a block coordinate descent (BCD) algorithm \citep{bertsekas1999} to find the MAP estimates of the parameters that maximise $p(\mathbf{U}, \mathbf{V}|\mathbf{\Theta})$ for the optimal hyperparameters $\mathbf{\Theta}^{*}$ determined in the first subproblem. We choose this approach because EM and BCD algorithms are robust, entail simple algorithms, and have no hyperparameters that need to be adjusted. We describe the details of both steps in the next paragraphs. An overview of the complete algorithm is shown in \ALG{tefa}.

\topic{EM algorithm introduction.}
\paragraph{Hyperparameter inference (EM algorithm)}
In the first subproblem we solve the following optimisation problem:
\begin{equation}
\Theta^* = \argmax_{\Theta} \log p(\mathbf{X} | \Theta) \label{eq:em}\text{.}
\end{equation}
This optimisation is intractable because of the complicated dependency of the observed data $\mathbf{X}$ on the hyperparameters $\Theta$. The dependencies are much simpler if we take into account the factors $\mathbf{Z} = \mathbf{U}, \mathbf{V}$ and rewrite \EQ{em} as
\begin{align*}
\log p(\mathbf{X}|\Theta)&=\log \int p(\mathbf{X}, \mathbf{Z}|\Theta)d\mathbf{Z}\\
&=\log \int p(\mathbf{X}| \mathbf{Z}, \Theta)p(\mathbf{Z}|\Theta)d\mathbf{Z}\text{.}\\
\intertext{
We know both $p(\mathbf{X}|\alpha, \mathbf{Z})$ and , $p(\mathbf{Z}|\theta)$ in closed-form. To make the integral tractable we rely on a trick from importance sampling: most values of $\mathbf{Z}$ yield zero likelihood $p(\mathbf{X}|\alpha, \mathbf{Z})$ and so we do not need to integrate over that part of the space. Instead we estimate a distribution $\tilde q(\mathbf{Z})$ over the "relevant" part of the $\mathbf{Z}$ space and turn the integral into an expectation value,}
&=\log \int \frac{\tilde{p}(\mathbf{Z})}{\tilde{p}(\mathbf{Z})} p(\mathbf{X}| \mathbf{Z}, \Theta)p(\mathbf{Z}|\Theta)d\mathbf{Z}\\
&=\log E_{\tilde{p}}\left[\frac{p(\mathbf{X}|\mathbf{Z},\Theta) p(\mathbf{Z}|\Theta)}{\tilde{p}(\mathbf{Z})}\right]\text{.}\\
\intertext{Using Jensen's inequality we can derive a lower bound,}
&\geq E_{\tilde{p}}\left[\log \frac{p(\mathbf{X}|\mathbf{Z},\Theta) p(\mathbf{Z}|\Theta)}{\tilde{p}(\mathbf{Z})}\right]\\
&=E_{\tilde{p}}\left[\log p(\mathbf{X},\mathbf{Z}|\Theta)\right] + H[\tilde{p}]\\
&=-D_{KL}\left[ \tilde{p}(\mathbf{Z}) || p(\mathbf{Z}|\mathbf{X}, \Theta) \right] + \log p(\mathbf{X}|\Theta).\\
\intertext{The expectation value can be efficiently estimated by sampling from $\tilde q(\mathbf{Z})$. From the last line we observe that the lower bound gets tight whenever $\tilde q(\mathbf{Z})$ is equal to $p(\mathbf{Z}|\mathbf{X}, \Theta)$. Hence, maximising the lower bound with respect to $\tilde q(\mathbf{Z})$ will ensure that we approximate the true posterior. Due to the importance of this lower bound we identify it by the function $F$.}
&=F(\Theta, \tilde{p})
\end{align*}

Every local (global) maximum $\Theta^*$, $\tilde{p}^*$ of $F$ is also a local (global) maximum $\Theta^*$ of equation \EQ{em} \citep{neal1998}. We can find a local maximum $\tilde{p}^*$, $\Theta^*$ of $F$ using block coordinate descent with the following block updates:
\begin{itemize}
\item Set $\tilde{p}^{(t)} = \argmax_{\tilde{p}}F(\Theta^{(t-1)}, \tilde{p}) = p(\mathbf{Z}|\mathbf{X},\Theta^{(t-1)})$\\(E-step)
\item Set $\Theta^{(t)} = \argmax_{\Theta} F(\Theta, \tilde{p}^{(t)})  =\argmax_{\Theta} E_{\tilde{p}}\left[\log p(\mathbf{X},\mathbf{Z}|\Theta)\right]$\\(M-step)
\end{itemize}

\topic{The updates cannot be done optimally.}
The $\Theta$ and $\tilde{p}$ updates are equivalent to the E-, and M-steps of the EM algorithm \citep{neal1998}. Computing those updates is computational very expensive for our model. Therefore we do not conduct them optimally as stated above but only approximately by partial steps in the direction of the optimal solution.

\topic{Approximation of the E-stop.}
In the E-step the distribution $p(\mathbf{Z}| \mathbf{X}, \Theta^{(t-1)})$ is not tractable and we approximate the distribution with samples from a Markov chain using a blocked Gibbs sampler. Due to efficiency we do not drive the chain until it reached equilibrium state but stop very early after just one sample and use this sample to warm start the chain in the next coordinate descent iteration. Nonetheless, it can be shown that posterior approximations based on single samples decrease the value of $F$ \citep{neal1998}. We define the blocks of the blocked Gibbs sampler as the columns of $\mathbf{U}$ and $\mathbf{V}$ i. e. $P(\mathbf{U}_k | \mathbf{U}_{-k}, \mathbf{V}, \Theta, \mathbf{X})$ which can be sampled efficiently.

\topic{Approximation of the M-stop.}
In the M-step the parameters $\Theta$ are updated with their maximum likelihood estimates which can often be computed exactly using a closed form expression. On the other hand, if the updates can be computed only numerically we do not iterate until convergence but again stop early before convergence and warm start in the next iteration of the coordinate descent algorithm. Cases in which the maximum likelihood estimators cannot be calculated analytically are for example mixture distributions.

\topic{Formalising the second sub problem.}
\paragraph{Parameter inference (BCD algorithm)}
The EM algorithm yields approximations for the hyperparameters $\Theta$ and a sample from the posterior distributions over $\mathbf{U}$ and $\mathbf{V}$. We then determine determine (local) maxima of the conditional posterior densities over $\mathbf{U}$ and $\mathbf{V}$ by optimising
\begin{equation}
\mathbf{U}^*, \mathbf{V}^* = \argmin_{\mathbf{U}, \mathbf{V}} -\log p(\mathbf{U}, \mathbf{V}|\Theta, X) \text{.} \label{eq:map}
\end{equation}

\topic{Optimisation of the second sub problem with BCD.}
This optimisation can be computed efficiently using the block coordinate descent (BCD) algorithm. We use the same blocks that we already used during blocked Gibbs sampling, \ie the columns of $\mathbf{U}$ and $\mathbf{V}$. An optimal block update is given by the mode of the conditional distribution over the elements of the block (instead of a sample that we used during blocked Gibbs sampling). The update, i. e. the mode of the distribution, is often available analytically in closed form and can be calculated efficiently. Only few block updates are needed, since the last sample of the blocked Gibbs sampler serves as a very good initial value for the BCD algorithm.

\IncMargin{1em}
\begin{algorithm}[h]
 \KwIn{data $\mathbf{X}$, number of sources $K$, prior distributions $f^{(1)}, f^{(2)}$, \ldots}
 \KwOut{parameters $\mathbf{U}^{(1)}$, $\mathbf{U}^{(2)}$, \ldots, hyperparameters $\theta_{U^{(1)}}$, $\theta_{U^{(2)}}$, \ldots, $\alpha$}
 random initialise $\mathbf{U}^{(1)}$, $\mathbf{U}^{(2)}$, \ldots, $\theta_{U^{(1)}}$, $\theta_{U^{(2)}}$, \ldots, $\alpha$\;
 \While(\tcp*[f]{Expectation-maximisation iterations\hskip 1em}){not converged}{
   $\alpha \leftarrow \argmax_\alpha p(\alpha, \sim)$\tcp*[r]{M-step (1): ML-estimate\hskip 1em}\label{algorithm:tefa3}
   \ForEach{ factor $\mathbf{U}$ in ($\mathbf{U}^{(1)}$, $\mathbf{U}^{(2)}$, \ldots)}{
     \ForEach{column $\mathbf{U_k}$}{
       $\theta_{U_k} \leftarrow \argmax_{\theta_{U_k}} p(\theta_{U_k}, \sim)$\tcp*[r]{M-step (2): ML-estimates\hskip 1em}\label{algorithm:tefa6}
       $\mathbf{U}_k \sim p(\mathbf{U}_k|\sim)$\tcp*[r]{E-step: conditional samples\hskip 1em}\label{algorithm:tefa7}
     }
     rescale columns\tcp*[r]{for numerical stability\hskip 1em}
   }
 }
 \While(\tcp*[f]{Block coordinate descent iterations\hskip 1em}){not converged}{
   \ForEach{factor $\mathbf{U}$ in ($\mathbf{U}^{(1)}$, $\mathbf{U}^{(2)}$, \ldots)}{
     \ForEach{column $\mathbf{U_k}$}{
       $\mathbf{U}_k \leftarrow \argmax_{\mathbf{U}_k} p(\mathbf{U}_k|\sim)$\tcp*[r]{Block updates: conditional mode\hskip 1em}\label{algorithm:tefa15}
     }
     rescale columns\tcp*[r]{for numerical stability\hskip 1em}
   }
 }
 \caption{Fitting the parameters and hyperparameters of a \name{} model.\label{algorithm:tefa}}
\end{algorithm}

\paragraph{Efficient implementation}
\topic{Calculating conditional posterior distribution is computationally very intensive.}
The most costly step of our method is the computation of the conditional posterior distributions which are needed throughout the algorithm: we sample from them during the blocked Gibbs sampling step when optimising $\Theta$ and we calculate their mode during the BCD algorithm when optimising the columns of $\mathbf{U}$ and $\mathbf{V}$. The naive derivation of those distributions yields for a column $\mathbf{U}_k$:
\begin{equation}
\begin{split}
\underbrace{P(\mathbf{U}_k | \mathbf{U}_{-k}, \mathbf{V}, \Theta, \mathbf{X})}_{\text{posterior}}&=\underbrace{\mathcal{N}\left(\mathbf{U}_k \middle| \bm{\mu}^{(k)}, \sigma^{(k)}\right)}_{\text{likelihood}}\underbrace{f(\mathbf{U}_k | \theta_{U_k})}_{\text{prior}}\\
\text{with } \bm{\mu}^{(k)} &=\frac{\mathbf{\tilde{X}}^{(k)}\mathbf{V}_k}{\mathbf{V}_k^\top \mathbf{V}_k} \text{, } {\sigma^{(k)}} = \frac{1}{\sqrt{\alpha \mathbf{V}_k^\top \mathbf{V}_k}} \text{,} \\
\mathbf{\tilde{X}}^{(k)} &= \mathbf{X}- \mathbf{U}_{-k}\mathbf{V}_{-k}^\top\label{eq:posterior}
\end{split}
\end{equation}
and analogously for the columns of $\mathbf{V}$ (not shown). We use two techniques that reduce the computational work needed to obtain those distributions. Subsequently we only consider the case of updating a column $\mathbf{U}_k$ but the same applies for columns of $\mathbf{V}$ which is not written down explicitly.

\topic{Avoid calculation of residuals.}
First, we apply a technique introduced in \cite{cichocki2009} that circumvents the costly computation of the residuals $\mathbf{\tilde{X}}^{(k)}$. Instead of calculating the residuals for every $k$ we define matrices $\mathbf{A} = \mathbf{X}\mathbf{V}$ and $\mathbf{B}=\mathbf{V}^\top\mathbf{V}$ and use them to update all $K$ columns:

\begin{align}
\bm{\mu}^{(k)}&=\frac{\mathbf{A}_k-\mathbf{U}\mathbf{B}_k + \mathbf{U}_k B_{k,k}}{B_{k,k}}\label{eq:mu}\text{,}\\
\sigma^{(k)}&=\frac{1}{\sqrt{\alpha B_{k,k}}}\label{eq:sigma}\text{.}
\end{align}

\topic{Conduct most calculations in a small random subspace.}
Second, we exploit that the number of sources $K$ is typically much smaller than the dimensionality of the data \ie $K \ll M, N$. Therefore we can project the data to a much smaller space using orthonormal projection matrices without loosing much information and calculate $\bm{\mu}^{(k)}$ and $\sigma^{(k)}$ using less computational time and memory. We obtain projection matrices $\mathbf{Q}^{(U)} \in \mathbb{R}^{M \times M^{(R)}}$ and $\mathbf{Q}^{(V)} \in \mathbb{R}^{N \times N^{(R)}}$ using random projections \citep{halko2011}. The projection matrices provide a low rank approximations of the filter banks $\mathbf{U}\approx \mathbf{Q}^{(U)}\mathbf{U}^{(R)}$. Analogously to the calculations in the full space we define $\mathbf{A}^{(R)} = \mathbf{X}^{(R)}\mathbf{V}^{(R)}$ and $\mathbf{B}^{(R)} = {\mathbf{V}^{(R)}}^\top\mathbf{V}^{(R)}$ where $\mathbf{X}^{(R)} = {\mathbf{Q}^{(U)}}^\top\mathbf{X}\mathbf{Q}^{(V)}$ is the low rank approximation of the data. Hence, $\bm{\mu}^{(k)}$ and $\sigma^{(k)}$ can be approximated efficiently as follows:

\begin{align}
\bm{\mu}^{(k)}&\approx{\mathbf{Q}^{(U)}}^\top\frac{\mathbf{A}^{(R)}_k-\mathbf{U}^{(R)}\mathbf{B}^{(R)}_k + \mathbf{U}^{(R)}_k B^{(R)}_{k,k}}{B^{(R)}_{k,k}}\label{eq:muapprox}\text{,}\\
\sigma^{(k)}&\approx\frac{1}{\sqrt{\alpha B^{(R)}_{k,k}}}\label{eq:sigmasqapprox}\text{.}
\end{align}

Importantly, \EQ{muapprox} and \EQ{sigmasqapprox} do not depend on the data $\mathbf{X}$ anymore but only on its low rank approximation $\mathbf{X}^{(R)}$.

\topic{Explicitly point out the trade-off between speed and accuracy.}
The calculation of the mean and the variance of the likelihood (\EQ{posterior}) can be conducted exactly (\EQ{mu}, \EQ{sigma}) or approximately with random projections (\EQ{muapprox}, \EQ{sigmasqapprox}). The decision for one or the other approach depends on a trade-off between speed and accuracy which in turn depends on the number random projections $M^{(R)}$, $N^{(R)}$, the number of sources $K$ and the dimensionality of the data $M$, $N$ \eg the number of pixels and the number of time steps.

\begin{figure}
    \centering
    \input{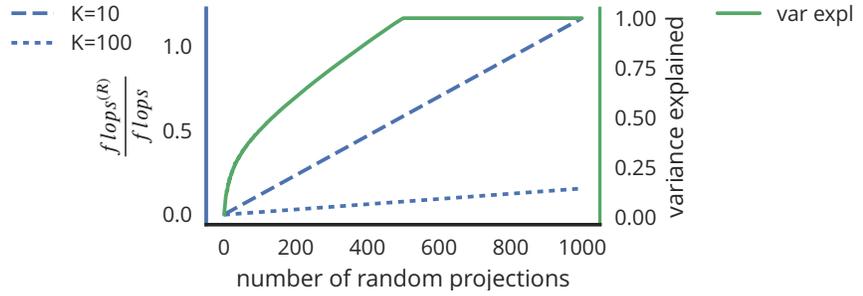}
    \caption{Assessing how dimensionality reduction benefits computational savings (blue) and affects data reconstruction quality (green) on a 2p calcium imaging recording with 128 $\times$ 128 pixels and 500 time steps (see section \nameref{sec:data} for details). We employ random projections in the spatial domain to reduce the dimensionality of the data. With 500 random projections it is possible to recover the full data set while reducing the computational complexity (in terms of floating points operations) by more then 92\% (for $K=100$ sources).
}
    \label{fig:flops}
\end{figure}

\topic{Explain the dependency of the parameters on the speed accuracy trade-off.}
In \FIG{flops} we discuss the trade-off using a 2p Calcium imaging recording with 128 $\times$ 128 pixels and 500 time steps (see section \nameref{sec:data} for details). Here we apply the dimensionality reduction only in the spatial domain. As expected, reducing the number of random projections leads to a decrease in computational demand (blue) and a less accurate reconstruction of the data (green). The benefit increases with the number sources. However, for a small number of sources or a large number of random projections the overhead of the dimensionality reduction step can also offset the benefits.

\topic{Resolve the trad-off in an example using experimental data.}
In the example in \FIG{flops}, for 100 sources a good trade-off is achieved with 500 random projections. In this case the number of floating point operations per iteration is reduced by more than 92\% while still maintaining perfect reconstruction. Even larger savings can be achieved if the dimensionality reduction is applied to all data dimensions.

\paragraph{Prior distributions}

\topic{Our framework has a large selection of prior distributions.}
In principle any probability distribution can be used as a prior distribution, but unfortunately only a subset of them can be implemented efficiently. A large number of tractable prior distributions are already implemented in the framework as listed in table \TABLE{priors}. Others such as truncated distributions, discrete distributions, and smoothness inducing priors can easily be integrated into our framework later.

\topic{Requirements to add new prior distributions.}
There are three requirements to add a new prior. First, there must exist a tractable maximum-likelihood estimator for the hyperparameters of the prior (which is used in line \ref{algorithm:tefa6} of \ALG{tefa}). Such an estimator can be calculated analytically or numerically using an iterative approach as long as the updates are calculated efficiently and increase the likelihood of the data.
The other two requirements concern the posterior distribution induced by the new prior. The density of the posterior distribution is proportional to the product of a normal density (coming from the likelihood) times the prior density. Our algorithm is required to sample efficiently from the posterior (used in line \ref{algorithm:tefa7} of  \ALG{tefa}) and calculate its mode (used in line \ref{algorithm:tefa15} of \ALG{tefa}).

\paragraph{Sampling from conditional posterior distributions}
\topic{Sampling from the conditional posterior can be challenging in DECOMPOSE}
Tensorflow (API r1.5 at the time of writing) provides highly optimised sampling methods for common distributions such as Normal distributions but not for some uncommon ones like non-negative normal. Therefore we ported a truncated normal sampling algorithm described in \cite{chopin2011} to TensorFlow. We implemented a vectorised version that allows us to sample whole filters at once even when every element is drawn from a distribution with different hyperparameters. This implementation allows us to sample efficiently from non-negative normal distributions which are a special subset of truncated normal distributions. The T and Lomax distribution are compound distributions whose conditional posterior distributions (which we need to sample) reduce to tractable Normal and Exponential distributions respectively.

\paragraph{Maximum likelihood estimation of the hyperparameters}
\topic{ML estimates for normal and exponential type priors}
For most distributions like Normal or Laplace priors it is straight-forward to compute the optimal hyperparameter update. One exception are the shifted non-negative versions where no closed form of the optimal parameter updates exist. In those cases we perform an inexact line search along the gradient which is guaranteed to improve the likelihood with every update. 

\topic{ML estimates for T and Lomax type priors}
Another exception are T and Lomax distributions for which no closed-form update exists. We update the parameters of the T distribution according to \cite{liu1995} which utilises the fact that T distributions are compound distributions which allows the application of the EM algorithm. In our framework we adapt the same technique to update the parameters of Lomax distributions. However the technique could not be applied to the shifted non-negative versions of the distributions because calculating the expectation over the latent variables leads to an intractable integral.

\newlength{\priorwidth}
\setlength{\priorwidth}{0.075\textwidth}
\newcolumntype{L}[1]{>{\centering\arraybackslash}m{#1}}%
\begin{table}
\centering
  \begin{tabular}{ L{2.5cm}L{1.1cm}L{1.1cm}L{1.1cm}L{1.1cm}}

  &
  \rot{90}{\begin{minipage}{1.5cm}centered\end{minipage}} &
  \rot{90}{\begin{minipage}{1.5cm}centered\\non-neg\end{minipage}} &
  \rot{90}{\begin{minipage}{1.5cm}shifted\end{minipage}} &
  \rot{90}{\begin{minipage}{1.5cm}shifted\\non-neg\end{minipage}}\tabularnewline
  \midrule

  Uniform &

\begingroup%
\makeatletter%
\begin{pgfpicture}%
\pgfpathrectangle{\pgfpointorigin}{\pgfqpoint{0.393701in}{0.393701in}}%
\pgfusepath{use as bounding box, clip}%
\begin{pgfscope}%
\pgfsetbuttcap%
\pgfsetmiterjoin%
\pgfsetlinewidth{0.000000pt}%
\definecolor{currentstroke}{rgb}{0.000000,0.000000,0.000000}%
\pgfsetstrokecolor{currentstroke}%
\pgfsetstrokeopacity{0.000000}%
\pgfsetdash{}{0pt}%
\pgfpathmoveto{\pgfqpoint{0.000000in}{0.000000in}}%
\pgfpathlineto{\pgfqpoint{0.393701in}{0.000000in}}%
\pgfpathlineto{\pgfqpoint{0.393701in}{0.393701in}}%
\pgfpathlineto{\pgfqpoint{0.000000in}{0.393701in}}%
\pgfpathclose%
\pgfusepath{}%
\end{pgfscope}%
\begin{pgfscope}%
\pgfsetbuttcap%
\pgfsetmiterjoin%
\pgfsetlinewidth{0.000000pt}%
\definecolor{currentstroke}{rgb}{0.000000,0.000000,0.000000}%
\pgfsetstrokecolor{currentstroke}%
\pgfsetstrokeopacity{0.000000}%
\pgfsetdash{}{0pt}%
\pgfpathmoveto{\pgfqpoint{0.000000in}{0.000000in}}%
\pgfpathlineto{\pgfqpoint{0.393701in}{0.000000in}}%
\pgfpathlineto{\pgfqpoint{0.393701in}{0.393701in}}%
\pgfpathlineto{\pgfqpoint{0.000000in}{0.393701in}}%
\pgfpathclose%
\pgfusepath{}%
\end{pgfscope}%
\begin{pgfscope}%
\pgfpathrectangle{\pgfqpoint{0.000000in}{0.000000in}}{\pgfqpoint{0.393701in}{0.393701in}}%
\pgfusepath{clip}%
\pgfsetroundcap%
\pgfsetroundjoin%
\pgfsetlinewidth{1.003750pt}%
\definecolor{currentstroke}{rgb}{0.121569,0.466667,0.705882}%
\pgfsetstrokecolor{currentstroke}%
\pgfsetdash{}{0pt}%
\pgfpathmoveto{\pgfqpoint{0.017896in}{0.196851in}}%
\pgfpathlineto{\pgfqpoint{0.375806in}{0.196851in}}%
\pgfpathlineto{\pgfqpoint{0.375806in}{0.196851in}}%
\pgfusepath{stroke}%
\end{pgfscope}%
\begin{pgfscope}%
\pgfsetrectcap%
\pgfsetmiterjoin%
\pgfsetlinewidth{1.254687pt}%
\definecolor{currentstroke}{rgb}{0.150000,0.150000,0.150000}%
\pgfsetstrokecolor{currentstroke}%
\pgfsetdash{}{0pt}%
\pgfpathmoveto{\pgfqpoint{0.196851in}{0.000000in}}%
\pgfpathlineto{\pgfqpoint{0.196851in}{0.393701in}}%
\pgfusepath{stroke}%
\end{pgfscope}%
\begin{pgfscope}%
\pgfsetrectcap%
\pgfsetmiterjoin%
\pgfsetlinewidth{1.254687pt}%
\definecolor{currentstroke}{rgb}{0.150000,0.150000,0.150000}%
\pgfsetstrokecolor{currentstroke}%
\pgfsetdash{}{0pt}%
\pgfpathmoveto{\pgfqpoint{0.000000in}{0.000000in}}%
\pgfpathlineto{\pgfqpoint{0.393701in}{0.000000in}}%
\pgfusepath{stroke}%
\end{pgfscope}%
\end{pgfpicture}%
\makeatother%
\endgroup%
 &

\begingroup%
\makeatletter%
\begin{pgfpicture}%
\pgfpathrectangle{\pgfpointorigin}{\pgfqpoint{0.393701in}{0.393701in}}%
\pgfusepath{use as bounding box, clip}%
\begin{pgfscope}%
\pgfsetbuttcap%
\pgfsetmiterjoin%
\pgfsetlinewidth{0.000000pt}%
\definecolor{currentstroke}{rgb}{0.000000,0.000000,0.000000}%
\pgfsetstrokecolor{currentstroke}%
\pgfsetstrokeopacity{0.000000}%
\pgfsetdash{}{0pt}%
\pgfpathmoveto{\pgfqpoint{0.000000in}{0.000000in}}%
\pgfpathlineto{\pgfqpoint{0.393701in}{0.000000in}}%
\pgfpathlineto{\pgfqpoint{0.393701in}{0.393701in}}%
\pgfpathlineto{\pgfqpoint{0.000000in}{0.393701in}}%
\pgfpathclose%
\pgfusepath{}%
\end{pgfscope}%
\begin{pgfscope}%
\pgfsetbuttcap%
\pgfsetmiterjoin%
\pgfsetlinewidth{0.000000pt}%
\definecolor{currentstroke}{rgb}{0.000000,0.000000,0.000000}%
\pgfsetstrokecolor{currentstroke}%
\pgfsetstrokeopacity{0.000000}%
\pgfsetdash{}{0pt}%
\pgfpathmoveto{\pgfqpoint{0.000000in}{0.000000in}}%
\pgfpathlineto{\pgfqpoint{0.393701in}{0.000000in}}%
\pgfpathlineto{\pgfqpoint{0.393701in}{0.393701in}}%
\pgfpathlineto{\pgfqpoint{0.000000in}{0.393701in}}%
\pgfpathclose%
\pgfusepath{}%
\end{pgfscope}%
\begin{pgfscope}%
\pgfpathrectangle{\pgfqpoint{0.000000in}{0.000000in}}{\pgfqpoint{0.393701in}{0.393701in}}%
\pgfusepath{clip}%
\pgfsetroundcap%
\pgfsetroundjoin%
\pgfsetlinewidth{1.003750pt}%
\definecolor{currentstroke}{rgb}{0.121569,0.466667,0.705882}%
\pgfsetstrokecolor{currentstroke}%
\pgfsetdash{}{0pt}%
\pgfpathmoveto{\pgfqpoint{0.017896in}{0.196851in}}%
\pgfpathlineto{\pgfqpoint{0.021511in}{0.196851in}}%
\pgfpathlineto{\pgfqpoint{0.025126in}{0.196851in}}%
\pgfpathlineto{\pgfqpoint{0.028741in}{0.196851in}}%
\pgfpathlineto{\pgfqpoint{0.032357in}{0.196851in}}%
\pgfpathlineto{\pgfqpoint{0.035972in}{0.196851in}}%
\pgfpathlineto{\pgfqpoint{0.039587in}{0.196851in}}%
\pgfpathlineto{\pgfqpoint{0.043202in}{0.196851in}}%
\pgfpathlineto{\pgfqpoint{0.046818in}{0.196851in}}%
\pgfpathlineto{\pgfqpoint{0.050433in}{0.196851in}}%
\pgfpathlineto{\pgfqpoint{0.054048in}{0.196851in}}%
\pgfpathlineto{\pgfqpoint{0.057663in}{0.196851in}}%
\pgfpathlineto{\pgfqpoint{0.061279in}{0.196851in}}%
\pgfpathlineto{\pgfqpoint{0.064894in}{0.196851in}}%
\pgfpathlineto{\pgfqpoint{0.068509in}{0.196851in}}%
\pgfpathlineto{\pgfqpoint{0.072124in}{0.196851in}}%
\pgfpathlineto{\pgfqpoint{0.075740in}{0.196851in}}%
\pgfpathlineto{\pgfqpoint{0.079355in}{0.196851in}}%
\pgfpathlineto{\pgfqpoint{0.082970in}{0.196851in}}%
\pgfpathlineto{\pgfqpoint{0.086585in}{0.196851in}}%
\pgfpathlineto{\pgfqpoint{0.090201in}{0.196851in}}%
\pgfpathlineto{\pgfqpoint{0.093816in}{0.196851in}}%
\pgfpathlineto{\pgfqpoint{0.097431in}{0.196851in}}%
\pgfpathlineto{\pgfqpoint{0.101046in}{0.196851in}}%
\pgfpathlineto{\pgfqpoint{0.104662in}{0.196851in}}%
\pgfpathlineto{\pgfqpoint{0.108277in}{0.196851in}}%
\pgfpathlineto{\pgfqpoint{0.111892in}{0.196851in}}%
\pgfpathlineto{\pgfqpoint{0.115507in}{0.196851in}}%
\pgfpathlineto{\pgfqpoint{0.119123in}{0.196851in}}%
\pgfpathlineto{\pgfqpoint{0.122738in}{0.196851in}}%
\pgfpathlineto{\pgfqpoint{0.126353in}{0.196851in}}%
\pgfpathlineto{\pgfqpoint{0.129968in}{0.196851in}}%
\pgfpathlineto{\pgfqpoint{0.133584in}{0.196851in}}%
\pgfpathlineto{\pgfqpoint{0.137199in}{0.196851in}}%
\pgfpathlineto{\pgfqpoint{0.140814in}{0.196851in}}%
\pgfpathlineto{\pgfqpoint{0.144429in}{0.196851in}}%
\pgfpathlineto{\pgfqpoint{0.148045in}{0.196851in}}%
\pgfpathlineto{\pgfqpoint{0.151660in}{0.196851in}}%
\pgfpathlineto{\pgfqpoint{0.155275in}{0.196851in}}%
\pgfpathlineto{\pgfqpoint{0.158890in}{0.196851in}}%
\pgfpathlineto{\pgfqpoint{0.162506in}{0.196851in}}%
\pgfpathlineto{\pgfqpoint{0.166121in}{0.196851in}}%
\pgfpathlineto{\pgfqpoint{0.169736in}{0.196851in}}%
\pgfpathlineto{\pgfqpoint{0.173351in}{0.196851in}}%
\pgfpathlineto{\pgfqpoint{0.176967in}{0.196851in}}%
\pgfpathlineto{\pgfqpoint{0.180582in}{0.196851in}}%
\pgfpathlineto{\pgfqpoint{0.184197in}{0.196851in}}%
\pgfpathlineto{\pgfqpoint{0.187812in}{0.196851in}}%
\pgfpathlineto{\pgfqpoint{0.191428in}{0.196851in}}%
\pgfpathlineto{\pgfqpoint{0.195043in}{0.196851in}}%
\pgfpathlineto{\pgfqpoint{0.198658in}{0.196851in}}%
\pgfpathlineto{\pgfqpoint{0.202273in}{0.196851in}}%
\pgfpathlineto{\pgfqpoint{0.205889in}{0.196851in}}%
\pgfpathlineto{\pgfqpoint{0.209504in}{0.196851in}}%
\pgfpathlineto{\pgfqpoint{0.213119in}{0.196851in}}%
\pgfpathlineto{\pgfqpoint{0.216734in}{0.196851in}}%
\pgfpathlineto{\pgfqpoint{0.220350in}{0.196851in}}%
\pgfpathlineto{\pgfqpoint{0.223965in}{0.196851in}}%
\pgfpathlineto{\pgfqpoint{0.227580in}{0.196851in}}%
\pgfpathlineto{\pgfqpoint{0.231195in}{0.196851in}}%
\pgfpathlineto{\pgfqpoint{0.234811in}{0.196851in}}%
\pgfpathlineto{\pgfqpoint{0.238426in}{0.196851in}}%
\pgfpathlineto{\pgfqpoint{0.242041in}{0.196851in}}%
\pgfpathlineto{\pgfqpoint{0.245656in}{0.196851in}}%
\pgfpathlineto{\pgfqpoint{0.249272in}{0.196851in}}%
\pgfpathlineto{\pgfqpoint{0.252887in}{0.196851in}}%
\pgfpathlineto{\pgfqpoint{0.256502in}{0.196851in}}%
\pgfpathlineto{\pgfqpoint{0.260117in}{0.196851in}}%
\pgfpathlineto{\pgfqpoint{0.263733in}{0.196851in}}%
\pgfpathlineto{\pgfqpoint{0.267348in}{0.196851in}}%
\pgfpathlineto{\pgfqpoint{0.270963in}{0.196851in}}%
\pgfpathlineto{\pgfqpoint{0.274578in}{0.196851in}}%
\pgfpathlineto{\pgfqpoint{0.278194in}{0.196851in}}%
\pgfpathlineto{\pgfqpoint{0.281809in}{0.196851in}}%
\pgfpathlineto{\pgfqpoint{0.285424in}{0.196851in}}%
\pgfpathlineto{\pgfqpoint{0.289039in}{0.196851in}}%
\pgfpathlineto{\pgfqpoint{0.292655in}{0.196851in}}%
\pgfpathlineto{\pgfqpoint{0.296270in}{0.196851in}}%
\pgfpathlineto{\pgfqpoint{0.299885in}{0.196851in}}%
\pgfpathlineto{\pgfqpoint{0.303500in}{0.196851in}}%
\pgfpathlineto{\pgfqpoint{0.307116in}{0.196851in}}%
\pgfpathlineto{\pgfqpoint{0.310731in}{0.196851in}}%
\pgfpathlineto{\pgfqpoint{0.314346in}{0.196851in}}%
\pgfpathlineto{\pgfqpoint{0.317961in}{0.196851in}}%
\pgfpathlineto{\pgfqpoint{0.321577in}{0.196851in}}%
\pgfpathlineto{\pgfqpoint{0.325192in}{0.196851in}}%
\pgfpathlineto{\pgfqpoint{0.328807in}{0.196851in}}%
\pgfpathlineto{\pgfqpoint{0.332422in}{0.196851in}}%
\pgfpathlineto{\pgfqpoint{0.336038in}{0.196851in}}%
\pgfpathlineto{\pgfqpoint{0.339653in}{0.196851in}}%
\pgfpathlineto{\pgfqpoint{0.343268in}{0.196851in}}%
\pgfpathlineto{\pgfqpoint{0.346883in}{0.196851in}}%
\pgfpathlineto{\pgfqpoint{0.350499in}{0.196851in}}%
\pgfpathlineto{\pgfqpoint{0.354114in}{0.196851in}}%
\pgfpathlineto{\pgfqpoint{0.357729in}{0.196851in}}%
\pgfpathlineto{\pgfqpoint{0.361344in}{0.196851in}}%
\pgfpathlineto{\pgfqpoint{0.364960in}{0.196851in}}%
\pgfpathlineto{\pgfqpoint{0.368575in}{0.196851in}}%
\pgfpathlineto{\pgfqpoint{0.372190in}{0.196851in}}%
\pgfpathlineto{\pgfqpoint{0.375805in}{0.196851in}}%
\pgfusepath{stroke}%
\end{pgfscope}%
\begin{pgfscope}%
\pgfsetrectcap%
\pgfsetmiterjoin%
\pgfsetlinewidth{1.254687pt}%
\definecolor{currentstroke}{rgb}{0.150000,0.150000,0.150000}%
\pgfsetstrokecolor{currentstroke}%
\pgfsetdash{}{0pt}%
\pgfpathmoveto{\pgfqpoint{0.000000in}{0.000000in}}%
\pgfpathlineto{\pgfqpoint{0.000000in}{0.393701in}}%
\pgfusepath{stroke}%
\end{pgfscope}%
\begin{pgfscope}%
\pgfsetrectcap%
\pgfsetmiterjoin%
\pgfsetlinewidth{1.254687pt}%
\definecolor{currentstroke}{rgb}{0.150000,0.150000,0.150000}%
\pgfsetstrokecolor{currentstroke}%
\pgfsetdash{}{0pt}%
\pgfpathmoveto{\pgfqpoint{0.000000in}{0.000000in}}%
\pgfpathlineto{\pgfqpoint{0.393701in}{0.000000in}}%
\pgfusepath{stroke}%
\end{pgfscope}%
\end{pgfpicture}%
\makeatother%
\endgroup%
 &
  -&
  -\tabularnewline
  \midrule

  Normal &

\begingroup%
\makeatletter%
\begin{pgfpicture}%
\pgfpathrectangle{\pgfpointorigin}{\pgfqpoint{0.393701in}{0.393701in}}%
\pgfusepath{use as bounding box, clip}%
\begin{pgfscope}%
\pgfsetbuttcap%
\pgfsetmiterjoin%
\pgfsetlinewidth{0.000000pt}%
\definecolor{currentstroke}{rgb}{0.000000,0.000000,0.000000}%
\pgfsetstrokecolor{currentstroke}%
\pgfsetstrokeopacity{0.000000}%
\pgfsetdash{}{0pt}%
\pgfpathmoveto{\pgfqpoint{0.000000in}{0.000000in}}%
\pgfpathlineto{\pgfqpoint{0.393701in}{0.000000in}}%
\pgfpathlineto{\pgfqpoint{0.393701in}{0.393701in}}%
\pgfpathlineto{\pgfqpoint{0.000000in}{0.393701in}}%
\pgfpathclose%
\pgfusepath{}%
\end{pgfscope}%
\begin{pgfscope}%
\pgfsetbuttcap%
\pgfsetmiterjoin%
\pgfsetlinewidth{0.000000pt}%
\definecolor{currentstroke}{rgb}{0.000000,0.000000,0.000000}%
\pgfsetstrokecolor{currentstroke}%
\pgfsetstrokeopacity{0.000000}%
\pgfsetdash{}{0pt}%
\pgfpathmoveto{\pgfqpoint{0.000000in}{0.000000in}}%
\pgfpathlineto{\pgfqpoint{0.393701in}{0.000000in}}%
\pgfpathlineto{\pgfqpoint{0.393701in}{0.393701in}}%
\pgfpathlineto{\pgfqpoint{0.000000in}{0.393701in}}%
\pgfpathclose%
\pgfusepath{}%
\end{pgfscope}%
\begin{pgfscope}%
\pgfpathrectangle{\pgfqpoint{0.000000in}{0.000000in}}{\pgfqpoint{0.393701in}{0.393701in}}%
\pgfusepath{clip}%
\pgfsetroundcap%
\pgfsetroundjoin%
\pgfsetlinewidth{1.003750pt}%
\definecolor{currentstroke}{rgb}{0.121569,0.466667,0.705882}%
\pgfsetstrokecolor{currentstroke}%
\pgfsetdash{}{0pt}%
\pgfpathmoveto{\pgfqpoint{0.017896in}{0.017896in}}%
\pgfpathlineto{\pgfqpoint{0.034082in}{0.022622in}}%
\pgfpathlineto{\pgfqpoint{0.046672in}{0.029092in}}%
\pgfpathlineto{\pgfqpoint{0.059262in}{0.039192in}}%
\pgfpathlineto{\pgfqpoint{0.070053in}{0.051679in}}%
\pgfpathlineto{\pgfqpoint{0.080844in}{0.068505in}}%
\pgfpathlineto{\pgfqpoint{0.093434in}{0.094415in}}%
\pgfpathlineto{\pgfqpoint{0.106024in}{0.127441in}}%
\pgfpathlineto{\pgfqpoint{0.122211in}{0.179337in}}%
\pgfpathlineto{\pgfqpoint{0.168973in}{0.338400in}}%
\pgfpathlineto{\pgfqpoint{0.179764in}{0.361298in}}%
\pgfpathlineto{\pgfqpoint{0.186959in}{0.370904in}}%
\pgfpathlineto{\pgfqpoint{0.194153in}{0.375477in}}%
\pgfpathlineto{\pgfqpoint{0.199548in}{0.375477in}}%
\pgfpathlineto{\pgfqpoint{0.204944in}{0.372530in}}%
\pgfpathlineto{\pgfqpoint{0.212138in}{0.364153in}}%
\pgfpathlineto{\pgfqpoint{0.221131in}{0.347070in}}%
\pgfpathlineto{\pgfqpoint{0.231922in}{0.318393in}}%
\pgfpathlineto{\pgfqpoint{0.248109in}{0.264102in}}%
\pgfpathlineto{\pgfqpoint{0.285878in}{0.132722in}}%
\pgfpathlineto{\pgfqpoint{0.300267in}{0.094415in}}%
\pgfpathlineto{\pgfqpoint{0.312857in}{0.068505in}}%
\pgfpathlineto{\pgfqpoint{0.325446in}{0.049316in}}%
\pgfpathlineto{\pgfqpoint{0.338036in}{0.035865in}}%
\pgfpathlineto{\pgfqpoint{0.350626in}{0.026925in}}%
\pgfpathlineto{\pgfqpoint{0.365014in}{0.020681in}}%
\pgfpathlineto{\pgfqpoint{0.375806in}{0.017896in}}%
\pgfpathlineto{\pgfqpoint{0.375806in}{0.017896in}}%
\pgfusepath{stroke}%
\end{pgfscope}%
\begin{pgfscope}%
\pgfsetrectcap%
\pgfsetmiterjoin%
\pgfsetlinewidth{1.254687pt}%
\definecolor{currentstroke}{rgb}{0.150000,0.150000,0.150000}%
\pgfsetstrokecolor{currentstroke}%
\pgfsetdash{}{0pt}%
\pgfpathmoveto{\pgfqpoint{0.196851in}{0.000000in}}%
\pgfpathlineto{\pgfqpoint{0.196851in}{0.393701in}}%
\pgfusepath{stroke}%
\end{pgfscope}%
\begin{pgfscope}%
\pgfsetrectcap%
\pgfsetmiterjoin%
\pgfsetlinewidth{1.254687pt}%
\definecolor{currentstroke}{rgb}{0.150000,0.150000,0.150000}%
\pgfsetstrokecolor{currentstroke}%
\pgfsetdash{}{0pt}%
\pgfpathmoveto{\pgfqpoint{0.000000in}{0.000000in}}%
\pgfpathlineto{\pgfqpoint{0.393701in}{0.000000in}}%
\pgfusepath{stroke}%
\end{pgfscope}%
\end{pgfpicture}%
\makeatother%
\endgroup%
 &

\begingroup%
\makeatletter%
\begin{pgfpicture}%
\pgfpathrectangle{\pgfpointorigin}{\pgfqpoint{0.393701in}{0.393701in}}%
\pgfusepath{use as bounding box, clip}%
\begin{pgfscope}%
\pgfsetbuttcap%
\pgfsetmiterjoin%
\pgfsetlinewidth{0.000000pt}%
\definecolor{currentstroke}{rgb}{0.000000,0.000000,0.000000}%
\pgfsetstrokecolor{currentstroke}%
\pgfsetstrokeopacity{0.000000}%
\pgfsetdash{}{0pt}%
\pgfpathmoveto{\pgfqpoint{0.000000in}{0.000000in}}%
\pgfpathlineto{\pgfqpoint{0.393701in}{0.000000in}}%
\pgfpathlineto{\pgfqpoint{0.393701in}{0.393701in}}%
\pgfpathlineto{\pgfqpoint{0.000000in}{0.393701in}}%
\pgfpathclose%
\pgfusepath{}%
\end{pgfscope}%
\begin{pgfscope}%
\pgfsetbuttcap%
\pgfsetmiterjoin%
\pgfsetlinewidth{0.000000pt}%
\definecolor{currentstroke}{rgb}{0.000000,0.000000,0.000000}%
\pgfsetstrokecolor{currentstroke}%
\pgfsetstrokeopacity{0.000000}%
\pgfsetdash{}{0pt}%
\pgfpathmoveto{\pgfqpoint{0.000000in}{0.000000in}}%
\pgfpathlineto{\pgfqpoint{0.393701in}{0.000000in}}%
\pgfpathlineto{\pgfqpoint{0.393701in}{0.393701in}}%
\pgfpathlineto{\pgfqpoint{0.000000in}{0.393701in}}%
\pgfpathclose%
\pgfusepath{}%
\end{pgfscope}%
\begin{pgfscope}%
\pgfpathrectangle{\pgfqpoint{0.000000in}{0.000000in}}{\pgfqpoint{0.393701in}{0.393701in}}%
\pgfusepath{clip}%
\pgfsetroundcap%
\pgfsetroundjoin%
\pgfsetlinewidth{1.003750pt}%
\definecolor{currentstroke}{rgb}{0.121569,0.466667,0.705882}%
\pgfsetstrokecolor{currentstroke}%
\pgfsetdash{}{0pt}%
\pgfpathmoveto{\pgfqpoint{0.017896in}{0.375806in}}%
\pgfpathlineto{\pgfqpoint{0.021511in}{0.375639in}}%
\pgfpathlineto{\pgfqpoint{0.025126in}{0.375141in}}%
\pgfpathlineto{\pgfqpoint{0.028741in}{0.374313in}}%
\pgfpathlineto{\pgfqpoint{0.032357in}{0.373156in}}%
\pgfpathlineto{\pgfqpoint{0.035972in}{0.371675in}}%
\pgfpathlineto{\pgfqpoint{0.039587in}{0.369872in}}%
\pgfpathlineto{\pgfqpoint{0.043202in}{0.367754in}}%
\pgfpathlineto{\pgfqpoint{0.046818in}{0.365325in}}%
\pgfpathlineto{\pgfqpoint{0.050433in}{0.362592in}}%
\pgfpathlineto{\pgfqpoint{0.054048in}{0.359564in}}%
\pgfpathlineto{\pgfqpoint{0.057663in}{0.356247in}}%
\pgfpathlineto{\pgfqpoint{0.061279in}{0.352650in}}%
\pgfpathlineto{\pgfqpoint{0.064894in}{0.348784in}}%
\pgfpathlineto{\pgfqpoint{0.068509in}{0.344658in}}%
\pgfpathlineto{\pgfqpoint{0.072124in}{0.340282in}}%
\pgfpathlineto{\pgfqpoint{0.075740in}{0.335669in}}%
\pgfpathlineto{\pgfqpoint{0.079355in}{0.330831in}}%
\pgfpathlineto{\pgfqpoint{0.082970in}{0.325778in}}%
\pgfpathlineto{\pgfqpoint{0.086585in}{0.320524in}}%
\pgfpathlineto{\pgfqpoint{0.090201in}{0.315082in}}%
\pgfpathlineto{\pgfqpoint{0.093816in}{0.309465in}}%
\pgfpathlineto{\pgfqpoint{0.097431in}{0.303686in}}%
\pgfpathlineto{\pgfqpoint{0.101046in}{0.297760in}}%
\pgfpathlineto{\pgfqpoint{0.104662in}{0.291699in}}%
\pgfpathlineto{\pgfqpoint{0.108277in}{0.285519in}}%
\pgfpathlineto{\pgfqpoint{0.111892in}{0.279232in}}%
\pgfpathlineto{\pgfqpoint{0.115507in}{0.272853in}}%
\pgfpathlineto{\pgfqpoint{0.119123in}{0.266395in}}%
\pgfpathlineto{\pgfqpoint{0.122738in}{0.259872in}}%
\pgfpathlineto{\pgfqpoint{0.126353in}{0.253297in}}%
\pgfpathlineto{\pgfqpoint{0.129968in}{0.246685in}}%
\pgfpathlineto{\pgfqpoint{0.133584in}{0.240047in}}%
\pgfpathlineto{\pgfqpoint{0.137199in}{0.233397in}}%
\pgfpathlineto{\pgfqpoint{0.140814in}{0.226747in}}%
\pgfpathlineto{\pgfqpoint{0.144429in}{0.220109in}}%
\pgfpathlineto{\pgfqpoint{0.148045in}{0.213494in}}%
\pgfpathlineto{\pgfqpoint{0.151660in}{0.206914in}}%
\pgfpathlineto{\pgfqpoint{0.155275in}{0.200380in}}%
\pgfpathlineto{\pgfqpoint{0.158890in}{0.193902in}}%
\pgfpathlineto{\pgfqpoint{0.162506in}{0.187489in}}%
\pgfpathlineto{\pgfqpoint{0.166121in}{0.181151in}}%
\pgfpathlineto{\pgfqpoint{0.169736in}{0.174896in}}%
\pgfpathlineto{\pgfqpoint{0.173351in}{0.168733in}}%
\pgfpathlineto{\pgfqpoint{0.176967in}{0.162669in}}%
\pgfpathlineto{\pgfqpoint{0.180582in}{0.156711in}}%
\pgfpathlineto{\pgfqpoint{0.184197in}{0.150866in}}%
\pgfpathlineto{\pgfqpoint{0.187812in}{0.145140in}}%
\pgfpathlineto{\pgfqpoint{0.191428in}{0.139538in}}%
\pgfpathlineto{\pgfqpoint{0.195043in}{0.134064in}}%
\pgfpathlineto{\pgfqpoint{0.198658in}{0.128723in}}%
\pgfpathlineto{\pgfqpoint{0.202273in}{0.123519in}}%
\pgfpathlineto{\pgfqpoint{0.205889in}{0.118454in}}%
\pgfpathlineto{\pgfqpoint{0.209504in}{0.113532in}}%
\pgfpathlineto{\pgfqpoint{0.213119in}{0.108755in}}%
\pgfpathlineto{\pgfqpoint{0.216734in}{0.104123in}}%
\pgfpathlineto{\pgfqpoint{0.220350in}{0.099639in}}%
\pgfpathlineto{\pgfqpoint{0.223965in}{0.095303in}}%
\pgfpathlineto{\pgfqpoint{0.227580in}{0.091115in}}%
\pgfpathlineto{\pgfqpoint{0.231195in}{0.087075in}}%
\pgfpathlineto{\pgfqpoint{0.234811in}{0.083183in}}%
\pgfpathlineto{\pgfqpoint{0.238426in}{0.079437in}}%
\pgfpathlineto{\pgfqpoint{0.242041in}{0.075837in}}%
\pgfpathlineto{\pgfqpoint{0.245656in}{0.072381in}}%
\pgfpathlineto{\pgfqpoint{0.249272in}{0.069067in}}%
\pgfpathlineto{\pgfqpoint{0.252887in}{0.065893in}}%
\pgfpathlineto{\pgfqpoint{0.256502in}{0.062857in}}%
\pgfpathlineto{\pgfqpoint{0.260117in}{0.059955in}}%
\pgfpathlineto{\pgfqpoint{0.263733in}{0.057186in}}%
\pgfpathlineto{\pgfqpoint{0.267348in}{0.054545in}}%
\pgfpathlineto{\pgfqpoint{0.270963in}{0.052031in}}%
\pgfpathlineto{\pgfqpoint{0.274578in}{0.049639in}}%
\pgfpathlineto{\pgfqpoint{0.278194in}{0.047366in}}%
\pgfpathlineto{\pgfqpoint{0.281809in}{0.045209in}}%
\pgfpathlineto{\pgfqpoint{0.285424in}{0.043164in}}%
\pgfpathlineto{\pgfqpoint{0.289039in}{0.041227in}}%
\pgfpathlineto{\pgfqpoint{0.292655in}{0.039395in}}%
\pgfpathlineto{\pgfqpoint{0.296270in}{0.037664in}}%
\pgfpathlineto{\pgfqpoint{0.299885in}{0.036030in}}%
\pgfpathlineto{\pgfqpoint{0.303500in}{0.034489in}}%
\pgfpathlineto{\pgfqpoint{0.307116in}{0.033038in}}%
\pgfpathlineto{\pgfqpoint{0.310731in}{0.031672in}}%
\pgfpathlineto{\pgfqpoint{0.314346in}{0.030389in}}%
\pgfpathlineto{\pgfqpoint{0.317961in}{0.029184in}}%
\pgfpathlineto{\pgfqpoint{0.321577in}{0.028054in}}%
\pgfpathlineto{\pgfqpoint{0.325192in}{0.026995in}}%
\pgfpathlineto{\pgfqpoint{0.328807in}{0.026005in}}%
\pgfpathlineto{\pgfqpoint{0.332422in}{0.025079in}}%
\pgfpathlineto{\pgfqpoint{0.336038in}{0.024214in}}%
\pgfpathlineto{\pgfqpoint{0.339653in}{0.023407in}}%
\pgfpathlineto{\pgfqpoint{0.343268in}{0.022655in}}%
\pgfpathlineto{\pgfqpoint{0.346883in}{0.021954in}}%
\pgfpathlineto{\pgfqpoint{0.350499in}{0.021303in}}%
\pgfpathlineto{\pgfqpoint{0.354114in}{0.020698in}}%
\pgfpathlineto{\pgfqpoint{0.357729in}{0.020137in}}%
\pgfpathlineto{\pgfqpoint{0.361344in}{0.019616in}}%
\pgfpathlineto{\pgfqpoint{0.364960in}{0.019134in}}%
\pgfpathlineto{\pgfqpoint{0.368575in}{0.018688in}}%
\pgfpathlineto{\pgfqpoint{0.372190in}{0.018276in}}%
\pgfpathlineto{\pgfqpoint{0.375805in}{0.017896in}}%
\pgfusepath{stroke}%
\end{pgfscope}%
\begin{pgfscope}%
\pgfsetrectcap%
\pgfsetmiterjoin%
\pgfsetlinewidth{1.254687pt}%
\definecolor{currentstroke}{rgb}{0.150000,0.150000,0.150000}%
\pgfsetstrokecolor{currentstroke}%
\pgfsetdash{}{0pt}%
\pgfpathmoveto{\pgfqpoint{0.000000in}{-0.000000in}}%
\pgfpathlineto{\pgfqpoint{0.000000in}{0.393701in}}%
\pgfusepath{stroke}%
\end{pgfscope}%
\begin{pgfscope}%
\pgfsetrectcap%
\pgfsetmiterjoin%
\pgfsetlinewidth{1.254687pt}%
\definecolor{currentstroke}{rgb}{0.150000,0.150000,0.150000}%
\pgfsetstrokecolor{currentstroke}%
\pgfsetdash{}{0pt}%
\pgfpathmoveto{\pgfqpoint{0.000000in}{0.000000in}}%
\pgfpathlineto{\pgfqpoint{0.393701in}{0.000000in}}%
\pgfusepath{stroke}%
\end{pgfscope}%
\end{pgfpicture}%
\makeatother%
\endgroup%
 &

\begingroup%
\makeatletter%
\begin{pgfpicture}%
\pgfpathrectangle{\pgfpointorigin}{\pgfqpoint{0.393701in}{0.393701in}}%
\pgfusepath{use as bounding box, clip}%
\begin{pgfscope}%
\pgfsetbuttcap%
\pgfsetmiterjoin%
\pgfsetlinewidth{0.000000pt}%
\definecolor{currentstroke}{rgb}{0.000000,0.000000,0.000000}%
\pgfsetstrokecolor{currentstroke}%
\pgfsetstrokeopacity{0.000000}%
\pgfsetdash{}{0pt}%
\pgfpathmoveto{\pgfqpoint{0.000000in}{0.000000in}}%
\pgfpathlineto{\pgfqpoint{0.393701in}{0.000000in}}%
\pgfpathlineto{\pgfqpoint{0.393701in}{0.393701in}}%
\pgfpathlineto{\pgfqpoint{0.000000in}{0.393701in}}%
\pgfpathclose%
\pgfusepath{}%
\end{pgfscope}%
\begin{pgfscope}%
\pgfsetbuttcap%
\pgfsetmiterjoin%
\pgfsetlinewidth{0.000000pt}%
\definecolor{currentstroke}{rgb}{0.000000,0.000000,0.000000}%
\pgfsetstrokecolor{currentstroke}%
\pgfsetstrokeopacity{0.000000}%
\pgfsetdash{}{0pt}%
\pgfpathmoveto{\pgfqpoint{0.000000in}{0.000000in}}%
\pgfpathlineto{\pgfqpoint{0.393701in}{0.000000in}}%
\pgfpathlineto{\pgfqpoint{0.393701in}{0.393701in}}%
\pgfpathlineto{\pgfqpoint{0.000000in}{0.393701in}}%
\pgfpathclose%
\pgfusepath{}%
\end{pgfscope}%
\begin{pgfscope}%
\pgfpathrectangle{\pgfqpoint{0.000000in}{0.000000in}}{\pgfqpoint{0.393701in}{0.393701in}}%
\pgfusepath{clip}%
\pgfsetroundcap%
\pgfsetroundjoin%
\pgfsetlinewidth{1.003750pt}%
\definecolor{currentstroke}{rgb}{0.121569,0.466667,0.705882}%
\pgfsetstrokecolor{currentstroke}%
\pgfsetdash{}{0pt}%
\pgfpathmoveto{\pgfqpoint{0.017896in}{0.017896in}}%
\pgfpathlineto{\pgfqpoint{0.064658in}{0.019808in}}%
\pgfpathlineto{\pgfqpoint{0.084442in}{0.023364in}}%
\pgfpathlineto{\pgfqpoint{0.098830in}{0.028660in}}%
\pgfpathlineto{\pgfqpoint{0.111420in}{0.036372in}}%
\pgfpathlineto{\pgfqpoint{0.124010in}{0.048162in}}%
\pgfpathlineto{\pgfqpoint{0.134801in}{0.062452in}}%
\pgfpathlineto{\pgfqpoint{0.147391in}{0.084983in}}%
\pgfpathlineto{\pgfqpoint{0.159980in}{0.114470in}}%
\pgfpathlineto{\pgfqpoint{0.174369in}{0.156538in}}%
\pgfpathlineto{\pgfqpoint{0.194153in}{0.225128in}}%
\pgfpathlineto{\pgfqpoint{0.221131in}{0.318100in}}%
\pgfpathlineto{\pgfqpoint{0.233721in}{0.350641in}}%
\pgfpathlineto{\pgfqpoint{0.242713in}{0.366384in}}%
\pgfpathlineto{\pgfqpoint{0.249908in}{0.373642in}}%
\pgfpathlineto{\pgfqpoint{0.255303in}{0.375751in}}%
\pgfpathlineto{\pgfqpoint{0.260699in}{0.374939in}}%
\pgfpathlineto{\pgfqpoint{0.266094in}{0.371224in}}%
\pgfpathlineto{\pgfqpoint{0.273289in}{0.361924in}}%
\pgfpathlineto{\pgfqpoint{0.282281in}{0.343902in}}%
\pgfpathlineto{\pgfqpoint{0.293073in}{0.314482in}}%
\pgfpathlineto{\pgfqpoint{0.309259in}{0.259928in}}%
\pgfpathlineto{\pgfqpoint{0.343432in}{0.141595in}}%
\pgfpathlineto{\pgfqpoint{0.357820in}{0.102400in}}%
\pgfpathlineto{\pgfqpoint{0.370410in}{0.075604in}}%
\pgfpathlineto{\pgfqpoint{0.375806in}{0.066232in}}%
\pgfpathlineto{\pgfqpoint{0.375806in}{0.066232in}}%
\pgfusepath{stroke}%
\end{pgfscope}%
\begin{pgfscope}%
\pgfsetrectcap%
\pgfsetmiterjoin%
\pgfsetlinewidth{1.254687pt}%
\definecolor{currentstroke}{rgb}{0.150000,0.150000,0.150000}%
\pgfsetstrokecolor{currentstroke}%
\pgfsetdash{}{0pt}%
\pgfpathmoveto{\pgfqpoint{0.196851in}{0.000000in}}%
\pgfpathlineto{\pgfqpoint{0.196851in}{0.393701in}}%
\pgfusepath{stroke}%
\end{pgfscope}%
\begin{pgfscope}%
\pgfsetrectcap%
\pgfsetmiterjoin%
\pgfsetlinewidth{1.254687pt}%
\definecolor{currentstroke}{rgb}{0.150000,0.150000,0.150000}%
\pgfsetstrokecolor{currentstroke}%
\pgfsetdash{}{0pt}%
\pgfpathmoveto{\pgfqpoint{0.000000in}{0.000000in}}%
\pgfpathlineto{\pgfqpoint{0.393701in}{0.000000in}}%
\pgfusepath{stroke}%
\end{pgfscope}%
\end{pgfpicture}%
\makeatother%
\endgroup%
 &

\begingroup%
\makeatletter%
\begin{pgfpicture}%
\pgfpathrectangle{\pgfpointorigin}{\pgfqpoint{0.393701in}{0.393701in}}%
\pgfusepath{use as bounding box, clip}%
\begin{pgfscope}%
\pgfsetbuttcap%
\pgfsetmiterjoin%
\pgfsetlinewidth{0.000000pt}%
\definecolor{currentstroke}{rgb}{0.000000,0.000000,0.000000}%
\pgfsetstrokecolor{currentstroke}%
\pgfsetstrokeopacity{0.000000}%
\pgfsetdash{}{0pt}%
\pgfpathmoveto{\pgfqpoint{0.000000in}{0.000000in}}%
\pgfpathlineto{\pgfqpoint{0.393701in}{0.000000in}}%
\pgfpathlineto{\pgfqpoint{0.393701in}{0.393701in}}%
\pgfpathlineto{\pgfqpoint{0.000000in}{0.393701in}}%
\pgfpathclose%
\pgfusepath{}%
\end{pgfscope}%
\begin{pgfscope}%
\pgfsetbuttcap%
\pgfsetmiterjoin%
\pgfsetlinewidth{0.000000pt}%
\definecolor{currentstroke}{rgb}{0.000000,0.000000,0.000000}%
\pgfsetstrokecolor{currentstroke}%
\pgfsetstrokeopacity{0.000000}%
\pgfsetdash{}{0pt}%
\pgfpathmoveto{\pgfqpoint{0.000000in}{0.000000in}}%
\pgfpathlineto{\pgfqpoint{0.393701in}{0.000000in}}%
\pgfpathlineto{\pgfqpoint{0.393701in}{0.393701in}}%
\pgfpathlineto{\pgfqpoint{0.000000in}{0.393701in}}%
\pgfpathclose%
\pgfusepath{}%
\end{pgfscope}%
\begin{pgfscope}%
\pgfpathrectangle{\pgfqpoint{0.000000in}{0.000000in}}{\pgfqpoint{0.393701in}{0.393701in}}%
\pgfusepath{clip}%
\pgfsetroundcap%
\pgfsetroundjoin%
\pgfsetlinewidth{1.003750pt}%
\definecolor{currentstroke}{rgb}{0.121569,0.466667,0.705882}%
\pgfsetstrokecolor{currentstroke}%
\pgfsetdash{}{0pt}%
\pgfpathmoveto{\pgfqpoint{0.017896in}{0.212937in}}%
\pgfpathlineto{\pgfqpoint{0.021511in}{0.220543in}}%
\pgfpathlineto{\pgfqpoint{0.025126in}{0.228134in}}%
\pgfpathlineto{\pgfqpoint{0.028741in}{0.235697in}}%
\pgfpathlineto{\pgfqpoint{0.032357in}{0.243215in}}%
\pgfpathlineto{\pgfqpoint{0.035972in}{0.250676in}}%
\pgfpathlineto{\pgfqpoint{0.039587in}{0.258061in}}%
\pgfpathlineto{\pgfqpoint{0.043202in}{0.265357in}}%
\pgfpathlineto{\pgfqpoint{0.046818in}{0.272547in}}%
\pgfpathlineto{\pgfqpoint{0.050433in}{0.279616in}}%
\pgfpathlineto{\pgfqpoint{0.054048in}{0.286547in}}%
\pgfpathlineto{\pgfqpoint{0.057663in}{0.293325in}}%
\pgfpathlineto{\pgfqpoint{0.061279in}{0.299934in}}%
\pgfpathlineto{\pgfqpoint{0.064894in}{0.306358in}}%
\pgfpathlineto{\pgfqpoint{0.068509in}{0.312582in}}%
\pgfpathlineto{\pgfqpoint{0.072124in}{0.318590in}}%
\pgfpathlineto{\pgfqpoint{0.075740in}{0.324369in}}%
\pgfpathlineto{\pgfqpoint{0.079355in}{0.329903in}}%
\pgfpathlineto{\pgfqpoint{0.082970in}{0.335179in}}%
\pgfpathlineto{\pgfqpoint{0.086585in}{0.340183in}}%
\pgfpathlineto{\pgfqpoint{0.090201in}{0.344902in}}%
\pgfpathlineto{\pgfqpoint{0.093816in}{0.349323in}}%
\pgfpathlineto{\pgfqpoint{0.097431in}{0.353437in}}%
\pgfpathlineto{\pgfqpoint{0.101046in}{0.357230in}}%
\pgfpathlineto{\pgfqpoint{0.104662in}{0.360694in}}%
\pgfpathlineto{\pgfqpoint{0.108277in}{0.363819in}}%
\pgfpathlineto{\pgfqpoint{0.111892in}{0.366597in}}%
\pgfpathlineto{\pgfqpoint{0.115507in}{0.369020in}}%
\pgfpathlineto{\pgfqpoint{0.119123in}{0.371081in}}%
\pgfpathlineto{\pgfqpoint{0.122738in}{0.372776in}}%
\pgfpathlineto{\pgfqpoint{0.126353in}{0.374099in}}%
\pgfpathlineto{\pgfqpoint{0.129968in}{0.375046in}}%
\pgfpathlineto{\pgfqpoint{0.133584in}{0.375615in}}%
\pgfpathlineto{\pgfqpoint{0.137199in}{0.375806in}}%
\pgfpathlineto{\pgfqpoint{0.140814in}{0.375615in}}%
\pgfpathlineto{\pgfqpoint{0.144429in}{0.375046in}}%
\pgfpathlineto{\pgfqpoint{0.148045in}{0.374099in}}%
\pgfpathlineto{\pgfqpoint{0.151660in}{0.372776in}}%
\pgfpathlineto{\pgfqpoint{0.155275in}{0.371081in}}%
\pgfpathlineto{\pgfqpoint{0.158890in}{0.369020in}}%
\pgfpathlineto{\pgfqpoint{0.162506in}{0.366597in}}%
\pgfpathlineto{\pgfqpoint{0.166121in}{0.363819in}}%
\pgfpathlineto{\pgfqpoint{0.169736in}{0.360694in}}%
\pgfpathlineto{\pgfqpoint{0.173351in}{0.357230in}}%
\pgfpathlineto{\pgfqpoint{0.176967in}{0.353437in}}%
\pgfpathlineto{\pgfqpoint{0.180582in}{0.349323in}}%
\pgfpathlineto{\pgfqpoint{0.184197in}{0.344902in}}%
\pgfpathlineto{\pgfqpoint{0.187812in}{0.340183in}}%
\pgfpathlineto{\pgfqpoint{0.191428in}{0.335179in}}%
\pgfpathlineto{\pgfqpoint{0.195043in}{0.329903in}}%
\pgfpathlineto{\pgfqpoint{0.198658in}{0.324369in}}%
\pgfpathlineto{\pgfqpoint{0.202273in}{0.318590in}}%
\pgfpathlineto{\pgfqpoint{0.205889in}{0.312582in}}%
\pgfpathlineto{\pgfqpoint{0.209504in}{0.306358in}}%
\pgfpathlineto{\pgfqpoint{0.213119in}{0.299934in}}%
\pgfpathlineto{\pgfqpoint{0.216734in}{0.293325in}}%
\pgfpathlineto{\pgfqpoint{0.220350in}{0.286547in}}%
\pgfpathlineto{\pgfqpoint{0.223965in}{0.279616in}}%
\pgfpathlineto{\pgfqpoint{0.227580in}{0.272547in}}%
\pgfpathlineto{\pgfqpoint{0.231195in}{0.265357in}}%
\pgfpathlineto{\pgfqpoint{0.234811in}{0.258061in}}%
\pgfpathlineto{\pgfqpoint{0.238426in}{0.250676in}}%
\pgfpathlineto{\pgfqpoint{0.242041in}{0.243215in}}%
\pgfpathlineto{\pgfqpoint{0.245656in}{0.235697in}}%
\pgfpathlineto{\pgfqpoint{0.249272in}{0.228134in}}%
\pgfpathlineto{\pgfqpoint{0.252887in}{0.220543in}}%
\pgfpathlineto{\pgfqpoint{0.256502in}{0.212937in}}%
\pgfpathlineto{\pgfqpoint{0.260117in}{0.205331in}}%
\pgfpathlineto{\pgfqpoint{0.263733in}{0.197740in}}%
\pgfpathlineto{\pgfqpoint{0.267348in}{0.190175in}}%
\pgfpathlineto{\pgfqpoint{0.270963in}{0.182650in}}%
\pgfpathlineto{\pgfqpoint{0.274578in}{0.175177in}}%
\pgfpathlineto{\pgfqpoint{0.278194in}{0.167767in}}%
\pgfpathlineto{\pgfqpoint{0.281809in}{0.160433in}}%
\pgfpathlineto{\pgfqpoint{0.285424in}{0.153185in}}%
\pgfpathlineto{\pgfqpoint{0.289039in}{0.146031in}}%
\pgfpathlineto{\pgfqpoint{0.292655in}{0.138983in}}%
\pgfpathlineto{\pgfqpoint{0.296270in}{0.132048in}}%
\pgfpathlineto{\pgfqpoint{0.299885in}{0.125234in}}%
\pgfpathlineto{\pgfqpoint{0.303500in}{0.118549in}}%
\pgfpathlineto{\pgfqpoint{0.307116in}{0.112000in}}%
\pgfpathlineto{\pgfqpoint{0.310731in}{0.105593in}}%
\pgfpathlineto{\pgfqpoint{0.314346in}{0.099333in}}%
\pgfpathlineto{\pgfqpoint{0.317961in}{0.093225in}}%
\pgfpathlineto{\pgfqpoint{0.321577in}{0.087273in}}%
\pgfpathlineto{\pgfqpoint{0.325192in}{0.081481in}}%
\pgfpathlineto{\pgfqpoint{0.328807in}{0.075851in}}%
\pgfpathlineto{\pgfqpoint{0.332422in}{0.070387in}}%
\pgfpathlineto{\pgfqpoint{0.336038in}{0.065090in}}%
\pgfpathlineto{\pgfqpoint{0.339653in}{0.059962in}}%
\pgfpathlineto{\pgfqpoint{0.343268in}{0.055003in}}%
\pgfpathlineto{\pgfqpoint{0.346883in}{0.050213in}}%
\pgfpathlineto{\pgfqpoint{0.350499in}{0.045593in}}%
\pgfpathlineto{\pgfqpoint{0.354114in}{0.041142in}}%
\pgfpathlineto{\pgfqpoint{0.357729in}{0.036858in}}%
\pgfpathlineto{\pgfqpoint{0.361344in}{0.032741in}}%
\pgfpathlineto{\pgfqpoint{0.364960in}{0.028788in}}%
\pgfpathlineto{\pgfqpoint{0.368575in}{0.024998in}}%
\pgfpathlineto{\pgfqpoint{0.372190in}{0.021368in}}%
\pgfpathlineto{\pgfqpoint{0.375805in}{0.017896in}}%
\pgfusepath{stroke}%
\end{pgfscope}%
\begin{pgfscope}%
\pgfsetrectcap%
\pgfsetmiterjoin%
\pgfsetlinewidth{1.254687pt}%
\definecolor{currentstroke}{rgb}{0.150000,0.150000,0.150000}%
\pgfsetstrokecolor{currentstroke}%
\pgfsetdash{}{0pt}%
\pgfpathmoveto{\pgfqpoint{0.000000in}{0.000000in}}%
\pgfpathlineto{\pgfqpoint{0.000000in}{0.393701in}}%
\pgfusepath{stroke}%
\end{pgfscope}%
\begin{pgfscope}%
\pgfsetrectcap%
\pgfsetmiterjoin%
\pgfsetlinewidth{1.254687pt}%
\definecolor{currentstroke}{rgb}{0.150000,0.150000,0.150000}%
\pgfsetstrokecolor{currentstroke}%
\pgfsetdash{}{0pt}%
\pgfpathmoveto{\pgfqpoint{0.000000in}{0.000000in}}%
\pgfpathlineto{\pgfqpoint{0.393701in}{0.000000in}}%
\pgfusepath{stroke}%
\end{pgfscope}%
\end{pgfpicture}%
\makeatother%
\endgroup%
\tabularnewline
  \midrule

  Laplace &

\begingroup%
\makeatletter%
\begin{pgfpicture}%
\pgfpathrectangle{\pgfpointorigin}{\pgfqpoint{0.393701in}{0.393701in}}%
\pgfusepath{use as bounding box, clip}%
\begin{pgfscope}%
\pgfsetbuttcap%
\pgfsetmiterjoin%
\pgfsetlinewidth{0.000000pt}%
\definecolor{currentstroke}{rgb}{0.000000,0.000000,0.000000}%
\pgfsetstrokecolor{currentstroke}%
\pgfsetstrokeopacity{0.000000}%
\pgfsetdash{}{0pt}%
\pgfpathmoveto{\pgfqpoint{0.000000in}{0.000000in}}%
\pgfpathlineto{\pgfqpoint{0.393701in}{0.000000in}}%
\pgfpathlineto{\pgfqpoint{0.393701in}{0.393701in}}%
\pgfpathlineto{\pgfqpoint{0.000000in}{0.393701in}}%
\pgfpathclose%
\pgfusepath{}%
\end{pgfscope}%
\begin{pgfscope}%
\pgfsetbuttcap%
\pgfsetmiterjoin%
\pgfsetlinewidth{0.000000pt}%
\definecolor{currentstroke}{rgb}{0.000000,0.000000,0.000000}%
\pgfsetstrokecolor{currentstroke}%
\pgfsetstrokeopacity{0.000000}%
\pgfsetdash{}{0pt}%
\pgfpathmoveto{\pgfqpoint{0.000000in}{0.000000in}}%
\pgfpathlineto{\pgfqpoint{0.393701in}{0.000000in}}%
\pgfpathlineto{\pgfqpoint{0.393701in}{0.393701in}}%
\pgfpathlineto{\pgfqpoint{0.000000in}{0.393701in}}%
\pgfpathclose%
\pgfusepath{}%
\end{pgfscope}%
\begin{pgfscope}%
\pgfpathrectangle{\pgfqpoint{0.000000in}{0.000000in}}{\pgfqpoint{0.393701in}{0.393701in}}%
\pgfusepath{clip}%
\pgfsetroundcap%
\pgfsetroundjoin%
\pgfsetlinewidth{1.003750pt}%
\definecolor{currentstroke}{rgb}{0.121569,0.466667,0.705882}%
\pgfsetstrokecolor{currentstroke}%
\pgfsetdash{}{0pt}%
\pgfpathmoveto{\pgfqpoint{0.017896in}{0.017896in}}%
\pgfpathlineto{\pgfqpoint{0.041277in}{0.027039in}}%
\pgfpathlineto{\pgfqpoint{0.061061in}{0.038127in}}%
\pgfpathlineto{\pgfqpoint{0.079046in}{0.051952in}}%
\pgfpathlineto{\pgfqpoint{0.095233in}{0.068508in}}%
\pgfpathlineto{\pgfqpoint{0.109621in}{0.087512in}}%
\pgfpathlineto{\pgfqpoint{0.124010in}{0.111699in}}%
\pgfpathlineto{\pgfqpoint{0.138398in}{0.142484in}}%
\pgfpathlineto{\pgfqpoint{0.152786in}{0.181667in}}%
\pgfpathlineto{\pgfqpoint{0.165376in}{0.224628in}}%
\pgfpathlineto{\pgfqpoint{0.177966in}{0.277683in}}%
\pgfpathlineto{\pgfqpoint{0.190556in}{0.343205in}}%
\pgfpathlineto{\pgfqpoint{0.195951in}{0.375805in}}%
\pgfpathlineto{\pgfqpoint{0.197750in}{0.375806in}}%
\pgfpathlineto{\pgfqpoint{0.212138in}{0.295015in}}%
\pgfpathlineto{\pgfqpoint{0.226526in}{0.231539in}}%
\pgfpathlineto{\pgfqpoint{0.240915in}{0.181667in}}%
\pgfpathlineto{\pgfqpoint{0.255303in}{0.142484in}}%
\pgfpathlineto{\pgfqpoint{0.269691in}{0.111699in}}%
\pgfpathlineto{\pgfqpoint{0.284080in}{0.087512in}}%
\pgfpathlineto{\pgfqpoint{0.300267in}{0.066439in}}%
\pgfpathlineto{\pgfqpoint{0.316454in}{0.050374in}}%
\pgfpathlineto{\pgfqpoint{0.334439in}{0.036961in}}%
\pgfpathlineto{\pgfqpoint{0.354223in}{0.026201in}}%
\pgfpathlineto{\pgfqpoint{0.375806in}{0.017896in}}%
\pgfpathlineto{\pgfqpoint{0.375806in}{0.017896in}}%
\pgfusepath{stroke}%
\end{pgfscope}%
\begin{pgfscope}%
\pgfsetrectcap%
\pgfsetmiterjoin%
\pgfsetlinewidth{1.254687pt}%
\definecolor{currentstroke}{rgb}{0.150000,0.150000,0.150000}%
\pgfsetstrokecolor{currentstroke}%
\pgfsetdash{}{0pt}%
\pgfpathmoveto{\pgfqpoint{0.196851in}{0.000000in}}%
\pgfpathlineto{\pgfqpoint{0.196851in}{0.393701in}}%
\pgfusepath{stroke}%
\end{pgfscope}%
\begin{pgfscope}%
\pgfsetrectcap%
\pgfsetmiterjoin%
\pgfsetlinewidth{1.254687pt}%
\definecolor{currentstroke}{rgb}{0.150000,0.150000,0.150000}%
\pgfsetstrokecolor{currentstroke}%
\pgfsetdash{}{0pt}%
\pgfpathmoveto{\pgfqpoint{0.000000in}{0.000000in}}%
\pgfpathlineto{\pgfqpoint{0.393701in}{0.000000in}}%
\pgfusepath{stroke}%
\end{pgfscope}%
\end{pgfpicture}%
\makeatother%
\endgroup%
 &

\begingroup%
\makeatletter%
\begin{pgfpicture}%
\pgfpathrectangle{\pgfpointorigin}{\pgfqpoint{0.393701in}{0.393701in}}%
\pgfusepath{use as bounding box, clip}%
\begin{pgfscope}%
\pgfsetbuttcap%
\pgfsetmiterjoin%
\pgfsetlinewidth{0.000000pt}%
\definecolor{currentstroke}{rgb}{0.000000,0.000000,0.000000}%
\pgfsetstrokecolor{currentstroke}%
\pgfsetstrokeopacity{0.000000}%
\pgfsetdash{}{0pt}%
\pgfpathmoveto{\pgfqpoint{0.000000in}{0.000000in}}%
\pgfpathlineto{\pgfqpoint{0.393701in}{0.000000in}}%
\pgfpathlineto{\pgfqpoint{0.393701in}{0.393701in}}%
\pgfpathlineto{\pgfqpoint{0.000000in}{0.393701in}}%
\pgfpathclose%
\pgfusepath{}%
\end{pgfscope}%
\begin{pgfscope}%
\pgfsetbuttcap%
\pgfsetmiterjoin%
\pgfsetlinewidth{0.000000pt}%
\definecolor{currentstroke}{rgb}{0.000000,0.000000,0.000000}%
\pgfsetstrokecolor{currentstroke}%
\pgfsetstrokeopacity{0.000000}%
\pgfsetdash{}{0pt}%
\pgfpathmoveto{\pgfqpoint{0.000000in}{0.000000in}}%
\pgfpathlineto{\pgfqpoint{0.393701in}{0.000000in}}%
\pgfpathlineto{\pgfqpoint{0.393701in}{0.393701in}}%
\pgfpathlineto{\pgfqpoint{0.000000in}{0.393701in}}%
\pgfpathclose%
\pgfusepath{}%
\end{pgfscope}%
\begin{pgfscope}%
\pgfpathrectangle{\pgfqpoint{0.000000in}{0.000000in}}{\pgfqpoint{0.393701in}{0.393701in}}%
\pgfusepath{clip}%
\pgfsetroundcap%
\pgfsetroundjoin%
\pgfsetlinewidth{1.003750pt}%
\definecolor{currentstroke}{rgb}{0.121569,0.466667,0.705882}%
\pgfsetstrokecolor{currentstroke}%
\pgfsetdash{}{0pt}%
\pgfpathmoveto{\pgfqpoint{0.017896in}{0.375806in}}%
\pgfpathlineto{\pgfqpoint{0.021511in}{0.364563in}}%
\pgfpathlineto{\pgfqpoint{0.025126in}{0.353655in}}%
\pgfpathlineto{\pgfqpoint{0.028741in}{0.343074in}}%
\pgfpathlineto{\pgfqpoint{0.032357in}{0.332808in}}%
\pgfpathlineto{\pgfqpoint{0.035972in}{0.322849in}}%
\pgfpathlineto{\pgfqpoint{0.039587in}{0.313186in}}%
\pgfpathlineto{\pgfqpoint{0.043202in}{0.303813in}}%
\pgfpathlineto{\pgfqpoint{0.046818in}{0.294719in}}%
\pgfpathlineto{\pgfqpoint{0.050433in}{0.285896in}}%
\pgfpathlineto{\pgfqpoint{0.054048in}{0.277337in}}%
\pgfpathlineto{\pgfqpoint{0.057663in}{0.269033in}}%
\pgfpathlineto{\pgfqpoint{0.061279in}{0.260978in}}%
\pgfpathlineto{\pgfqpoint{0.064894in}{0.253162in}}%
\pgfpathlineto{\pgfqpoint{0.068509in}{0.245580in}}%
\pgfpathlineto{\pgfqpoint{0.072124in}{0.238224in}}%
\pgfpathlineto{\pgfqpoint{0.075740in}{0.231088in}}%
\pgfpathlineto{\pgfqpoint{0.079355in}{0.224165in}}%
\pgfpathlineto{\pgfqpoint{0.082970in}{0.217448in}}%
\pgfpathlineto{\pgfqpoint{0.086585in}{0.210932in}}%
\pgfpathlineto{\pgfqpoint{0.090201in}{0.204611in}}%
\pgfpathlineto{\pgfqpoint{0.093816in}{0.198478in}}%
\pgfpathlineto{\pgfqpoint{0.097431in}{0.192528in}}%
\pgfpathlineto{\pgfqpoint{0.101046in}{0.186756in}}%
\pgfpathlineto{\pgfqpoint{0.104662in}{0.181156in}}%
\pgfpathlineto{\pgfqpoint{0.108277in}{0.175723in}}%
\pgfpathlineto{\pgfqpoint{0.111892in}{0.170452in}}%
\pgfpathlineto{\pgfqpoint{0.115507in}{0.165339in}}%
\pgfpathlineto{\pgfqpoint{0.119123in}{0.160378in}}%
\pgfpathlineto{\pgfqpoint{0.122738in}{0.155565in}}%
\pgfpathlineto{\pgfqpoint{0.126353in}{0.150896in}}%
\pgfpathlineto{\pgfqpoint{0.129968in}{0.146367in}}%
\pgfpathlineto{\pgfqpoint{0.133584in}{0.141972in}}%
\pgfpathlineto{\pgfqpoint{0.137199in}{0.137709in}}%
\pgfpathlineto{\pgfqpoint{0.140814in}{0.133573in}}%
\pgfpathlineto{\pgfqpoint{0.144429in}{0.129561in}}%
\pgfpathlineto{\pgfqpoint{0.148045in}{0.125668in}}%
\pgfpathlineto{\pgfqpoint{0.151660in}{0.121891in}}%
\pgfpathlineto{\pgfqpoint{0.155275in}{0.118227in}}%
\pgfpathlineto{\pgfqpoint{0.158890in}{0.114673in}}%
\pgfpathlineto{\pgfqpoint{0.162506in}{0.111224in}}%
\pgfpathlineto{\pgfqpoint{0.166121in}{0.107879in}}%
\pgfpathlineto{\pgfqpoint{0.169736in}{0.104633in}}%
\pgfpathlineto{\pgfqpoint{0.173351in}{0.101485in}}%
\pgfpathlineto{\pgfqpoint{0.176967in}{0.098430in}}%
\pgfpathlineto{\pgfqpoint{0.180582in}{0.095466in}}%
\pgfpathlineto{\pgfqpoint{0.184197in}{0.092591in}}%
\pgfpathlineto{\pgfqpoint{0.187812in}{0.089802in}}%
\pgfpathlineto{\pgfqpoint{0.191428in}{0.087096in}}%
\pgfpathlineto{\pgfqpoint{0.195043in}{0.084471in}}%
\pgfpathlineto{\pgfqpoint{0.198658in}{0.081924in}}%
\pgfpathlineto{\pgfqpoint{0.202273in}{0.079453in}}%
\pgfpathlineto{\pgfqpoint{0.205889in}{0.077056in}}%
\pgfpathlineto{\pgfqpoint{0.209504in}{0.074730in}}%
\pgfpathlineto{\pgfqpoint{0.213119in}{0.072474in}}%
\pgfpathlineto{\pgfqpoint{0.216734in}{0.070285in}}%
\pgfpathlineto{\pgfqpoint{0.220350in}{0.068162in}}%
\pgfpathlineto{\pgfqpoint{0.223965in}{0.066101in}}%
\pgfpathlineto{\pgfqpoint{0.227580in}{0.064103in}}%
\pgfpathlineto{\pgfqpoint{0.231195in}{0.062164in}}%
\pgfpathlineto{\pgfqpoint{0.234811in}{0.060283in}}%
\pgfpathlineto{\pgfqpoint{0.238426in}{0.058458in}}%
\pgfpathlineto{\pgfqpoint{0.242041in}{0.056687in}}%
\pgfpathlineto{\pgfqpoint{0.245656in}{0.054970in}}%
\pgfpathlineto{\pgfqpoint{0.249272in}{0.053303in}}%
\pgfpathlineto{\pgfqpoint{0.252887in}{0.051687in}}%
\pgfpathlineto{\pgfqpoint{0.256502in}{0.050118in}}%
\pgfpathlineto{\pgfqpoint{0.260117in}{0.048597in}}%
\pgfpathlineto{\pgfqpoint{0.263733in}{0.047121in}}%
\pgfpathlineto{\pgfqpoint{0.267348in}{0.045689in}}%
\pgfpathlineto{\pgfqpoint{0.270963in}{0.044299in}}%
\pgfpathlineto{\pgfqpoint{0.274578in}{0.042951in}}%
\pgfpathlineto{\pgfqpoint{0.278194in}{0.041644in}}%
\pgfpathlineto{\pgfqpoint{0.281809in}{0.040375in}}%
\pgfpathlineto{\pgfqpoint{0.285424in}{0.039144in}}%
\pgfpathlineto{\pgfqpoint{0.289039in}{0.037950in}}%
\pgfpathlineto{\pgfqpoint{0.292655in}{0.036792in}}%
\pgfpathlineto{\pgfqpoint{0.296270in}{0.035668in}}%
\pgfpathlineto{\pgfqpoint{0.299885in}{0.034578in}}%
\pgfpathlineto{\pgfqpoint{0.303500in}{0.033520in}}%
\pgfpathlineto{\pgfqpoint{0.307116in}{0.032494in}}%
\pgfpathlineto{\pgfqpoint{0.310731in}{0.031499in}}%
\pgfpathlineto{\pgfqpoint{0.314346in}{0.030533in}}%
\pgfpathlineto{\pgfqpoint{0.317961in}{0.029596in}}%
\pgfpathlineto{\pgfqpoint{0.321577in}{0.028687in}}%
\pgfpathlineto{\pgfqpoint{0.325192in}{0.027805in}}%
\pgfpathlineto{\pgfqpoint{0.328807in}{0.026950in}}%
\pgfpathlineto{\pgfqpoint{0.332422in}{0.026120in}}%
\pgfpathlineto{\pgfqpoint{0.336038in}{0.025314in}}%
\pgfpathlineto{\pgfqpoint{0.339653in}{0.024533in}}%
\pgfpathlineto{\pgfqpoint{0.343268in}{0.023775in}}%
\pgfpathlineto{\pgfqpoint{0.346883in}{0.023040in}}%
\pgfpathlineto{\pgfqpoint{0.350499in}{0.022327in}}%
\pgfpathlineto{\pgfqpoint{0.354114in}{0.021635in}}%
\pgfpathlineto{\pgfqpoint{0.357729in}{0.020963in}}%
\pgfpathlineto{\pgfqpoint{0.361344in}{0.020312in}}%
\pgfpathlineto{\pgfqpoint{0.364960in}{0.019680in}}%
\pgfpathlineto{\pgfqpoint{0.368575in}{0.019067in}}%
\pgfpathlineto{\pgfqpoint{0.372190in}{0.018472in}}%
\pgfpathlineto{\pgfqpoint{0.375805in}{0.017896in}}%
\pgfusepath{stroke}%
\end{pgfscope}%
\begin{pgfscope}%
\pgfsetrectcap%
\pgfsetmiterjoin%
\pgfsetlinewidth{1.254687pt}%
\definecolor{currentstroke}{rgb}{0.150000,0.150000,0.150000}%
\pgfsetstrokecolor{currentstroke}%
\pgfsetdash{}{0pt}%
\pgfpathmoveto{\pgfqpoint{0.000000in}{0.000000in}}%
\pgfpathlineto{\pgfqpoint{0.000000in}{0.393701in}}%
\pgfusepath{stroke}%
\end{pgfscope}%
\begin{pgfscope}%
\pgfsetrectcap%
\pgfsetmiterjoin%
\pgfsetlinewidth{1.254687pt}%
\definecolor{currentstroke}{rgb}{0.150000,0.150000,0.150000}%
\pgfsetstrokecolor{currentstroke}%
\pgfsetdash{}{0pt}%
\pgfpathmoveto{\pgfqpoint{0.000000in}{0.000000in}}%
\pgfpathlineto{\pgfqpoint{0.393701in}{0.000000in}}%
\pgfusepath{stroke}%
\end{pgfscope}%
\end{pgfpicture}%
\makeatother%
\endgroup%
 &

\begingroup%
\makeatletter%
\begin{pgfpicture}%
\pgfpathrectangle{\pgfpointorigin}{\pgfqpoint{0.393701in}{0.393701in}}%
\pgfusepath{use as bounding box, clip}%
\begin{pgfscope}%
\pgfsetbuttcap%
\pgfsetmiterjoin%
\pgfsetlinewidth{0.000000pt}%
\definecolor{currentstroke}{rgb}{0.000000,0.000000,0.000000}%
\pgfsetstrokecolor{currentstroke}%
\pgfsetstrokeopacity{0.000000}%
\pgfsetdash{}{0pt}%
\pgfpathmoveto{\pgfqpoint{0.000000in}{0.000000in}}%
\pgfpathlineto{\pgfqpoint{0.393701in}{0.000000in}}%
\pgfpathlineto{\pgfqpoint{0.393701in}{0.393701in}}%
\pgfpathlineto{\pgfqpoint{0.000000in}{0.393701in}}%
\pgfpathclose%
\pgfusepath{}%
\end{pgfscope}%
\begin{pgfscope}%
\pgfsetbuttcap%
\pgfsetmiterjoin%
\pgfsetlinewidth{0.000000pt}%
\definecolor{currentstroke}{rgb}{0.000000,0.000000,0.000000}%
\pgfsetstrokecolor{currentstroke}%
\pgfsetstrokeopacity{0.000000}%
\pgfsetdash{}{0pt}%
\pgfpathmoveto{\pgfqpoint{0.000000in}{0.000000in}}%
\pgfpathlineto{\pgfqpoint{0.393701in}{0.000000in}}%
\pgfpathlineto{\pgfqpoint{0.393701in}{0.393701in}}%
\pgfpathlineto{\pgfqpoint{0.000000in}{0.393701in}}%
\pgfpathclose%
\pgfusepath{}%
\end{pgfscope}%
\begin{pgfscope}%
\pgfpathrectangle{\pgfqpoint{0.000000in}{0.000000in}}{\pgfqpoint{0.393701in}{0.393701in}}%
\pgfusepath{clip}%
\pgfsetroundcap%
\pgfsetroundjoin%
\pgfsetlinewidth{1.003750pt}%
\definecolor{currentstroke}{rgb}{0.121569,0.466667,0.705882}%
\pgfsetstrokecolor{currentstroke}%
\pgfsetdash{}{0pt}%
\pgfpathmoveto{\pgfqpoint{0.017896in}{0.017895in}}%
\pgfpathlineto{\pgfqpoint{0.053866in}{0.023479in}}%
\pgfpathlineto{\pgfqpoint{0.080844in}{0.030530in}}%
\pgfpathlineto{\pgfqpoint{0.104226in}{0.039830in}}%
\pgfpathlineto{\pgfqpoint{0.124010in}{0.051110in}}%
\pgfpathlineto{\pgfqpoint{0.141995in}{0.065172in}}%
\pgfpathlineto{\pgfqpoint{0.158182in}{0.082013in}}%
\pgfpathlineto{\pgfqpoint{0.172570in}{0.101344in}}%
\pgfpathlineto{\pgfqpoint{0.186959in}{0.125947in}}%
\pgfpathlineto{\pgfqpoint{0.201347in}{0.157262in}}%
\pgfpathlineto{\pgfqpoint{0.215735in}{0.197120in}}%
\pgfpathlineto{\pgfqpoint{0.228325in}{0.240819in}}%
\pgfpathlineto{\pgfqpoint{0.240915in}{0.294787in}}%
\pgfpathlineto{\pgfqpoint{0.253505in}{0.361437in}}%
\pgfpathlineto{\pgfqpoint{0.257102in}{0.375805in}}%
\pgfpathlineto{\pgfqpoint{0.271490in}{0.297652in}}%
\pgfpathlineto{\pgfqpoint{0.285878in}{0.236249in}}%
\pgfpathlineto{\pgfqpoint{0.300267in}{0.188005in}}%
\pgfpathlineto{\pgfqpoint{0.314655in}{0.150102in}}%
\pgfpathlineto{\pgfqpoint{0.329043in}{0.120321in}}%
\pgfpathlineto{\pgfqpoint{0.343432in}{0.096923in}}%
\pgfpathlineto{\pgfqpoint{0.359619in}{0.076539in}}%
\pgfpathlineto{\pgfqpoint{0.375806in}{0.060998in}}%
\pgfpathlineto{\pgfqpoint{0.375806in}{0.060998in}}%
\pgfusepath{stroke}%
\end{pgfscope}%
\begin{pgfscope}%
\pgfsetrectcap%
\pgfsetmiterjoin%
\pgfsetlinewidth{1.254687pt}%
\definecolor{currentstroke}{rgb}{0.150000,0.150000,0.150000}%
\pgfsetstrokecolor{currentstroke}%
\pgfsetdash{}{0pt}%
\pgfpathmoveto{\pgfqpoint{0.196851in}{0.000000in}}%
\pgfpathlineto{\pgfqpoint{0.196851in}{0.393701in}}%
\pgfusepath{stroke}%
\end{pgfscope}%
\begin{pgfscope}%
\pgfsetrectcap%
\pgfsetmiterjoin%
\pgfsetlinewidth{1.254687pt}%
\definecolor{currentstroke}{rgb}{0.150000,0.150000,0.150000}%
\pgfsetstrokecolor{currentstroke}%
\pgfsetdash{}{0pt}%
\pgfpathmoveto{\pgfqpoint{0.000000in}{0.000000in}}%
\pgfpathlineto{\pgfqpoint{0.393701in}{0.000000in}}%
\pgfusepath{stroke}%
\end{pgfscope}%
\end{pgfpicture}%
\makeatother%
\endgroup%
 &

\begingroup%
\makeatletter%
\begin{pgfpicture}%
\pgfpathrectangle{\pgfpointorigin}{\pgfqpoint{0.393701in}{0.393701in}}%
\pgfusepath{use as bounding box, clip}%
\begin{pgfscope}%
\pgfsetbuttcap%
\pgfsetmiterjoin%
\pgfsetlinewidth{0.000000pt}%
\definecolor{currentstroke}{rgb}{0.000000,0.000000,0.000000}%
\pgfsetstrokecolor{currentstroke}%
\pgfsetstrokeopacity{0.000000}%
\pgfsetdash{}{0pt}%
\pgfpathmoveto{\pgfqpoint{0.000000in}{0.000000in}}%
\pgfpathlineto{\pgfqpoint{0.393701in}{0.000000in}}%
\pgfpathlineto{\pgfqpoint{0.393701in}{0.393701in}}%
\pgfpathlineto{\pgfqpoint{0.000000in}{0.393701in}}%
\pgfpathclose%
\pgfusepath{}%
\end{pgfscope}%
\begin{pgfscope}%
\pgfsetbuttcap%
\pgfsetmiterjoin%
\pgfsetlinewidth{0.000000pt}%
\definecolor{currentstroke}{rgb}{0.000000,0.000000,0.000000}%
\pgfsetstrokecolor{currentstroke}%
\pgfsetstrokeopacity{0.000000}%
\pgfsetdash{}{0pt}%
\pgfpathmoveto{\pgfqpoint{0.000000in}{0.000000in}}%
\pgfpathlineto{\pgfqpoint{0.393701in}{0.000000in}}%
\pgfpathlineto{\pgfqpoint{0.393701in}{0.393701in}}%
\pgfpathlineto{\pgfqpoint{0.000000in}{0.393701in}}%
\pgfpathclose%
\pgfusepath{}%
\end{pgfscope}%
\begin{pgfscope}%
\pgfpathrectangle{\pgfqpoint{0.000000in}{0.000000in}}{\pgfqpoint{0.393701in}{0.393701in}}%
\pgfusepath{clip}%
\pgfsetroundcap%
\pgfsetroundjoin%
\pgfsetlinewidth{1.003750pt}%
\definecolor{currentstroke}{rgb}{0.121569,0.466667,0.705882}%
\pgfsetstrokecolor{currentstroke}%
\pgfsetdash{}{0pt}%
\pgfpathmoveto{\pgfqpoint{0.017896in}{0.114152in}}%
\pgfpathlineto{\pgfqpoint{0.021511in}{0.118837in}}%
\pgfpathlineto{\pgfqpoint{0.025126in}{0.123667in}}%
\pgfpathlineto{\pgfqpoint{0.028741in}{0.128644in}}%
\pgfpathlineto{\pgfqpoint{0.032357in}{0.133775in}}%
\pgfpathlineto{\pgfqpoint{0.035972in}{0.139064in}}%
\pgfpathlineto{\pgfqpoint{0.039587in}{0.144516in}}%
\pgfpathlineto{\pgfqpoint{0.043202in}{0.150135in}}%
\pgfpathlineto{\pgfqpoint{0.046818in}{0.155927in}}%
\pgfpathlineto{\pgfqpoint{0.050433in}{0.161897in}}%
\pgfpathlineto{\pgfqpoint{0.054048in}{0.168051in}}%
\pgfpathlineto{\pgfqpoint{0.057663in}{0.174395in}}%
\pgfpathlineto{\pgfqpoint{0.061279in}{0.180933in}}%
\pgfpathlineto{\pgfqpoint{0.064894in}{0.187673in}}%
\pgfpathlineto{\pgfqpoint{0.068509in}{0.194620in}}%
\pgfpathlineto{\pgfqpoint{0.072124in}{0.201781in}}%
\pgfpathlineto{\pgfqpoint{0.075740in}{0.209162in}}%
\pgfpathlineto{\pgfqpoint{0.079355in}{0.216770in}}%
\pgfpathlineto{\pgfqpoint{0.082970in}{0.224612in}}%
\pgfpathlineto{\pgfqpoint{0.086585in}{0.232696in}}%
\pgfpathlineto{\pgfqpoint{0.090201in}{0.241028in}}%
\pgfpathlineto{\pgfqpoint{0.093816in}{0.249617in}}%
\pgfpathlineto{\pgfqpoint{0.097431in}{0.258470in}}%
\pgfpathlineto{\pgfqpoint{0.101046in}{0.267595in}}%
\pgfpathlineto{\pgfqpoint{0.104662in}{0.277001in}}%
\pgfpathlineto{\pgfqpoint{0.108277in}{0.286696in}}%
\pgfpathlineto{\pgfqpoint{0.111892in}{0.296690in}}%
\pgfpathlineto{\pgfqpoint{0.115507in}{0.306991in}}%
\pgfpathlineto{\pgfqpoint{0.119123in}{0.317609in}}%
\pgfpathlineto{\pgfqpoint{0.122738in}{0.328554in}}%
\pgfpathlineto{\pgfqpoint{0.126353in}{0.339835in}}%
\pgfpathlineto{\pgfqpoint{0.129968in}{0.351464in}}%
\pgfpathlineto{\pgfqpoint{0.133584in}{0.363450in}}%
\pgfpathlineto{\pgfqpoint{0.137199in}{0.375806in}}%
\pgfpathlineto{\pgfqpoint{0.140814in}{0.363450in}}%
\pgfpathlineto{\pgfqpoint{0.144429in}{0.351464in}}%
\pgfpathlineto{\pgfqpoint{0.148045in}{0.339835in}}%
\pgfpathlineto{\pgfqpoint{0.151660in}{0.328554in}}%
\pgfpathlineto{\pgfqpoint{0.155275in}{0.317609in}}%
\pgfpathlineto{\pgfqpoint{0.158890in}{0.306991in}}%
\pgfpathlineto{\pgfqpoint{0.162506in}{0.296690in}}%
\pgfpathlineto{\pgfqpoint{0.166121in}{0.286696in}}%
\pgfpathlineto{\pgfqpoint{0.169736in}{0.277001in}}%
\pgfpathlineto{\pgfqpoint{0.173351in}{0.267595in}}%
\pgfpathlineto{\pgfqpoint{0.176967in}{0.258470in}}%
\pgfpathlineto{\pgfqpoint{0.180582in}{0.249617in}}%
\pgfpathlineto{\pgfqpoint{0.184197in}{0.241028in}}%
\pgfpathlineto{\pgfqpoint{0.187812in}{0.232696in}}%
\pgfpathlineto{\pgfqpoint{0.191428in}{0.224612in}}%
\pgfpathlineto{\pgfqpoint{0.195043in}{0.216770in}}%
\pgfpathlineto{\pgfqpoint{0.198658in}{0.209162in}}%
\pgfpathlineto{\pgfqpoint{0.202273in}{0.201781in}}%
\pgfpathlineto{\pgfqpoint{0.205889in}{0.194620in}}%
\pgfpathlineto{\pgfqpoint{0.209504in}{0.187673in}}%
\pgfpathlineto{\pgfqpoint{0.213119in}{0.180933in}}%
\pgfpathlineto{\pgfqpoint{0.216734in}{0.174395in}}%
\pgfpathlineto{\pgfqpoint{0.220350in}{0.168051in}}%
\pgfpathlineto{\pgfqpoint{0.223965in}{0.161897in}}%
\pgfpathlineto{\pgfqpoint{0.227580in}{0.155927in}}%
\pgfpathlineto{\pgfqpoint{0.231195in}{0.150135in}}%
\pgfpathlineto{\pgfqpoint{0.234811in}{0.144516in}}%
\pgfpathlineto{\pgfqpoint{0.238426in}{0.139064in}}%
\pgfpathlineto{\pgfqpoint{0.242041in}{0.133775in}}%
\pgfpathlineto{\pgfqpoint{0.245656in}{0.128644in}}%
\pgfpathlineto{\pgfqpoint{0.249272in}{0.123667in}}%
\pgfpathlineto{\pgfqpoint{0.252887in}{0.118837in}}%
\pgfpathlineto{\pgfqpoint{0.256502in}{0.114152in}}%
\pgfpathlineto{\pgfqpoint{0.260117in}{0.109607in}}%
\pgfpathlineto{\pgfqpoint{0.263733in}{0.105198in}}%
\pgfpathlineto{\pgfqpoint{0.267348in}{0.100920in}}%
\pgfpathlineto{\pgfqpoint{0.270963in}{0.096769in}}%
\pgfpathlineto{\pgfqpoint{0.274578in}{0.092743in}}%
\pgfpathlineto{\pgfqpoint{0.278194in}{0.088837in}}%
\pgfpathlineto{\pgfqpoint{0.281809in}{0.085047in}}%
\pgfpathlineto{\pgfqpoint{0.285424in}{0.081371in}}%
\pgfpathlineto{\pgfqpoint{0.289039in}{0.077804in}}%
\pgfpathlineto{\pgfqpoint{0.292655in}{0.074344in}}%
\pgfpathlineto{\pgfqpoint{0.296270in}{0.070987in}}%
\pgfpathlineto{\pgfqpoint{0.299885in}{0.067730in}}%
\pgfpathlineto{\pgfqpoint{0.303500in}{0.064570in}}%
\pgfpathlineto{\pgfqpoint{0.307116in}{0.061505in}}%
\pgfpathlineto{\pgfqpoint{0.310731in}{0.058531in}}%
\pgfpathlineto{\pgfqpoint{0.314346in}{0.055646in}}%
\pgfpathlineto{\pgfqpoint{0.317961in}{0.052848in}}%
\pgfpathlineto{\pgfqpoint{0.321577in}{0.050132in}}%
\pgfpathlineto{\pgfqpoint{0.325192in}{0.047498in}}%
\pgfpathlineto{\pgfqpoint{0.328807in}{0.044942in}}%
\pgfpathlineto{\pgfqpoint{0.332422in}{0.042463in}}%
\pgfpathlineto{\pgfqpoint{0.336038in}{0.040057in}}%
\pgfpathlineto{\pgfqpoint{0.339653in}{0.037724in}}%
\pgfpathlineto{\pgfqpoint{0.343268in}{0.035460in}}%
\pgfpathlineto{\pgfqpoint{0.346883in}{0.033264in}}%
\pgfpathlineto{\pgfqpoint{0.350499in}{0.031133in}}%
\pgfpathlineto{\pgfqpoint{0.354114in}{0.029066in}}%
\pgfpathlineto{\pgfqpoint{0.357729in}{0.027060in}}%
\pgfpathlineto{\pgfqpoint{0.361344in}{0.025114in}}%
\pgfpathlineto{\pgfqpoint{0.364960in}{0.023227in}}%
\pgfpathlineto{\pgfqpoint{0.368575in}{0.021396in}}%
\pgfpathlineto{\pgfqpoint{0.372190in}{0.019619in}}%
\pgfpathlineto{\pgfqpoint{0.375805in}{0.017896in}}%
\pgfusepath{stroke}%
\end{pgfscope}%
\begin{pgfscope}%
\pgfsetrectcap%
\pgfsetmiterjoin%
\pgfsetlinewidth{1.254687pt}%
\definecolor{currentstroke}{rgb}{0.150000,0.150000,0.150000}%
\pgfsetstrokecolor{currentstroke}%
\pgfsetdash{}{0pt}%
\pgfpathmoveto{\pgfqpoint{0.000000in}{0.000000in}}%
\pgfpathlineto{\pgfqpoint{0.000000in}{0.393701in}}%
\pgfusepath{stroke}%
\end{pgfscope}%
\begin{pgfscope}%
\pgfsetrectcap%
\pgfsetmiterjoin%
\pgfsetlinewidth{1.254687pt}%
\definecolor{currentstroke}{rgb}{0.150000,0.150000,0.150000}%
\pgfsetstrokecolor{currentstroke}%
\pgfsetdash{}{0pt}%
\pgfpathmoveto{\pgfqpoint{0.000000in}{0.000000in}}%
\pgfpathlineto{\pgfqpoint{0.393701in}{0.000000in}}%
\pgfusepath{stroke}%
\end{pgfscope}%
\end{pgfpicture}%
\makeatother%
\endgroup%
\tabularnewline
  \midrule

  T &

\begingroup%
\makeatletter%
\begin{pgfpicture}%
\pgfpathrectangle{\pgfpointorigin}{\pgfqpoint{0.393701in}{0.393701in}}%
\pgfusepath{use as bounding box, clip}%
\begin{pgfscope}%
\pgfsetbuttcap%
\pgfsetmiterjoin%
\pgfsetlinewidth{0.000000pt}%
\definecolor{currentstroke}{rgb}{0.000000,0.000000,0.000000}%
\pgfsetstrokecolor{currentstroke}%
\pgfsetstrokeopacity{0.000000}%
\pgfsetdash{}{0pt}%
\pgfpathmoveto{\pgfqpoint{0.000000in}{0.000000in}}%
\pgfpathlineto{\pgfqpoint{0.393701in}{0.000000in}}%
\pgfpathlineto{\pgfqpoint{0.393701in}{0.393701in}}%
\pgfpathlineto{\pgfqpoint{0.000000in}{0.393701in}}%
\pgfpathclose%
\pgfusepath{}%
\end{pgfscope}%
\begin{pgfscope}%
\pgfsetbuttcap%
\pgfsetmiterjoin%
\pgfsetlinewidth{0.000000pt}%
\definecolor{currentstroke}{rgb}{0.000000,0.000000,0.000000}%
\pgfsetstrokecolor{currentstroke}%
\pgfsetstrokeopacity{0.000000}%
\pgfsetdash{}{0pt}%
\pgfpathmoveto{\pgfqpoint{0.000000in}{0.000000in}}%
\pgfpathlineto{\pgfqpoint{0.393701in}{0.000000in}}%
\pgfpathlineto{\pgfqpoint{0.393701in}{0.393701in}}%
\pgfpathlineto{\pgfqpoint{0.000000in}{0.393701in}}%
\pgfpathclose%
\pgfusepath{}%
\end{pgfscope}%
\begin{pgfscope}%
\pgfpathrectangle{\pgfqpoint{0.000000in}{0.000000in}}{\pgfqpoint{0.393701in}{0.393701in}}%
\pgfusepath{clip}%
\pgfsetroundcap%
\pgfsetroundjoin%
\pgfsetlinewidth{1.003750pt}%
\definecolor{currentstroke}{rgb}{0.121569,0.466667,0.705882}%
\pgfsetstrokecolor{currentstroke}%
\pgfsetdash{}{0pt}%
\pgfpathmoveto{\pgfqpoint{0.017896in}{0.017896in}}%
\pgfpathlineto{\pgfqpoint{0.043075in}{0.027763in}}%
\pgfpathlineto{\pgfqpoint{0.064658in}{0.039574in}}%
\pgfpathlineto{\pgfqpoint{0.082643in}{0.053110in}}%
\pgfpathlineto{\pgfqpoint{0.097031in}{0.067582in}}%
\pgfpathlineto{\pgfqpoint{0.109621in}{0.084100in}}%
\pgfpathlineto{\pgfqpoint{0.122211in}{0.105725in}}%
\pgfpathlineto{\pgfqpoint{0.133002in}{0.129905in}}%
\pgfpathlineto{\pgfqpoint{0.143793in}{0.161145in}}%
\pgfpathlineto{\pgfqpoint{0.154585in}{0.201397in}}%
\pgfpathlineto{\pgfqpoint{0.167175in}{0.260857in}}%
\pgfpathlineto{\pgfqpoint{0.186959in}{0.357675in}}%
\pgfpathlineto{\pgfqpoint{0.192354in}{0.372014in}}%
\pgfpathlineto{\pgfqpoint{0.195951in}{0.375805in}}%
\pgfpathlineto{\pgfqpoint{0.199548in}{0.374532in}}%
\pgfpathlineto{\pgfqpoint{0.203145in}{0.368308in}}%
\pgfpathlineto{\pgfqpoint{0.208541in}{0.350963in}}%
\pgfpathlineto{\pgfqpoint{0.217534in}{0.308385in}}%
\pgfpathlineto{\pgfqpoint{0.239116in}{0.201397in}}%
\pgfpathlineto{\pgfqpoint{0.251706in}{0.155363in}}%
\pgfpathlineto{\pgfqpoint{0.264296in}{0.121156in}}%
\pgfpathlineto{\pgfqpoint{0.276886in}{0.095713in}}%
\pgfpathlineto{\pgfqpoint{0.291274in}{0.074142in}}%
\pgfpathlineto{\pgfqpoint{0.305662in}{0.058083in}}%
\pgfpathlineto{\pgfqpoint{0.321849in}{0.044496in}}%
\pgfpathlineto{\pgfqpoint{0.341633in}{0.032219in}}%
\pgfpathlineto{\pgfqpoint{0.365014in}{0.021721in}}%
\pgfpathlineto{\pgfqpoint{0.375806in}{0.017896in}}%
\pgfpathlineto{\pgfqpoint{0.375806in}{0.017896in}}%
\pgfusepath{stroke}%
\end{pgfscope}%
\begin{pgfscope}%
\pgfsetrectcap%
\pgfsetmiterjoin%
\pgfsetlinewidth{1.254687pt}%
\definecolor{currentstroke}{rgb}{0.150000,0.150000,0.150000}%
\pgfsetstrokecolor{currentstroke}%
\pgfsetdash{}{0pt}%
\pgfpathmoveto{\pgfqpoint{0.196851in}{0.000000in}}%
\pgfpathlineto{\pgfqpoint{0.196851in}{0.393701in}}%
\pgfusepath{stroke}%
\end{pgfscope}%
\begin{pgfscope}%
\pgfsetrectcap%
\pgfsetmiterjoin%
\pgfsetlinewidth{1.254687pt}%
\definecolor{currentstroke}{rgb}{0.150000,0.150000,0.150000}%
\pgfsetstrokecolor{currentstroke}%
\pgfsetdash{}{0pt}%
\pgfpathmoveto{\pgfqpoint{0.000000in}{0.000000in}}%
\pgfpathlineto{\pgfqpoint{0.393701in}{0.000000in}}%
\pgfusepath{stroke}%
\end{pgfscope}%
\end{pgfpicture}%
\makeatother%
\endgroup%
 &

\begingroup%
\makeatletter%
\begin{pgfpicture}%
\pgfpathrectangle{\pgfpointorigin}{\pgfqpoint{0.393701in}{0.393701in}}%
\pgfusepath{use as bounding box, clip}%
\begin{pgfscope}%
\pgfsetbuttcap%
\pgfsetmiterjoin%
\pgfsetlinewidth{0.000000pt}%
\definecolor{currentstroke}{rgb}{0.000000,0.000000,0.000000}%
\pgfsetstrokecolor{currentstroke}%
\pgfsetstrokeopacity{0.000000}%
\pgfsetdash{}{0pt}%
\pgfpathmoveto{\pgfqpoint{0.000000in}{0.000000in}}%
\pgfpathlineto{\pgfqpoint{0.393701in}{0.000000in}}%
\pgfpathlineto{\pgfqpoint{0.393701in}{0.393701in}}%
\pgfpathlineto{\pgfqpoint{0.000000in}{0.393701in}}%
\pgfpathclose%
\pgfusepath{}%
\end{pgfscope}%
\begin{pgfscope}%
\pgfsetbuttcap%
\pgfsetmiterjoin%
\pgfsetlinewidth{0.000000pt}%
\definecolor{currentstroke}{rgb}{0.000000,0.000000,0.000000}%
\pgfsetstrokecolor{currentstroke}%
\pgfsetstrokeopacity{0.000000}%
\pgfsetdash{}{0pt}%
\pgfpathmoveto{\pgfqpoint{0.000000in}{0.000000in}}%
\pgfpathlineto{\pgfqpoint{0.393701in}{0.000000in}}%
\pgfpathlineto{\pgfqpoint{0.393701in}{0.393701in}}%
\pgfpathlineto{\pgfqpoint{0.000000in}{0.393701in}}%
\pgfpathclose%
\pgfusepath{}%
\end{pgfscope}%
\begin{pgfscope}%
\pgfpathrectangle{\pgfqpoint{0.000000in}{0.000000in}}{\pgfqpoint{0.393701in}{0.393701in}}%
\pgfusepath{clip}%
\pgfsetroundcap%
\pgfsetroundjoin%
\pgfsetlinewidth{1.003750pt}%
\definecolor{currentstroke}{rgb}{0.121569,0.466667,0.705882}%
\pgfsetstrokecolor{currentstroke}%
\pgfsetdash{}{0pt}%
\pgfpathmoveto{\pgfqpoint{0.017896in}{0.375806in}}%
\pgfpathlineto{\pgfqpoint{0.021511in}{0.375161in}}%
\pgfpathlineto{\pgfqpoint{0.025126in}{0.373241in}}%
\pgfpathlineto{\pgfqpoint{0.028741in}{0.370091in}}%
\pgfpathlineto{\pgfqpoint{0.032357in}{0.365781in}}%
\pgfpathlineto{\pgfqpoint{0.035972in}{0.360404in}}%
\pgfpathlineto{\pgfqpoint{0.039587in}{0.354069in}}%
\pgfpathlineto{\pgfqpoint{0.043202in}{0.346899in}}%
\pgfpathlineto{\pgfqpoint{0.046818in}{0.339022in}}%
\pgfpathlineto{\pgfqpoint{0.050433in}{0.330566in}}%
\pgfpathlineto{\pgfqpoint{0.054048in}{0.321659in}}%
\pgfpathlineto{\pgfqpoint{0.057663in}{0.312419in}}%
\pgfpathlineto{\pgfqpoint{0.061279in}{0.302956in}}%
\pgfpathlineto{\pgfqpoint{0.064894in}{0.293368in}}%
\pgfpathlineto{\pgfqpoint{0.068509in}{0.283742in}}%
\pgfpathlineto{\pgfqpoint{0.072124in}{0.274154in}}%
\pgfpathlineto{\pgfqpoint{0.075740in}{0.264665in}}%
\pgfpathlineto{\pgfqpoint{0.079355in}{0.255327in}}%
\pgfpathlineto{\pgfqpoint{0.082970in}{0.246182in}}%
\pgfpathlineto{\pgfqpoint{0.086585in}{0.237263in}}%
\pgfpathlineto{\pgfqpoint{0.090201in}{0.228593in}}%
\pgfpathlineto{\pgfqpoint{0.093816in}{0.220191in}}%
\pgfpathlineto{\pgfqpoint{0.097431in}{0.212068in}}%
\pgfpathlineto{\pgfqpoint{0.101046in}{0.204232in}}%
\pgfpathlineto{\pgfqpoint{0.104662in}{0.196684in}}%
\pgfpathlineto{\pgfqpoint{0.108277in}{0.189424in}}%
\pgfpathlineto{\pgfqpoint{0.111892in}{0.182450in}}%
\pgfpathlineto{\pgfqpoint{0.115507in}{0.175757in}}%
\pgfpathlineto{\pgfqpoint{0.119123in}{0.169337in}}%
\pgfpathlineto{\pgfqpoint{0.122738in}{0.163184in}}%
\pgfpathlineto{\pgfqpoint{0.126353in}{0.157288in}}%
\pgfpathlineto{\pgfqpoint{0.129968in}{0.151640in}}%
\pgfpathlineto{\pgfqpoint{0.133584in}{0.146231in}}%
\pgfpathlineto{\pgfqpoint{0.137199in}{0.141051in}}%
\pgfpathlineto{\pgfqpoint{0.140814in}{0.136091in}}%
\pgfpathlineto{\pgfqpoint{0.144429in}{0.131340in}}%
\pgfpathlineto{\pgfqpoint{0.148045in}{0.126790in}}%
\pgfpathlineto{\pgfqpoint{0.151660in}{0.122430in}}%
\pgfpathlineto{\pgfqpoint{0.155275in}{0.118253in}}%
\pgfpathlineto{\pgfqpoint{0.158890in}{0.114250in}}%
\pgfpathlineto{\pgfqpoint{0.162506in}{0.110411in}}%
\pgfpathlineto{\pgfqpoint{0.166121in}{0.106730in}}%
\pgfpathlineto{\pgfqpoint{0.169736in}{0.103199in}}%
\pgfpathlineto{\pgfqpoint{0.173351in}{0.099809in}}%
\pgfpathlineto{\pgfqpoint{0.176967in}{0.096555in}}%
\pgfpathlineto{\pgfqpoint{0.180582in}{0.093430in}}%
\pgfpathlineto{\pgfqpoint{0.184197in}{0.090427in}}%
\pgfpathlineto{\pgfqpoint{0.187812in}{0.087540in}}%
\pgfpathlineto{\pgfqpoint{0.191428in}{0.084764in}}%
\pgfpathlineto{\pgfqpoint{0.195043in}{0.082094in}}%
\pgfpathlineto{\pgfqpoint{0.198658in}{0.079524in}}%
\pgfpathlineto{\pgfqpoint{0.202273in}{0.077050in}}%
\pgfpathlineto{\pgfqpoint{0.205889in}{0.074667in}}%
\pgfpathlineto{\pgfqpoint{0.209504in}{0.072370in}}%
\pgfpathlineto{\pgfqpoint{0.213119in}{0.070157in}}%
\pgfpathlineto{\pgfqpoint{0.216734in}{0.068022in}}%
\pgfpathlineto{\pgfqpoint{0.220350in}{0.065962in}}%
\pgfpathlineto{\pgfqpoint{0.223965in}{0.063974in}}%
\pgfpathlineto{\pgfqpoint{0.227580in}{0.062055in}}%
\pgfpathlineto{\pgfqpoint{0.231195in}{0.060201in}}%
\pgfpathlineto{\pgfqpoint{0.234811in}{0.058409in}}%
\pgfpathlineto{\pgfqpoint{0.238426in}{0.056677in}}%
\pgfpathlineto{\pgfqpoint{0.242041in}{0.055002in}}%
\pgfpathlineto{\pgfqpoint{0.245656in}{0.053382in}}%
\pgfpathlineto{\pgfqpoint{0.249272in}{0.051814in}}%
\pgfpathlineto{\pgfqpoint{0.252887in}{0.050296in}}%
\pgfpathlineto{\pgfqpoint{0.256502in}{0.048825in}}%
\pgfpathlineto{\pgfqpoint{0.260117in}{0.047401in}}%
\pgfpathlineto{\pgfqpoint{0.263733in}{0.046021in}}%
\pgfpathlineto{\pgfqpoint{0.267348in}{0.044682in}}%
\pgfpathlineto{\pgfqpoint{0.270963in}{0.043384in}}%
\pgfpathlineto{\pgfqpoint{0.274578in}{0.042125in}}%
\pgfpathlineto{\pgfqpoint{0.278194in}{0.040903in}}%
\pgfpathlineto{\pgfqpoint{0.281809in}{0.039717in}}%
\pgfpathlineto{\pgfqpoint{0.285424in}{0.038565in}}%
\pgfpathlineto{\pgfqpoint{0.289039in}{0.037446in}}%
\pgfpathlineto{\pgfqpoint{0.292655in}{0.036359in}}%
\pgfpathlineto{\pgfqpoint{0.296270in}{0.035302in}}%
\pgfpathlineto{\pgfqpoint{0.299885in}{0.034275in}}%
\pgfpathlineto{\pgfqpoint{0.303500in}{0.033276in}}%
\pgfpathlineto{\pgfqpoint{0.307116in}{0.032304in}}%
\pgfpathlineto{\pgfqpoint{0.310731in}{0.031358in}}%
\pgfpathlineto{\pgfqpoint{0.314346in}{0.030437in}}%
\pgfpathlineto{\pgfqpoint{0.317961in}{0.029541in}}%
\pgfpathlineto{\pgfqpoint{0.321577in}{0.028668in}}%
\pgfpathlineto{\pgfqpoint{0.325192in}{0.027818in}}%
\pgfpathlineto{\pgfqpoint{0.328807in}{0.026989in}}%
\pgfpathlineto{\pgfqpoint{0.332422in}{0.026182in}}%
\pgfpathlineto{\pgfqpoint{0.336038in}{0.025395in}}%
\pgfpathlineto{\pgfqpoint{0.339653in}{0.024627in}}%
\pgfpathlineto{\pgfqpoint{0.343268in}{0.023879in}}%
\pgfpathlineto{\pgfqpoint{0.346883in}{0.023148in}}%
\pgfpathlineto{\pgfqpoint{0.350499in}{0.022436in}}%
\pgfpathlineto{\pgfqpoint{0.354114in}{0.021740in}}%
\pgfpathlineto{\pgfqpoint{0.357729in}{0.021061in}}%
\pgfpathlineto{\pgfqpoint{0.361344in}{0.020398in}}%
\pgfpathlineto{\pgfqpoint{0.364960in}{0.019750in}}%
\pgfpathlineto{\pgfqpoint{0.368575in}{0.019118in}}%
\pgfpathlineto{\pgfqpoint{0.372190in}{0.018500in}}%
\pgfpathlineto{\pgfqpoint{0.375805in}{0.017895in}}%
\pgfusepath{stroke}%
\end{pgfscope}%
\begin{pgfscope}%
\pgfsetrectcap%
\pgfsetmiterjoin%
\pgfsetlinewidth{1.254687pt}%
\definecolor{currentstroke}{rgb}{0.150000,0.150000,0.150000}%
\pgfsetstrokecolor{currentstroke}%
\pgfsetdash{}{0pt}%
\pgfpathmoveto{\pgfqpoint{0.000000in}{-0.000000in}}%
\pgfpathlineto{\pgfqpoint{0.000000in}{0.393701in}}%
\pgfusepath{stroke}%
\end{pgfscope}%
\begin{pgfscope}%
\pgfsetrectcap%
\pgfsetmiterjoin%
\pgfsetlinewidth{1.254687pt}%
\definecolor{currentstroke}{rgb}{0.150000,0.150000,0.150000}%
\pgfsetstrokecolor{currentstroke}%
\pgfsetdash{}{0pt}%
\pgfpathmoveto{\pgfqpoint{0.000000in}{0.000000in}}%
\pgfpathlineto{\pgfqpoint{0.393701in}{0.000000in}}%
\pgfusepath{stroke}%
\end{pgfscope}%
\end{pgfpicture}%
\makeatother%
\endgroup%
 &

\begingroup%
\makeatletter%
\begin{pgfpicture}%
\pgfpathrectangle{\pgfpointorigin}{\pgfqpoint{0.393701in}{0.393701in}}%
\pgfusepath{use as bounding box, clip}%
\begin{pgfscope}%
\pgfsetbuttcap%
\pgfsetmiterjoin%
\pgfsetlinewidth{0.000000pt}%
\definecolor{currentstroke}{rgb}{0.000000,0.000000,0.000000}%
\pgfsetstrokecolor{currentstroke}%
\pgfsetstrokeopacity{0.000000}%
\pgfsetdash{}{0pt}%
\pgfpathmoveto{\pgfqpoint{0.000000in}{0.000000in}}%
\pgfpathlineto{\pgfqpoint{0.393701in}{0.000000in}}%
\pgfpathlineto{\pgfqpoint{0.393701in}{0.393701in}}%
\pgfpathlineto{\pgfqpoint{0.000000in}{0.393701in}}%
\pgfpathclose%
\pgfusepath{}%
\end{pgfscope}%
\begin{pgfscope}%
\pgfsetbuttcap%
\pgfsetmiterjoin%
\pgfsetlinewidth{0.000000pt}%
\definecolor{currentstroke}{rgb}{0.000000,0.000000,0.000000}%
\pgfsetstrokecolor{currentstroke}%
\pgfsetstrokeopacity{0.000000}%
\pgfsetdash{}{0pt}%
\pgfpathmoveto{\pgfqpoint{0.000000in}{0.000000in}}%
\pgfpathlineto{\pgfqpoint{0.393701in}{0.000000in}}%
\pgfpathlineto{\pgfqpoint{0.393701in}{0.393701in}}%
\pgfpathlineto{\pgfqpoint{0.000000in}{0.393701in}}%
\pgfpathclose%
\pgfusepath{}%
\end{pgfscope}%
\begin{pgfscope}%
\pgfpathrectangle{\pgfqpoint{0.000000in}{0.000000in}}{\pgfqpoint{0.393701in}{0.393701in}}%
\pgfusepath{clip}%
\pgfsetroundcap%
\pgfsetroundjoin%
\pgfsetlinewidth{1.003750pt}%
\definecolor{currentstroke}{rgb}{0.121569,0.466667,0.705882}%
\pgfsetstrokecolor{currentstroke}%
\pgfsetdash{}{0pt}%
\pgfpathmoveto{\pgfqpoint{0.017896in}{0.017896in}}%
\pgfpathlineto{\pgfqpoint{0.057463in}{0.025926in}}%
\pgfpathlineto{\pgfqpoint{0.088039in}{0.035154in}}%
\pgfpathlineto{\pgfqpoint{0.111420in}{0.045205in}}%
\pgfpathlineto{\pgfqpoint{0.131204in}{0.056954in}}%
\pgfpathlineto{\pgfqpoint{0.147391in}{0.069948in}}%
\pgfpathlineto{\pgfqpoint{0.161779in}{0.085298in}}%
\pgfpathlineto{\pgfqpoint{0.174369in}{0.102968in}}%
\pgfpathlineto{\pgfqpoint{0.186959in}{0.126285in}}%
\pgfpathlineto{\pgfqpoint{0.197750in}{0.152506in}}%
\pgfpathlineto{\pgfqpoint{0.208541in}{0.186415in}}%
\pgfpathlineto{\pgfqpoint{0.221131in}{0.237838in}}%
\pgfpathlineto{\pgfqpoint{0.249908in}{0.367812in}}%
\pgfpathlineto{\pgfqpoint{0.255303in}{0.375601in}}%
\pgfpathlineto{\pgfqpoint{0.258900in}{0.374784in}}%
\pgfpathlineto{\pgfqpoint{0.262497in}{0.369182in}}%
\pgfpathlineto{\pgfqpoint{0.267893in}{0.352964in}}%
\pgfpathlineto{\pgfqpoint{0.276886in}{0.312401in}}%
\pgfpathlineto{\pgfqpoint{0.298468in}{0.209237in}}%
\pgfpathlineto{\pgfqpoint{0.311058in}{0.164672in}}%
\pgfpathlineto{\pgfqpoint{0.323648in}{0.131546in}}%
\pgfpathlineto{\pgfqpoint{0.336238in}{0.106913in}}%
\pgfpathlineto{\pgfqpoint{0.350626in}{0.086038in}}%
\pgfpathlineto{\pgfqpoint{0.365014in}{0.070506in}}%
\pgfpathlineto{\pgfqpoint{0.375806in}{0.061338in}}%
\pgfpathlineto{\pgfqpoint{0.375806in}{0.061338in}}%
\pgfusepath{stroke}%
\end{pgfscope}%
\begin{pgfscope}%
\pgfsetrectcap%
\pgfsetmiterjoin%
\pgfsetlinewidth{1.254687pt}%
\definecolor{currentstroke}{rgb}{0.150000,0.150000,0.150000}%
\pgfsetstrokecolor{currentstroke}%
\pgfsetdash{}{0pt}%
\pgfpathmoveto{\pgfqpoint{0.196851in}{0.000000in}}%
\pgfpathlineto{\pgfqpoint{0.196851in}{0.393701in}}%
\pgfusepath{stroke}%
\end{pgfscope}%
\begin{pgfscope}%
\pgfsetrectcap%
\pgfsetmiterjoin%
\pgfsetlinewidth{1.254687pt}%
\definecolor{currentstroke}{rgb}{0.150000,0.150000,0.150000}%
\pgfsetstrokecolor{currentstroke}%
\pgfsetdash{}{0pt}%
\pgfpathmoveto{\pgfqpoint{0.000000in}{0.000000in}}%
\pgfpathlineto{\pgfqpoint{0.393701in}{0.000000in}}%
\pgfusepath{stroke}%
\end{pgfscope}%
\end{pgfpicture}%
\makeatother%
\endgroup%
 &

\begingroup%
\makeatletter%
\begin{pgfpicture}%
\pgfpathrectangle{\pgfpointorigin}{\pgfqpoint{0.393701in}{0.393701in}}%
\pgfusepath{use as bounding box, clip}%
\begin{pgfscope}%
\pgfsetbuttcap%
\pgfsetmiterjoin%
\pgfsetlinewidth{0.000000pt}%
\definecolor{currentstroke}{rgb}{0.000000,0.000000,0.000000}%
\pgfsetstrokecolor{currentstroke}%
\pgfsetstrokeopacity{0.000000}%
\pgfsetdash{}{0pt}%
\pgfpathmoveto{\pgfqpoint{0.000000in}{0.000000in}}%
\pgfpathlineto{\pgfqpoint{0.393701in}{0.000000in}}%
\pgfpathlineto{\pgfqpoint{0.393701in}{0.393701in}}%
\pgfpathlineto{\pgfqpoint{0.000000in}{0.393701in}}%
\pgfpathclose%
\pgfusepath{}%
\end{pgfscope}%
\begin{pgfscope}%
\pgfsetbuttcap%
\pgfsetmiterjoin%
\pgfsetlinewidth{0.000000pt}%
\definecolor{currentstroke}{rgb}{0.000000,0.000000,0.000000}%
\pgfsetstrokecolor{currentstroke}%
\pgfsetstrokeopacity{0.000000}%
\pgfsetdash{}{0pt}%
\pgfpathmoveto{\pgfqpoint{0.000000in}{0.000000in}}%
\pgfpathlineto{\pgfqpoint{0.393701in}{0.000000in}}%
\pgfpathlineto{\pgfqpoint{0.393701in}{0.393701in}}%
\pgfpathlineto{\pgfqpoint{0.000000in}{0.393701in}}%
\pgfpathclose%
\pgfusepath{}%
\end{pgfscope}%
\begin{pgfscope}%
\pgfpathrectangle{\pgfqpoint{0.000000in}{0.000000in}}{\pgfqpoint{0.393701in}{0.393701in}}%
\pgfusepath{clip}%
\pgfsetroundcap%
\pgfsetroundjoin%
\pgfsetlinewidth{1.003750pt}%
\definecolor{currentstroke}{rgb}{0.839216,0.152941,0.156863}%
\pgfsetstrokecolor{currentstroke}%
\pgfsetdash{}{0pt}%
\pgfpathmoveto{\pgfqpoint{0.017896in}{0.118845in}}%
\pgfpathlineto{\pgfqpoint{0.021511in}{0.124515in}}%
\pgfpathlineto{\pgfqpoint{0.025126in}{0.130435in}}%
\pgfpathlineto{\pgfqpoint{0.028741in}{0.136617in}}%
\pgfpathlineto{\pgfqpoint{0.032357in}{0.143071in}}%
\pgfpathlineto{\pgfqpoint{0.035972in}{0.149807in}}%
\pgfpathlineto{\pgfqpoint{0.039587in}{0.156834in}}%
\pgfpathlineto{\pgfqpoint{0.043202in}{0.164160in}}%
\pgfpathlineto{\pgfqpoint{0.046818in}{0.171794in}}%
\pgfpathlineto{\pgfqpoint{0.050433in}{0.179740in}}%
\pgfpathlineto{\pgfqpoint{0.054048in}{0.188002in}}%
\pgfpathlineto{\pgfqpoint{0.057663in}{0.196580in}}%
\pgfpathlineto{\pgfqpoint{0.061279in}{0.205471in}}%
\pgfpathlineto{\pgfqpoint{0.064894in}{0.214668in}}%
\pgfpathlineto{\pgfqpoint{0.068509in}{0.224158in}}%
\pgfpathlineto{\pgfqpoint{0.072124in}{0.233921in}}%
\pgfpathlineto{\pgfqpoint{0.075740in}{0.243931in}}%
\pgfpathlineto{\pgfqpoint{0.079355in}{0.254151in}}%
\pgfpathlineto{\pgfqpoint{0.082970in}{0.264538in}}%
\pgfpathlineto{\pgfqpoint{0.086585in}{0.275034in}}%
\pgfpathlineto{\pgfqpoint{0.090201in}{0.285570in}}%
\pgfpathlineto{\pgfqpoint{0.093816in}{0.296064in}}%
\pgfpathlineto{\pgfqpoint{0.097431in}{0.306423in}}%
\pgfpathlineto{\pgfqpoint{0.101046in}{0.316537in}}%
\pgfpathlineto{\pgfqpoint{0.104662in}{0.326287in}}%
\pgfpathlineto{\pgfqpoint{0.108277in}{0.335542in}}%
\pgfpathlineto{\pgfqpoint{0.111892in}{0.344165in}}%
\pgfpathlineto{\pgfqpoint{0.115507in}{0.352013in}}%
\pgfpathlineto{\pgfqpoint{0.119123in}{0.358947in}}%
\pgfpathlineto{\pgfqpoint{0.122738in}{0.364833in}}%
\pgfpathlineto{\pgfqpoint{0.126353in}{0.369551in}}%
\pgfpathlineto{\pgfqpoint{0.129968in}{0.372999in}}%
\pgfpathlineto{\pgfqpoint{0.133584in}{0.375100in}}%
\pgfpathlineto{\pgfqpoint{0.137199in}{0.375806in}}%
\pgfpathlineto{\pgfqpoint{0.140814in}{0.375100in}}%
\pgfpathlineto{\pgfqpoint{0.144429in}{0.372999in}}%
\pgfpathlineto{\pgfqpoint{0.148045in}{0.369551in}}%
\pgfpathlineto{\pgfqpoint{0.151660in}{0.364833in}}%
\pgfpathlineto{\pgfqpoint{0.155275in}{0.358947in}}%
\pgfpathlineto{\pgfqpoint{0.158890in}{0.352013in}}%
\pgfpathlineto{\pgfqpoint{0.162506in}{0.344165in}}%
\pgfpathlineto{\pgfqpoint{0.166121in}{0.335542in}}%
\pgfpathlineto{\pgfqpoint{0.169736in}{0.326287in}}%
\pgfpathlineto{\pgfqpoint{0.173351in}{0.316537in}}%
\pgfpathlineto{\pgfqpoint{0.176967in}{0.306423in}}%
\pgfpathlineto{\pgfqpoint{0.180582in}{0.296064in}}%
\pgfpathlineto{\pgfqpoint{0.184197in}{0.285570in}}%
\pgfpathlineto{\pgfqpoint{0.187812in}{0.275034in}}%
\pgfpathlineto{\pgfqpoint{0.191428in}{0.264538in}}%
\pgfpathlineto{\pgfqpoint{0.195043in}{0.254151in}}%
\pgfpathlineto{\pgfqpoint{0.198658in}{0.243931in}}%
\pgfpathlineto{\pgfqpoint{0.202273in}{0.233921in}}%
\pgfpathlineto{\pgfqpoint{0.205889in}{0.224158in}}%
\pgfpathlineto{\pgfqpoint{0.209504in}{0.214668in}}%
\pgfpathlineto{\pgfqpoint{0.213119in}{0.205471in}}%
\pgfpathlineto{\pgfqpoint{0.216734in}{0.196580in}}%
\pgfpathlineto{\pgfqpoint{0.220350in}{0.188002in}}%
\pgfpathlineto{\pgfqpoint{0.223965in}{0.179740in}}%
\pgfpathlineto{\pgfqpoint{0.227580in}{0.171794in}}%
\pgfpathlineto{\pgfqpoint{0.231195in}{0.164160in}}%
\pgfpathlineto{\pgfqpoint{0.234811in}{0.156834in}}%
\pgfpathlineto{\pgfqpoint{0.238426in}{0.149807in}}%
\pgfpathlineto{\pgfqpoint{0.242041in}{0.143071in}}%
\pgfpathlineto{\pgfqpoint{0.245656in}{0.136617in}}%
\pgfpathlineto{\pgfqpoint{0.249272in}{0.130435in}}%
\pgfpathlineto{\pgfqpoint{0.252887in}{0.124515in}}%
\pgfpathlineto{\pgfqpoint{0.256502in}{0.118845in}}%
\pgfpathlineto{\pgfqpoint{0.260117in}{0.113415in}}%
\pgfpathlineto{\pgfqpoint{0.263733in}{0.108215in}}%
\pgfpathlineto{\pgfqpoint{0.267348in}{0.103235in}}%
\pgfpathlineto{\pgfqpoint{0.270963in}{0.098463in}}%
\pgfpathlineto{\pgfqpoint{0.274578in}{0.093891in}}%
\pgfpathlineto{\pgfqpoint{0.278194in}{0.089509in}}%
\pgfpathlineto{\pgfqpoint{0.281809in}{0.085307in}}%
\pgfpathlineto{\pgfqpoint{0.285424in}{0.081278in}}%
\pgfpathlineto{\pgfqpoint{0.289039in}{0.077412in}}%
\pgfpathlineto{\pgfqpoint{0.292655in}{0.073702in}}%
\pgfpathlineto{\pgfqpoint{0.296270in}{0.070140in}}%
\pgfpathlineto{\pgfqpoint{0.299885in}{0.066719in}}%
\pgfpathlineto{\pgfqpoint{0.303500in}{0.063432in}}%
\pgfpathlineto{\pgfqpoint{0.307116in}{0.060272in}}%
\pgfpathlineto{\pgfqpoint{0.310731in}{0.057234in}}%
\pgfpathlineto{\pgfqpoint{0.314346in}{0.054311in}}%
\pgfpathlineto{\pgfqpoint{0.317961in}{0.051498in}}%
\pgfpathlineto{\pgfqpoint{0.321577in}{0.048790in}}%
\pgfpathlineto{\pgfqpoint{0.325192in}{0.046181in}}%
\pgfpathlineto{\pgfqpoint{0.328807in}{0.043668in}}%
\pgfpathlineto{\pgfqpoint{0.332422in}{0.041244in}}%
\pgfpathlineto{\pgfqpoint{0.336038in}{0.038908in}}%
\pgfpathlineto{\pgfqpoint{0.339653in}{0.036653in}}%
\pgfpathlineto{\pgfqpoint{0.343268in}{0.034477in}}%
\pgfpathlineto{\pgfqpoint{0.346883in}{0.032376in}}%
\pgfpathlineto{\pgfqpoint{0.350499in}{0.030347in}}%
\pgfpathlineto{\pgfqpoint{0.354114in}{0.028385in}}%
\pgfpathlineto{\pgfqpoint{0.357729in}{0.026490in}}%
\pgfpathlineto{\pgfqpoint{0.361344in}{0.024656in}}%
\pgfpathlineto{\pgfqpoint{0.364960in}{0.022883in}}%
\pgfpathlineto{\pgfqpoint{0.368575in}{0.021166in}}%
\pgfpathlineto{\pgfqpoint{0.372190in}{0.019505in}}%
\pgfpathlineto{\pgfqpoint{0.375805in}{0.017896in}}%
\pgfusepath{stroke}%
\end{pgfscope}%
\begin{pgfscope}%
\pgfsetrectcap%
\pgfsetmiterjoin%
\pgfsetlinewidth{1.254687pt}%
\definecolor{currentstroke}{rgb}{0.150000,0.150000,0.150000}%
\pgfsetstrokecolor{currentstroke}%
\pgfsetdash{}{0pt}%
\pgfpathmoveto{\pgfqpoint{0.000000in}{0.000000in}}%
\pgfpathlineto{\pgfqpoint{0.000000in}{0.393701in}}%
\pgfusepath{stroke}%
\end{pgfscope}%
\begin{pgfscope}%
\pgfsetrectcap%
\pgfsetmiterjoin%
\pgfsetlinewidth{1.254687pt}%
\definecolor{currentstroke}{rgb}{0.150000,0.150000,0.150000}%
\pgfsetstrokecolor{currentstroke}%
\pgfsetdash{}{0pt}%
\pgfpathmoveto{\pgfqpoint{0.000000in}{0.000000in}}%
\pgfpathlineto{\pgfqpoint{0.393701in}{0.000000in}}%
\pgfusepath{stroke}%
\end{pgfscope}%
\end{pgfpicture}%
\makeatother%
\endgroup%
\tabularnewline
  \midrule

  Lomax &

\begingroup%
\makeatletter%
\begin{pgfpicture}%
\pgfpathrectangle{\pgfpointorigin}{\pgfqpoint{0.393701in}{0.393701in}}%
\pgfusepath{use as bounding box, clip}%
\begin{pgfscope}%
\pgfsetbuttcap%
\pgfsetmiterjoin%
\pgfsetlinewidth{0.000000pt}%
\definecolor{currentstroke}{rgb}{0.000000,0.000000,0.000000}%
\pgfsetstrokecolor{currentstroke}%
\pgfsetstrokeopacity{0.000000}%
\pgfsetdash{}{0pt}%
\pgfpathmoveto{\pgfqpoint{0.000000in}{0.000000in}}%
\pgfpathlineto{\pgfqpoint{0.393701in}{0.000000in}}%
\pgfpathlineto{\pgfqpoint{0.393701in}{0.393701in}}%
\pgfpathlineto{\pgfqpoint{0.000000in}{0.393701in}}%
\pgfpathclose%
\pgfusepath{}%
\end{pgfscope}%
\begin{pgfscope}%
\pgfsetbuttcap%
\pgfsetmiterjoin%
\pgfsetlinewidth{0.000000pt}%
\definecolor{currentstroke}{rgb}{0.000000,0.000000,0.000000}%
\pgfsetstrokecolor{currentstroke}%
\pgfsetstrokeopacity{0.000000}%
\pgfsetdash{}{0pt}%
\pgfpathmoveto{\pgfqpoint{0.000000in}{0.000000in}}%
\pgfpathlineto{\pgfqpoint{0.393701in}{0.000000in}}%
\pgfpathlineto{\pgfqpoint{0.393701in}{0.393701in}}%
\pgfpathlineto{\pgfqpoint{0.000000in}{0.393701in}}%
\pgfpathclose%
\pgfusepath{}%
\end{pgfscope}%
\begin{pgfscope}%
\pgfpathrectangle{\pgfqpoint{0.000000in}{0.000000in}}{\pgfqpoint{0.393701in}{0.393701in}}%
\pgfusepath{clip}%
\pgfsetroundcap%
\pgfsetroundjoin%
\pgfsetlinewidth{1.003750pt}%
\definecolor{currentstroke}{rgb}{0.121569,0.466667,0.705882}%
\pgfsetstrokecolor{currentstroke}%
\pgfsetdash{}{0pt}%
\pgfpathmoveto{\pgfqpoint{0.017896in}{0.017896in}}%
\pgfpathlineto{\pgfqpoint{0.075449in}{0.020901in}}%
\pgfpathlineto{\pgfqpoint{0.102427in}{0.024978in}}%
\pgfpathlineto{\pgfqpoint{0.120412in}{0.030490in}}%
\pgfpathlineto{\pgfqpoint{0.133002in}{0.037173in}}%
\pgfpathlineto{\pgfqpoint{0.143793in}{0.046344in}}%
\pgfpathlineto{\pgfqpoint{0.152786in}{0.058157in}}%
\pgfpathlineto{\pgfqpoint{0.159980in}{0.072067in}}%
\pgfpathlineto{\pgfqpoint{0.167175in}{0.092284in}}%
\pgfpathlineto{\pgfqpoint{0.174369in}{0.122570in}}%
\pgfpathlineto{\pgfqpoint{0.181563in}{0.169593in}}%
\pgfpathlineto{\pgfqpoint{0.186959in}{0.222958in}}%
\pgfpathlineto{\pgfqpoint{0.192354in}{0.301723in}}%
\pgfpathlineto{\pgfqpoint{0.197750in}{0.375805in}}%
\pgfpathlineto{\pgfqpoint{0.204944in}{0.245782in}}%
\pgfpathlineto{\pgfqpoint{0.212138in}{0.169593in}}%
\pgfpathlineto{\pgfqpoint{0.219332in}{0.122570in}}%
\pgfpathlineto{\pgfqpoint{0.226526in}{0.092284in}}%
\pgfpathlineto{\pgfqpoint{0.235519in}{0.068105in}}%
\pgfpathlineto{\pgfqpoint{0.244512in}{0.052834in}}%
\pgfpathlineto{\pgfqpoint{0.255303in}{0.041236in}}%
\pgfpathlineto{\pgfqpoint{0.267893in}{0.032964in}}%
\pgfpathlineto{\pgfqpoint{0.282281in}{0.027309in}}%
\pgfpathlineto{\pgfqpoint{0.302065in}{0.022944in}}%
\pgfpathlineto{\pgfqpoint{0.332640in}{0.019716in}}%
\pgfpathlineto{\pgfqpoint{0.375806in}{0.017896in}}%
\pgfpathlineto{\pgfqpoint{0.375806in}{0.017896in}}%
\pgfusepath{stroke}%
\end{pgfscope}%
\begin{pgfscope}%
\pgfsetrectcap%
\pgfsetmiterjoin%
\pgfsetlinewidth{1.254687pt}%
\definecolor{currentstroke}{rgb}{0.150000,0.150000,0.150000}%
\pgfsetstrokecolor{currentstroke}%
\pgfsetdash{}{0pt}%
\pgfpathmoveto{\pgfqpoint{0.196851in}{0.000000in}}%
\pgfpathlineto{\pgfqpoint{0.196851in}{0.393701in}}%
\pgfusepath{stroke}%
\end{pgfscope}%
\begin{pgfscope}%
\pgfsetrectcap%
\pgfsetmiterjoin%
\pgfsetlinewidth{1.254687pt}%
\definecolor{currentstroke}{rgb}{0.150000,0.150000,0.150000}%
\pgfsetstrokecolor{currentstroke}%
\pgfsetdash{}{0pt}%
\pgfpathmoveto{\pgfqpoint{0.000000in}{0.000000in}}%
\pgfpathlineto{\pgfqpoint{0.393701in}{0.000000in}}%
\pgfusepath{stroke}%
\end{pgfscope}%
\end{pgfpicture}%
\makeatother%
\endgroup%
 &

\begingroup%
\makeatletter%
\begin{pgfpicture}%
\pgfpathrectangle{\pgfpointorigin}{\pgfqpoint{0.393701in}{0.393701in}}%
\pgfusepath{use as bounding box, clip}%
\begin{pgfscope}%
\pgfsetbuttcap%
\pgfsetmiterjoin%
\pgfsetlinewidth{0.000000pt}%
\definecolor{currentstroke}{rgb}{0.000000,0.000000,0.000000}%
\pgfsetstrokecolor{currentstroke}%
\pgfsetstrokeopacity{0.000000}%
\pgfsetdash{}{0pt}%
\pgfpathmoveto{\pgfqpoint{0.000000in}{0.000000in}}%
\pgfpathlineto{\pgfqpoint{0.393701in}{0.000000in}}%
\pgfpathlineto{\pgfqpoint{0.393701in}{0.393701in}}%
\pgfpathlineto{\pgfqpoint{0.000000in}{0.393701in}}%
\pgfpathclose%
\pgfusepath{}%
\end{pgfscope}%
\begin{pgfscope}%
\pgfsetbuttcap%
\pgfsetmiterjoin%
\pgfsetlinewidth{0.000000pt}%
\definecolor{currentstroke}{rgb}{0.000000,0.000000,0.000000}%
\pgfsetstrokecolor{currentstroke}%
\pgfsetstrokeopacity{0.000000}%
\pgfsetdash{}{0pt}%
\pgfpathmoveto{\pgfqpoint{0.000000in}{0.000000in}}%
\pgfpathlineto{\pgfqpoint{0.393701in}{0.000000in}}%
\pgfpathlineto{\pgfqpoint{0.393701in}{0.393701in}}%
\pgfpathlineto{\pgfqpoint{0.000000in}{0.393701in}}%
\pgfpathclose%
\pgfusepath{}%
\end{pgfscope}%
\begin{pgfscope}%
\pgfpathrectangle{\pgfqpoint{0.000000in}{0.000000in}}{\pgfqpoint{0.393701in}{0.393701in}}%
\pgfusepath{clip}%
\pgfsetroundcap%
\pgfsetroundjoin%
\pgfsetlinewidth{1.003750pt}%
\definecolor{currentstroke}{rgb}{0.121569,0.466667,0.705882}%
\pgfsetstrokecolor{currentstroke}%
\pgfsetdash{}{0pt}%
\pgfpathmoveto{\pgfqpoint{0.017896in}{0.375806in}}%
\pgfpathlineto{\pgfqpoint{0.021511in}{0.335362in}}%
\pgfpathlineto{\pgfqpoint{0.025126in}{0.300452in}}%
\pgfpathlineto{\pgfqpoint{0.028741in}{0.270191in}}%
\pgfpathlineto{\pgfqpoint{0.032357in}{0.243856in}}%
\pgfpathlineto{\pgfqpoint{0.035972in}{0.220852in}}%
\pgfpathlineto{\pgfqpoint{0.039587in}{0.200684in}}%
\pgfpathlineto{\pgfqpoint{0.043202in}{0.182944in}}%
\pgfpathlineto{\pgfqpoint{0.046818in}{0.167289in}}%
\pgfpathlineto{\pgfqpoint{0.050433in}{0.153433in}}%
\pgfpathlineto{\pgfqpoint{0.054048in}{0.141131in}}%
\pgfpathlineto{\pgfqpoint{0.057663in}{0.130181in}}%
\pgfpathlineto{\pgfqpoint{0.061279in}{0.120407in}}%
\pgfpathlineto{\pgfqpoint{0.064894in}{0.111661in}}%
\pgfpathlineto{\pgfqpoint{0.068509in}{0.103817in}}%
\pgfpathlineto{\pgfqpoint{0.072124in}{0.096764in}}%
\pgfpathlineto{\pgfqpoint{0.075740in}{0.090409in}}%
\pgfpathlineto{\pgfqpoint{0.079355in}{0.084671in}}%
\pgfpathlineto{\pgfqpoint{0.082970in}{0.079479in}}%
\pgfpathlineto{\pgfqpoint{0.086585in}{0.074771in}}%
\pgfpathlineto{\pgfqpoint{0.090201in}{0.070496in}}%
\pgfpathlineto{\pgfqpoint{0.093816in}{0.066605in}}%
\pgfpathlineto{\pgfqpoint{0.097431in}{0.063059in}}%
\pgfpathlineto{\pgfqpoint{0.101046in}{0.059821in}}%
\pgfpathlineto{\pgfqpoint{0.104662in}{0.056859in}}%
\pgfpathlineto{\pgfqpoint{0.108277in}{0.054146in}}%
\pgfpathlineto{\pgfqpoint{0.111892in}{0.051658in}}%
\pgfpathlineto{\pgfqpoint{0.115507in}{0.049371in}}%
\pgfpathlineto{\pgfqpoint{0.119123in}{0.047268in}}%
\pgfpathlineto{\pgfqpoint{0.122738in}{0.045330in}}%
\pgfpathlineto{\pgfqpoint{0.126353in}{0.043542in}}%
\pgfpathlineto{\pgfqpoint{0.129968in}{0.041891in}}%
\pgfpathlineto{\pgfqpoint{0.133584in}{0.040363in}}%
\pgfpathlineto{\pgfqpoint{0.137199in}{0.038949in}}%
\pgfpathlineto{\pgfqpoint{0.140814in}{0.037638in}}%
\pgfpathlineto{\pgfqpoint{0.144429in}{0.036421in}}%
\pgfpathlineto{\pgfqpoint{0.148045in}{0.035291in}}%
\pgfpathlineto{\pgfqpoint{0.151660in}{0.034239in}}%
\pgfpathlineto{\pgfqpoint{0.155275in}{0.033260in}}%
\pgfpathlineto{\pgfqpoint{0.158890in}{0.032348in}}%
\pgfpathlineto{\pgfqpoint{0.162506in}{0.031497in}}%
\pgfpathlineto{\pgfqpoint{0.166121in}{0.030702in}}%
\pgfpathlineto{\pgfqpoint{0.169736in}{0.029959in}}%
\pgfpathlineto{\pgfqpoint{0.173351in}{0.029264in}}%
\pgfpathlineto{\pgfqpoint{0.176967in}{0.028614in}}%
\pgfpathlineto{\pgfqpoint{0.180582in}{0.028004in}}%
\pgfpathlineto{\pgfqpoint{0.184197in}{0.027432in}}%
\pgfpathlineto{\pgfqpoint{0.187812in}{0.026895in}}%
\pgfpathlineto{\pgfqpoint{0.191428in}{0.026391in}}%
\pgfpathlineto{\pgfqpoint{0.195043in}{0.025917in}}%
\pgfpathlineto{\pgfqpoint{0.198658in}{0.025471in}}%
\pgfpathlineto{\pgfqpoint{0.202273in}{0.025051in}}%
\pgfpathlineto{\pgfqpoint{0.205889in}{0.024655in}}%
\pgfpathlineto{\pgfqpoint{0.209504in}{0.024282in}}%
\pgfpathlineto{\pgfqpoint{0.213119in}{0.023930in}}%
\pgfpathlineto{\pgfqpoint{0.216734in}{0.023597in}}%
\pgfpathlineto{\pgfqpoint{0.220350in}{0.023283in}}%
\pgfpathlineto{\pgfqpoint{0.223965in}{0.022987in}}%
\pgfpathlineto{\pgfqpoint{0.227580in}{0.022706in}}%
\pgfpathlineto{\pgfqpoint{0.231195in}{0.022440in}}%
\pgfpathlineto{\pgfqpoint{0.234811in}{0.022188in}}%
\pgfpathlineto{\pgfqpoint{0.238426in}{0.021950in}}%
\pgfpathlineto{\pgfqpoint{0.242041in}{0.021724in}}%
\pgfpathlineto{\pgfqpoint{0.245656in}{0.021509in}}%
\pgfpathlineto{\pgfqpoint{0.249272in}{0.021305in}}%
\pgfpathlineto{\pgfqpoint{0.252887in}{0.021112in}}%
\pgfpathlineto{\pgfqpoint{0.256502in}{0.020928in}}%
\pgfpathlineto{\pgfqpoint{0.260117in}{0.020753in}}%
\pgfpathlineto{\pgfqpoint{0.263733in}{0.020587in}}%
\pgfpathlineto{\pgfqpoint{0.267348in}{0.020429in}}%
\pgfpathlineto{\pgfqpoint{0.270963in}{0.020278in}}%
\pgfpathlineto{\pgfqpoint{0.274578in}{0.020134in}}%
\pgfpathlineto{\pgfqpoint{0.278194in}{0.019998in}}%
\pgfpathlineto{\pgfqpoint{0.281809in}{0.019867in}}%
\pgfpathlineto{\pgfqpoint{0.285424in}{0.019743in}}%
\pgfpathlineto{\pgfqpoint{0.289039in}{0.019624in}}%
\pgfpathlineto{\pgfqpoint{0.292655in}{0.019511in}}%
\pgfpathlineto{\pgfqpoint{0.296270in}{0.019402in}}%
\pgfpathlineto{\pgfqpoint{0.299885in}{0.019299in}}%
\pgfpathlineto{\pgfqpoint{0.303500in}{0.019200in}}%
\pgfpathlineto{\pgfqpoint{0.307116in}{0.019105in}}%
\pgfpathlineto{\pgfqpoint{0.310731in}{0.019015in}}%
\pgfpathlineto{\pgfqpoint{0.314346in}{0.018928in}}%
\pgfpathlineto{\pgfqpoint{0.317961in}{0.018845in}}%
\pgfpathlineto{\pgfqpoint{0.321577in}{0.018766in}}%
\pgfpathlineto{\pgfqpoint{0.325192in}{0.018690in}}%
\pgfpathlineto{\pgfqpoint{0.328807in}{0.018617in}}%
\pgfpathlineto{\pgfqpoint{0.332422in}{0.018547in}}%
\pgfpathlineto{\pgfqpoint{0.336038in}{0.018480in}}%
\pgfpathlineto{\pgfqpoint{0.339653in}{0.018415in}}%
\pgfpathlineto{\pgfqpoint{0.343268in}{0.018354in}}%
\pgfpathlineto{\pgfqpoint{0.346883in}{0.018294in}}%
\pgfpathlineto{\pgfqpoint{0.350499in}{0.018237in}}%
\pgfpathlineto{\pgfqpoint{0.354114in}{0.018183in}}%
\pgfpathlineto{\pgfqpoint{0.357729in}{0.018130in}}%
\pgfpathlineto{\pgfqpoint{0.361344in}{0.018079in}}%
\pgfpathlineto{\pgfqpoint{0.364960in}{0.018031in}}%
\pgfpathlineto{\pgfqpoint{0.368575in}{0.017984in}}%
\pgfpathlineto{\pgfqpoint{0.372190in}{0.017939in}}%
\pgfpathlineto{\pgfqpoint{0.375805in}{0.017896in}}%
\pgfusepath{stroke}%
\end{pgfscope}%
\begin{pgfscope}%
\pgfsetrectcap%
\pgfsetmiterjoin%
\pgfsetlinewidth{1.254687pt}%
\definecolor{currentstroke}{rgb}{0.150000,0.150000,0.150000}%
\pgfsetstrokecolor{currentstroke}%
\pgfsetdash{}{0pt}%
\pgfpathmoveto{\pgfqpoint{0.000000in}{0.000000in}}%
\pgfpathlineto{\pgfqpoint{0.000000in}{0.393701in}}%
\pgfusepath{stroke}%
\end{pgfscope}%
\begin{pgfscope}%
\pgfsetrectcap%
\pgfsetmiterjoin%
\pgfsetlinewidth{1.254687pt}%
\definecolor{currentstroke}{rgb}{0.150000,0.150000,0.150000}%
\pgfsetstrokecolor{currentstroke}%
\pgfsetdash{}{0pt}%
\pgfpathmoveto{\pgfqpoint{0.000000in}{0.000000in}}%
\pgfpathlineto{\pgfqpoint{0.393701in}{0.000000in}}%
\pgfusepath{stroke}%
\end{pgfscope}%
\end{pgfpicture}%
\makeatother%
\endgroup%
 &

\begingroup%
\makeatletter%
\begin{pgfpicture}%
\pgfpathrectangle{\pgfpointorigin}{\pgfqpoint{0.393701in}{0.393701in}}%
\pgfusepath{use as bounding box, clip}%
\begin{pgfscope}%
\pgfsetbuttcap%
\pgfsetmiterjoin%
\pgfsetlinewidth{0.000000pt}%
\definecolor{currentstroke}{rgb}{0.000000,0.000000,0.000000}%
\pgfsetstrokecolor{currentstroke}%
\pgfsetstrokeopacity{0.000000}%
\pgfsetdash{}{0pt}%
\pgfpathmoveto{\pgfqpoint{0.000000in}{0.000000in}}%
\pgfpathlineto{\pgfqpoint{0.393701in}{0.000000in}}%
\pgfpathlineto{\pgfqpoint{0.393701in}{0.393701in}}%
\pgfpathlineto{\pgfqpoint{0.000000in}{0.393701in}}%
\pgfpathclose%
\pgfusepath{}%
\end{pgfscope}%
\begin{pgfscope}%
\pgfsetbuttcap%
\pgfsetmiterjoin%
\pgfsetlinewidth{0.000000pt}%
\definecolor{currentstroke}{rgb}{0.000000,0.000000,0.000000}%
\pgfsetstrokecolor{currentstroke}%
\pgfsetstrokeopacity{0.000000}%
\pgfsetdash{}{0pt}%
\pgfpathmoveto{\pgfqpoint{0.000000in}{0.000000in}}%
\pgfpathlineto{\pgfqpoint{0.393701in}{0.000000in}}%
\pgfpathlineto{\pgfqpoint{0.393701in}{0.393701in}}%
\pgfpathlineto{\pgfqpoint{0.000000in}{0.393701in}}%
\pgfpathclose%
\pgfusepath{}%
\end{pgfscope}%
\begin{pgfscope}%
\pgfpathrectangle{\pgfqpoint{0.000000in}{0.000000in}}{\pgfqpoint{0.393701in}{0.393701in}}%
\pgfusepath{clip}%
\pgfsetroundcap%
\pgfsetroundjoin%
\pgfsetlinewidth{1.003750pt}%
\definecolor{currentstroke}{rgb}{0.121569,0.466667,0.705882}%
\pgfsetstrokecolor{currentstroke}%
\pgfsetdash{}{0pt}%
\pgfpathmoveto{\pgfqpoint{0.017896in}{0.017896in}}%
\pgfpathlineto{\pgfqpoint{0.115017in}{0.020153in}}%
\pgfpathlineto{\pgfqpoint{0.152786in}{0.023876in}}%
\pgfpathlineto{\pgfqpoint{0.174369in}{0.028932in}}%
\pgfpathlineto{\pgfqpoint{0.188757in}{0.035209in}}%
\pgfpathlineto{\pgfqpoint{0.199548in}{0.042901in}}%
\pgfpathlineto{\pgfqpoint{0.208541in}{0.052711in}}%
\pgfpathlineto{\pgfqpoint{0.217534in}{0.067704in}}%
\pgfpathlineto{\pgfqpoint{0.224728in}{0.085803in}}%
\pgfpathlineto{\pgfqpoint{0.231922in}{0.112806in}}%
\pgfpathlineto{\pgfqpoint{0.239116in}{0.154531in}}%
\pgfpathlineto{\pgfqpoint{0.246310in}{0.221746in}}%
\pgfpathlineto{\pgfqpoint{0.251706in}{0.300950in}}%
\pgfpathlineto{\pgfqpoint{0.255303in}{0.375806in}}%
\pgfpathlineto{\pgfqpoint{0.257102in}{0.361749in}}%
\pgfpathlineto{\pgfqpoint{0.264296in}{0.236675in}}%
\pgfpathlineto{\pgfqpoint{0.271490in}{0.163562in}}%
\pgfpathlineto{\pgfqpoint{0.278684in}{0.118527in}}%
\pgfpathlineto{\pgfqpoint{0.285878in}{0.089567in}}%
\pgfpathlineto{\pgfqpoint{0.294871in}{0.066483in}}%
\pgfpathlineto{\pgfqpoint{0.303864in}{0.051924in}}%
\pgfpathlineto{\pgfqpoint{0.314655in}{0.040881in}}%
\pgfpathlineto{\pgfqpoint{0.327245in}{0.033014in}}%
\pgfpathlineto{\pgfqpoint{0.343432in}{0.027145in}}%
\pgfpathlineto{\pgfqpoint{0.365014in}{0.022989in}}%
\pgfpathlineto{\pgfqpoint{0.375806in}{0.021738in}}%
\pgfpathlineto{\pgfqpoint{0.375806in}{0.021738in}}%
\pgfusepath{stroke}%
\end{pgfscope}%
\begin{pgfscope}%
\pgfsetrectcap%
\pgfsetmiterjoin%
\pgfsetlinewidth{1.254687pt}%
\definecolor{currentstroke}{rgb}{0.150000,0.150000,0.150000}%
\pgfsetstrokecolor{currentstroke}%
\pgfsetdash{}{0pt}%
\pgfpathmoveto{\pgfqpoint{0.196851in}{-0.000000in}}%
\pgfpathlineto{\pgfqpoint{0.196851in}{0.393701in}}%
\pgfusepath{stroke}%
\end{pgfscope}%
\begin{pgfscope}%
\pgfsetrectcap%
\pgfsetmiterjoin%
\pgfsetlinewidth{1.254687pt}%
\definecolor{currentstroke}{rgb}{0.150000,0.150000,0.150000}%
\pgfsetstrokecolor{currentstroke}%
\pgfsetdash{}{0pt}%
\pgfpathmoveto{\pgfqpoint{0.000000in}{0.000000in}}%
\pgfpathlineto{\pgfqpoint{0.393701in}{0.000000in}}%
\pgfusepath{stroke}%
\end{pgfscope}%
\end{pgfpicture}%
\makeatother%
\endgroup%
 &

\begingroup%
\makeatletter%
\begin{pgfpicture}%
\pgfpathrectangle{\pgfpointorigin}{\pgfqpoint{0.393701in}{0.393701in}}%
\pgfusepath{use as bounding box, clip}%
\begin{pgfscope}%
\pgfsetbuttcap%
\pgfsetmiterjoin%
\pgfsetlinewidth{0.000000pt}%
\definecolor{currentstroke}{rgb}{0.000000,0.000000,0.000000}%
\pgfsetstrokecolor{currentstroke}%
\pgfsetstrokeopacity{0.000000}%
\pgfsetdash{}{0pt}%
\pgfpathmoveto{\pgfqpoint{0.000000in}{0.000000in}}%
\pgfpathlineto{\pgfqpoint{0.393701in}{0.000000in}}%
\pgfpathlineto{\pgfqpoint{0.393701in}{0.393701in}}%
\pgfpathlineto{\pgfqpoint{0.000000in}{0.393701in}}%
\pgfpathclose%
\pgfusepath{}%
\end{pgfscope}%
\begin{pgfscope}%
\pgfsetbuttcap%
\pgfsetmiterjoin%
\pgfsetlinewidth{0.000000pt}%
\definecolor{currentstroke}{rgb}{0.000000,0.000000,0.000000}%
\pgfsetstrokecolor{currentstroke}%
\pgfsetstrokeopacity{0.000000}%
\pgfsetdash{}{0pt}%
\pgfpathmoveto{\pgfqpoint{0.000000in}{0.000000in}}%
\pgfpathlineto{\pgfqpoint{0.393701in}{0.000000in}}%
\pgfpathlineto{\pgfqpoint{0.393701in}{0.393701in}}%
\pgfpathlineto{\pgfqpoint{0.000000in}{0.393701in}}%
\pgfpathclose%
\pgfusepath{}%
\end{pgfscope}%
\begin{pgfscope}%
\pgfpathrectangle{\pgfqpoint{0.000000in}{0.000000in}}{\pgfqpoint{0.393701in}{0.393701in}}%
\pgfusepath{clip}%
\pgfsetroundcap%
\pgfsetroundjoin%
\pgfsetlinewidth{1.003750pt}%
\definecolor{currentstroke}{rgb}{0.839216,0.152941,0.156863}%
\pgfsetstrokecolor{currentstroke}%
\pgfsetdash{}{0pt}%
\pgfpathmoveto{\pgfqpoint{0.017896in}{0.037497in}}%
\pgfpathlineto{\pgfqpoint{0.021511in}{0.039037in}}%
\pgfpathlineto{\pgfqpoint{0.025126in}{0.040703in}}%
\pgfpathlineto{\pgfqpoint{0.028741in}{0.042506in}}%
\pgfpathlineto{\pgfqpoint{0.032357in}{0.044460in}}%
\pgfpathlineto{\pgfqpoint{0.035972in}{0.046582in}}%
\pgfpathlineto{\pgfqpoint{0.039587in}{0.048888in}}%
\pgfpathlineto{\pgfqpoint{0.043202in}{0.051398in}}%
\pgfpathlineto{\pgfqpoint{0.046818in}{0.054134in}}%
\pgfpathlineto{\pgfqpoint{0.050433in}{0.057121in}}%
\pgfpathlineto{\pgfqpoint{0.054048in}{0.060387in}}%
\pgfpathlineto{\pgfqpoint{0.057663in}{0.063963in}}%
\pgfpathlineto{\pgfqpoint{0.061279in}{0.067887in}}%
\pgfpathlineto{\pgfqpoint{0.064894in}{0.072199in}}%
\pgfpathlineto{\pgfqpoint{0.068509in}{0.076947in}}%
\pgfpathlineto{\pgfqpoint{0.072124in}{0.082183in}}%
\pgfpathlineto{\pgfqpoint{0.075740in}{0.087970in}}%
\pgfpathlineto{\pgfqpoint{0.079355in}{0.094380in}}%
\pgfpathlineto{\pgfqpoint{0.082970in}{0.101493in}}%
\pgfpathlineto{\pgfqpoint{0.086585in}{0.109404in}}%
\pgfpathlineto{\pgfqpoint{0.090201in}{0.118225in}}%
\pgfpathlineto{\pgfqpoint{0.093816in}{0.128082in}}%
\pgfpathlineto{\pgfqpoint{0.097431in}{0.139126in}}%
\pgfpathlineto{\pgfqpoint{0.101046in}{0.151532in}}%
\pgfpathlineto{\pgfqpoint{0.104662in}{0.165508in}}%
\pgfpathlineto{\pgfqpoint{0.108277in}{0.181296in}}%
\pgfpathlineto{\pgfqpoint{0.111892in}{0.199188in}}%
\pgfpathlineto{\pgfqpoint{0.115507in}{0.219527in}}%
\pgfpathlineto{\pgfqpoint{0.119123in}{0.242729in}}%
\pgfpathlineto{\pgfqpoint{0.122738in}{0.269289in}}%
\pgfpathlineto{\pgfqpoint{0.126353in}{0.299808in}}%
\pgfpathlineto{\pgfqpoint{0.129968in}{0.335016in}}%
\pgfpathlineto{\pgfqpoint{0.133584in}{0.375806in}}%
\pgfpathlineto{\pgfqpoint{0.137199in}{0.375806in}}%
\pgfpathlineto{\pgfqpoint{0.140814in}{0.335016in}}%
\pgfpathlineto{\pgfqpoint{0.144429in}{0.299808in}}%
\pgfpathlineto{\pgfqpoint{0.148045in}{0.269289in}}%
\pgfpathlineto{\pgfqpoint{0.151660in}{0.242729in}}%
\pgfpathlineto{\pgfqpoint{0.155275in}{0.219527in}}%
\pgfpathlineto{\pgfqpoint{0.158890in}{0.199188in}}%
\pgfpathlineto{\pgfqpoint{0.162506in}{0.181296in}}%
\pgfpathlineto{\pgfqpoint{0.166121in}{0.165508in}}%
\pgfpathlineto{\pgfqpoint{0.169736in}{0.151532in}}%
\pgfpathlineto{\pgfqpoint{0.173351in}{0.139126in}}%
\pgfpathlineto{\pgfqpoint{0.176967in}{0.128082in}}%
\pgfpathlineto{\pgfqpoint{0.180582in}{0.118225in}}%
\pgfpathlineto{\pgfqpoint{0.184197in}{0.109404in}}%
\pgfpathlineto{\pgfqpoint{0.187812in}{0.101493in}}%
\pgfpathlineto{\pgfqpoint{0.191428in}{0.094380in}}%
\pgfpathlineto{\pgfqpoint{0.195043in}{0.087970in}}%
\pgfpathlineto{\pgfqpoint{0.198658in}{0.082183in}}%
\pgfpathlineto{\pgfqpoint{0.202273in}{0.076947in}}%
\pgfpathlineto{\pgfqpoint{0.205889in}{0.072199in}}%
\pgfpathlineto{\pgfqpoint{0.209504in}{0.067887in}}%
\pgfpathlineto{\pgfqpoint{0.213119in}{0.063963in}}%
\pgfpathlineto{\pgfqpoint{0.216734in}{0.060387in}}%
\pgfpathlineto{\pgfqpoint{0.220350in}{0.057121in}}%
\pgfpathlineto{\pgfqpoint{0.223965in}{0.054134in}}%
\pgfpathlineto{\pgfqpoint{0.227580in}{0.051398in}}%
\pgfpathlineto{\pgfqpoint{0.231195in}{0.048888in}}%
\pgfpathlineto{\pgfqpoint{0.234811in}{0.046582in}}%
\pgfpathlineto{\pgfqpoint{0.238426in}{0.044460in}}%
\pgfpathlineto{\pgfqpoint{0.242041in}{0.042506in}}%
\pgfpathlineto{\pgfqpoint{0.245656in}{0.040703in}}%
\pgfpathlineto{\pgfqpoint{0.249272in}{0.039037in}}%
\pgfpathlineto{\pgfqpoint{0.252887in}{0.037497in}}%
\pgfpathlineto{\pgfqpoint{0.256502in}{0.036071in}}%
\pgfpathlineto{\pgfqpoint{0.260117in}{0.034748in}}%
\pgfpathlineto{\pgfqpoint{0.263733in}{0.033521in}}%
\pgfpathlineto{\pgfqpoint{0.267348in}{0.032381in}}%
\pgfpathlineto{\pgfqpoint{0.270963in}{0.031321in}}%
\pgfpathlineto{\pgfqpoint{0.274578in}{0.030333in}}%
\pgfpathlineto{\pgfqpoint{0.278194in}{0.029413in}}%
\pgfpathlineto{\pgfqpoint{0.281809in}{0.028555in}}%
\pgfpathlineto{\pgfqpoint{0.285424in}{0.027753in}}%
\pgfpathlineto{\pgfqpoint{0.289039in}{0.027004in}}%
\pgfpathlineto{\pgfqpoint{0.292655in}{0.026303in}}%
\pgfpathlineto{\pgfqpoint{0.296270in}{0.025647in}}%
\pgfpathlineto{\pgfqpoint{0.299885in}{0.025032in}}%
\pgfpathlineto{\pgfqpoint{0.303500in}{0.024455in}}%
\pgfpathlineto{\pgfqpoint{0.307116in}{0.023914in}}%
\pgfpathlineto{\pgfqpoint{0.310731in}{0.023405in}}%
\pgfpathlineto{\pgfqpoint{0.314346in}{0.022927in}}%
\pgfpathlineto{\pgfqpoint{0.317961in}{0.022477in}}%
\pgfpathlineto{\pgfqpoint{0.321577in}{0.022054in}}%
\pgfpathlineto{\pgfqpoint{0.325192in}{0.021654in}}%
\pgfpathlineto{\pgfqpoint{0.328807in}{0.021278in}}%
\pgfpathlineto{\pgfqpoint{0.332422in}{0.020923in}}%
\pgfpathlineto{\pgfqpoint{0.336038in}{0.020588in}}%
\pgfpathlineto{\pgfqpoint{0.339653in}{0.020271in}}%
\pgfpathlineto{\pgfqpoint{0.343268in}{0.019972in}}%
\pgfpathlineto{\pgfqpoint{0.346883in}{0.019689in}}%
\pgfpathlineto{\pgfqpoint{0.350499in}{0.019421in}}%
\pgfpathlineto{\pgfqpoint{0.354114in}{0.019167in}}%
\pgfpathlineto{\pgfqpoint{0.357729in}{0.018926in}}%
\pgfpathlineto{\pgfqpoint{0.361344in}{0.018698in}}%
\pgfpathlineto{\pgfqpoint{0.364960in}{0.018481in}}%
\pgfpathlineto{\pgfqpoint{0.368575in}{0.018276in}}%
\pgfpathlineto{\pgfqpoint{0.372190in}{0.018081in}}%
\pgfpathlineto{\pgfqpoint{0.375805in}{0.017896in}}%
\pgfusepath{stroke}%
\end{pgfscope}%
\begin{pgfscope}%
\pgfsetrectcap%
\pgfsetmiterjoin%
\pgfsetlinewidth{1.254687pt}%
\definecolor{currentstroke}{rgb}{0.150000,0.150000,0.150000}%
\pgfsetstrokecolor{currentstroke}%
\pgfsetdash{}{0pt}%
\pgfpathmoveto{\pgfqpoint{0.000000in}{-0.000000in}}%
\pgfpathlineto{\pgfqpoint{0.000000in}{0.393701in}}%
\pgfusepath{stroke}%
\end{pgfscope}%
\begin{pgfscope}%
\pgfsetrectcap%
\pgfsetmiterjoin%
\pgfsetlinewidth{1.254687pt}%
\definecolor{currentstroke}{rgb}{0.150000,0.150000,0.150000}%
\pgfsetstrokecolor{currentstroke}%
\pgfsetdash{}{0pt}%
\pgfpathmoveto{\pgfqpoint{0.000000in}{0.000000in}}%
\pgfpathlineto{\pgfqpoint{0.393701in}{0.000000in}}%
\pgfusepath{stroke}%
\end{pgfscope}%
\end{pgfpicture}%
\makeatother%
\endgroup%
\tabularnewline
  \bottomrule
  \end{tabular}
  \caption{\label{tab:priors}Overview of the prior distributions. Rows refer to types of distributions and columns refer to properties of distributions. A cell of the table refers to a specific family of distributions which is depicted by the density of an example element of the family. Currently implemented prior distributions are plotted in blue and not yet implemented prior distributions are plotted in red.}
\end{table}

\subsection{Data}
\label{sec:data}
\topic{Description of the synthetic data set.}
The synthetic data set in Figure \FIG{hypest} consists of three sources plus background noise. Each source is an outer product of a dense filter and a sparse filter. The dense factors are normalised, non-negative, zero mean, and normally distributed. The sparse filters are exponentially distributed with scales $\beta_1$, $\beta_2$, and $\beta_3$, respectively. The signal is perturbed by independent Gaussian noise with variance $\sigma^2$. The filters have a size of 1000 elements each which leads to data sets of size 1000 $\times$ 1000. The results in Figure \FIG{hypest} are averaged across 100 randomly sampled data sets.

\topic{Description of the experimental data set.}
The experimental data set (used in \FIG{exdec}, \FIG{exdecall}, and \FIG{flops}) is a publicly available two-photon calcium imaging recording of a population of neurons \cite{frady2015}. The data set is analogously recorded to the data sets published in \cite{peters2014}. The data captures the activity of a population of layer 2/3 neurons in the motor cortex of a behaving mouse using the GCaMP5G indicator. The field of view spans \SI[product-units = repeat]{472 x 502}{\micro \metre} using a resolution of 512 $\times$ 512 pixels. The duration of the recording is about 2 minutes using a sampling rate of 28 Hz. The publicly available data set is subsampled to a spatial resolution of 128 $\times$ 128 pixel and a temporal resolution of 500 frames. For more information about the experimental protocol please consult \cite{peters2014}.

\topic{Description of the augmented experimental data set.}
The augmented data sets used in \FIG{srcRec} are based on the experimental data set introduced above. The augmentation is performed by a superposition of the real data with a ground truth cell of variance $\sigma^2_{GT}$. The result of the augmentation is a realistic data set $\mathbf{X}_{\sigma^2_{GT}}$ for which ground truth information is available.

\begin{figure}
    \centering
    \input{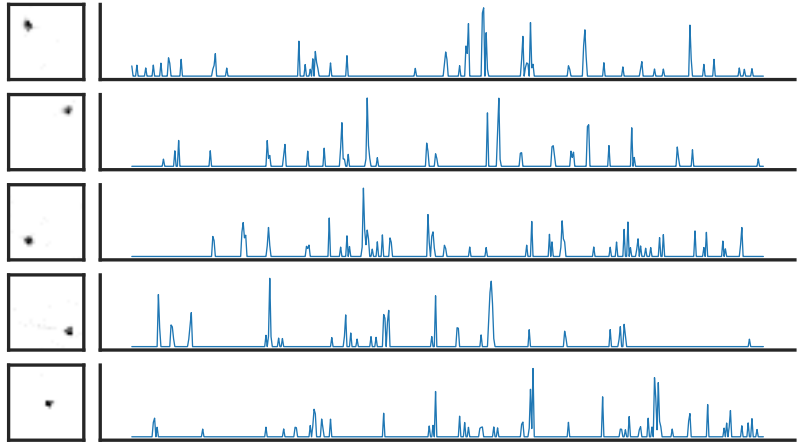}
    \caption{Temporal and spatial filters of the ground truth cells.}
    \label{fig:groundTruth}
\end{figure}

The augmentation process starts with spatially splitting the experimental data set into two parts: the first part of the data ($\mathbf{X}_{in}$) has a reduced spatial extend to cover only a small patch of 32 $\times$ 32 pixels. The second part of the data ($\mathbf{X}_{out}$) covers all pixels not included in the first part. On the latter part we run our method using a lomax - exponential prior combination. Among the separated sources we pick five sources that are very likely related to a neuron and consider them as ground truth cells (listed in \FIG{groundTruth}). We crop the spatial filters of the ground truth cells such that each cell lies inside a 32 $\times$ 32 window. Finally, we adjust the variance of each ground truth cell $S_k$ to match a certain target variance $\sigma^2_{GT}$ and superimpose them on $\mathbf{X}_{in}$: \begin{equation}
    \mathbf{X}_{\sigma^2_{GT}} = \sum_{k=1}^5 S_kc_k + \mathbf{X}_{in}\text{.}
\end{equation}

This technique results in ground truth data that is very realistic with respect to noise, background activity, and the spatial and temporal correlations among sources. Further, the technique allows us to control the separation difficulty by varying the variance of the ground truth source $\sigma^2_{GT}$ using a scaling coefficient $c_k$.

\subsection{Performance measure} \label{sec:quantify}
\topic{Reason to define a measure that demands ground truth information.}
Our goal is to use the ground truth data set to objectively evaluate the performance of different BSS models. The performance measure should capture how good different models are able to recover the ground-truth sources. Let $S_i$ be the ground truth cell and $\hat S_k$ the $k$-th source recovered by a model. The pearson correlation $\rho(S_i, \hat S_k)$ is a good measure for how well the two match and its absolute value is normalised to be between 0 (no match at all) and 1 (perfect match). 
For each run $n$ and ground-truth cell $S_i$ we determine the highest pearson correlation $\max_k \rho(S_i, \hat S_k^{(n)})$. The final score of a model is defined as the average over the medians of the highest correlations:
\begin{equation}
    \mathrm{score}(\mathrm{model}) = \mean_i\left(\median_n\left(\max_{k}\rho\left(S_{i}, \hat S_k^{(n)}\right)\right)\right).
\end{equation}

\section{Acknowledgments}
BE was supported by a European Commission Marie Sklodowska Curie grant (660328). WB was supported by a grant of the Carl Zeiss
Foundation (0563-2.8/558/3). AB and MB were supported by the DFG-funded CRC 1233 Robust Vision (sub project 12).

\nocite{*}
\bibliography{biblio}

\end{document}